%% file: acl_latex.tex
\documentclass[11pt]{article}

\usepackage[preprint]{acl}

\usepackage{times}
\usepackage{latexsym}

\usepackage[T1]{fontenc}
\usepackage[utf8]{inputenc}
\usepackage{microtype}
\usepackage{inconsolata}

\usepackage{amsmath}
\usepackage{amssymb}
\usepackage{booktabs}
\usepackage{multirow}
\usepackage{array}
\usepackage{graphicx}
\usepackage{xspace}
\usepackage{arydshln}
\usepackage[most]{tcolorbox}
\usepackage{xurl}

\newcommand{\method}{CoSTA\xspace}

\newcommand{\E}{\mathbf{E}}
\newcommand{\W}{\mathbf{W}}

\newcommand{\Dtts}{\mathcal{D}_{\mathrm{tts}}}
\newcommand{\Ddlg}{\mathcal{D}_{\mathrm{dlg}}}

\title{CoSTA: \underline{Co}mpressed \underline{S}peech \underline{T}oken \underline{A}daptation for Efficient Spoken Dialogue Models}

\author{
\textbf{Yingzhuo Deng}$^{1,2}$, 
\textbf{Chong Li}$^{1,2}$, 
\textbf{Bingxin Liu}$^{1,2}$, 
\textbf{Yi Shu}$^{2}$, 
\textbf{Tianyu Peng}$^{1,2}$,\\
\textbf{Can Xie}$^{1,2}$, 
\textbf{Siyao Song}$^{1,2}$, 
\textbf{Shuo Ren}$^{1}$, 
\textbf{Jiajun Zhang}$^{1,2,3}$\thanks{Corresponding author.}\\
$^{1}$Institute of Automation, Chinese Academy of Sciences\\
$^{2}$School of Artificial Intelligence, University of Chinese Academy of Sciences\\
$^{3}$Wuhan AI Research\\
\texttt{dengyingzhuo2024@ia.ac.cn}; \texttt{jjzhang@nlpr.ia.ac.cn}
}

\begin{document}
\maketitle

\begin{abstract}
Spoken Dialogue Models (SDMs) typically share a speech-token space for generation and synthesis, leading to long autoregressive sequences and costly inference.
Lower-rate speech tokenizers shorten sequences but may introduce compression--quality trade-offs and costly replacement without breaking this coupling.
To address this limitation, we propose \method, a compatibility-preserving post-hoc framework that decouples generation and synthesis token spaces while retaining the original speech tokenizer and decoder.
\method extends the generation token space with compressed representations via mean-initialized vocabulary extension, pronunciation adaptation, and dialogue adaptation, followed by deterministic expansion back to the original speech-token space before synthesis.
On bilingual speech-dialogue benchmarks, \method achieves approximately $1.7\times$ expanded speech-token generation speed, lower full-response latency, and higher throughput while preserving response quality.
Analyses show more sustained interaction under limited budgets and a practical trade-off among compression, decoding efficiency, and quality.
Further experiments demonstrate cross-architecture transfer and complementarity to low-rate tokenization.\footnote{Code: \url{https://github.com/yingzhuodeng/CoSTA}}
\end{abstract}

\section{Introduction}

\begin{figure}[t]
    \centering
    \includegraphics[width=\linewidth]{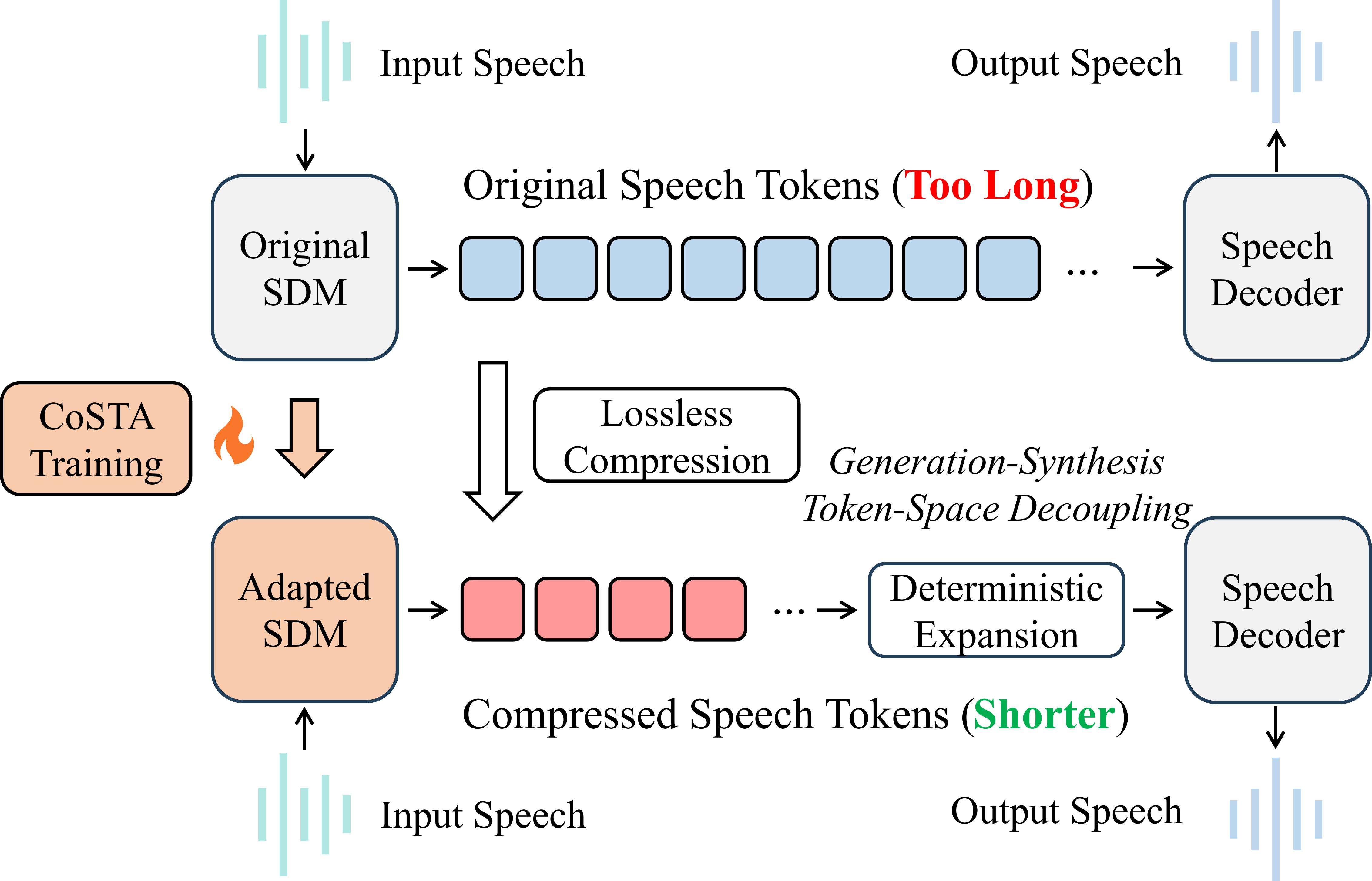}
    \vspace{-2.1em}
    \caption{Overview of \method. Existing SDMs couple generation and synthesis token spaces. \method extends the generation token space with compressed representations and deterministically expands them back to the original space before synthesis.}
    \label{fig:first}
    \vspace{-1.6em}
\end{figure}

Spoken dialogue models (SDMs) enable end-to-end speech interaction by modeling speech input and generating text tokens together with speech tokens, which are then synthesized into waveforms by a speech decoder~\cite{Moshi, GLM-4-Voice, Qwen2.5-Omni, Kimi-Audio, Step-Audio2}.
Existing SDMs typically use the same token space for autoregressive generation and speech synthesis, tightly coupling generation with the representation required by the speech decoder.
These speech-token sequences are often overly long, increasing autoregressive generation cost, complicating text-speech alignment, and making long-context speech modeling more difficult~\cite{aBPE, aBPE2}.
In practical interaction, this inefficiency slows responses and consumes generation budgets faster, limiting sustained interaction and spoken-response duration.

One solution is to design speech tokenizers with lower token rates~\cite{Moshi, Zeng, SoCodec, TaDiCodec, SpeechTokensReview}.
However, aggressive rate reduction may introduce a compression--quality trade-off, and replacing the tokenizer of an existing SDM may require retraining or adapting the tokenizer, paired speech decoder, and SDM together.
More fundamentally, this route still couples generation and synthesis to the same representation.
We therefore ask a different question:
\emph{Can we decouple the token space used for autoregressive generation from that required for speech synthesis, while preserving the original speech tokenizer and decoder?}

Speech-token sequences produced by existing tokenizers exhibit substantial redundancy and can be losslessly compressed using BPE~\cite{Gage1994BPE, Sennrich2016BPE, GPT2}.
Across diverse tokenizer, data, and language settings, we uncover an empirical scaling regularity: BPE compression shortens speech-token sequences with increasing vocabulary size, following an approximate log-linear trend on held-out speech.
This persistent compressibility suggests a route to more efficient generation without altering synthesis: using a compact autoregressive representation while deterministically recovering the original speech-token sequence required by the decoder.

This insight motivates \emph{generation--synthesis token-space decoupling} and \method, short for \textbf{Co}mpressed \textbf{S}peech \textbf{T}oken \textbf{A}daptation, a compatibility-preserving post-hoc token-space adaptation framework for existing large-scale SDMs.
As illustrated in Figure~\ref{fig:first}, \method extends the autoregressive generation token space with compressed representations while preserving the original synthesis interface.
\method adapts the SDM to generate these compressed representations through vocabulary extension with mean initialization, pronunciation adaptation, and dialogue adaptation.
At inference time, generated compressed units are deterministically expanded back to the original speech-token representations before synthesis, reducing autoregressive speech-token steps while leaving the original speech tokenizer and decoder unchanged.

We validate \method on two representative SDM architectures. 
In experiments with Step-Audio 2 mini \cite{Step-Audio2}, 
\method achieves \(1.736\times\) and \(1.659\times\) relative expanded speech-token generation speed in Chinese and English, respectively, while preserving response quality according to both automatic and human evaluations. 
These gains further translate into lower full-response latency, higher throughput, and more sustained interaction under limited budgets. 
Vocabulary scaling further reveals a practical trade-off among compression, decoding efficiency, and response quality. 
Experiments on a different SDM architecture further demonstrate cross-architecture transfer and complementarity to low-rate speech tokenization.

Our contributions are summarized as follows:
\begin{itemize}
    \item We uncover an approximate log-linear scaling regularity of speech-token compression over 10K--320K added units across eight tokenizer configurations, data sources, and languages.

    \item We formulate generation--synthesis token-space decoupling and introduce \method, a compatibility-preserving post-hoc adaptation framework that enables existing SDMs to generate compact, losslessly expandable speech-token sequences while retaining their original speech tokenizer and decoder.

    \item We demonstrate that \method improves model-side and system efficiency while preserving response quality, with additional evidence of cross-architecture transfer and complementarity to low-rate speech tokenization.

    \item We further characterize compressed-token generation: \method enables more sustained interaction under limited budgets, while vocabulary scaling reveals a practical trade-off among compression, efficiency, and quality.
\end{itemize}

\begin{figure}[t]
    \centering
    \includegraphics[width=\linewidth]{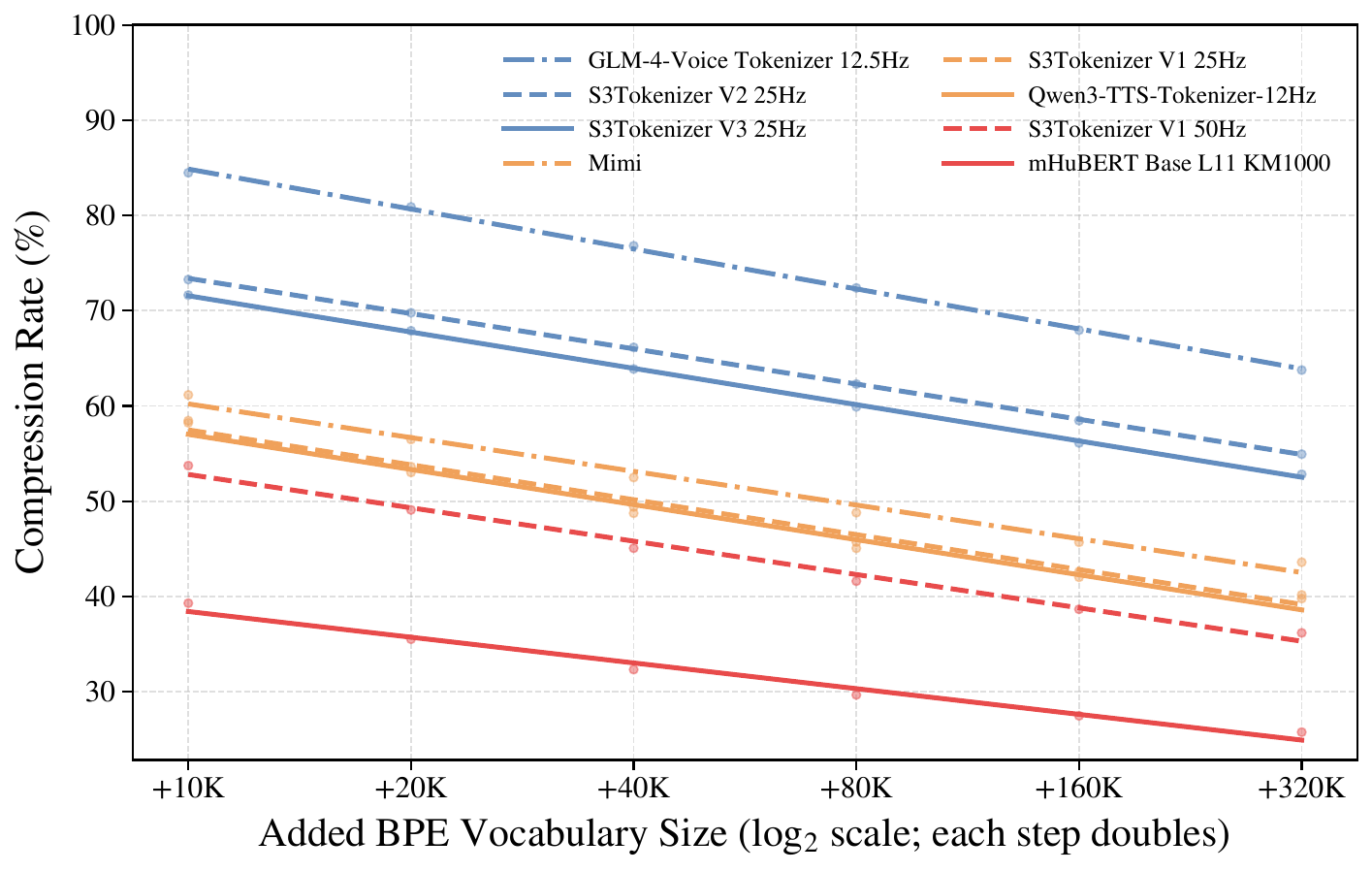}
    \vspace{-1.9em}
    \caption{Speech-token BPE scaling across eight tokenizer configurations. Compression rate decreases approximately linearly with $\log_2$ vocabulary size.}
    \label{fig:tokenizer-scaling}
    \vspace{-1.3em}
\end{figure}

\begin{figure*}[t]
    \centering
    \includegraphics[width=0.9\textwidth]{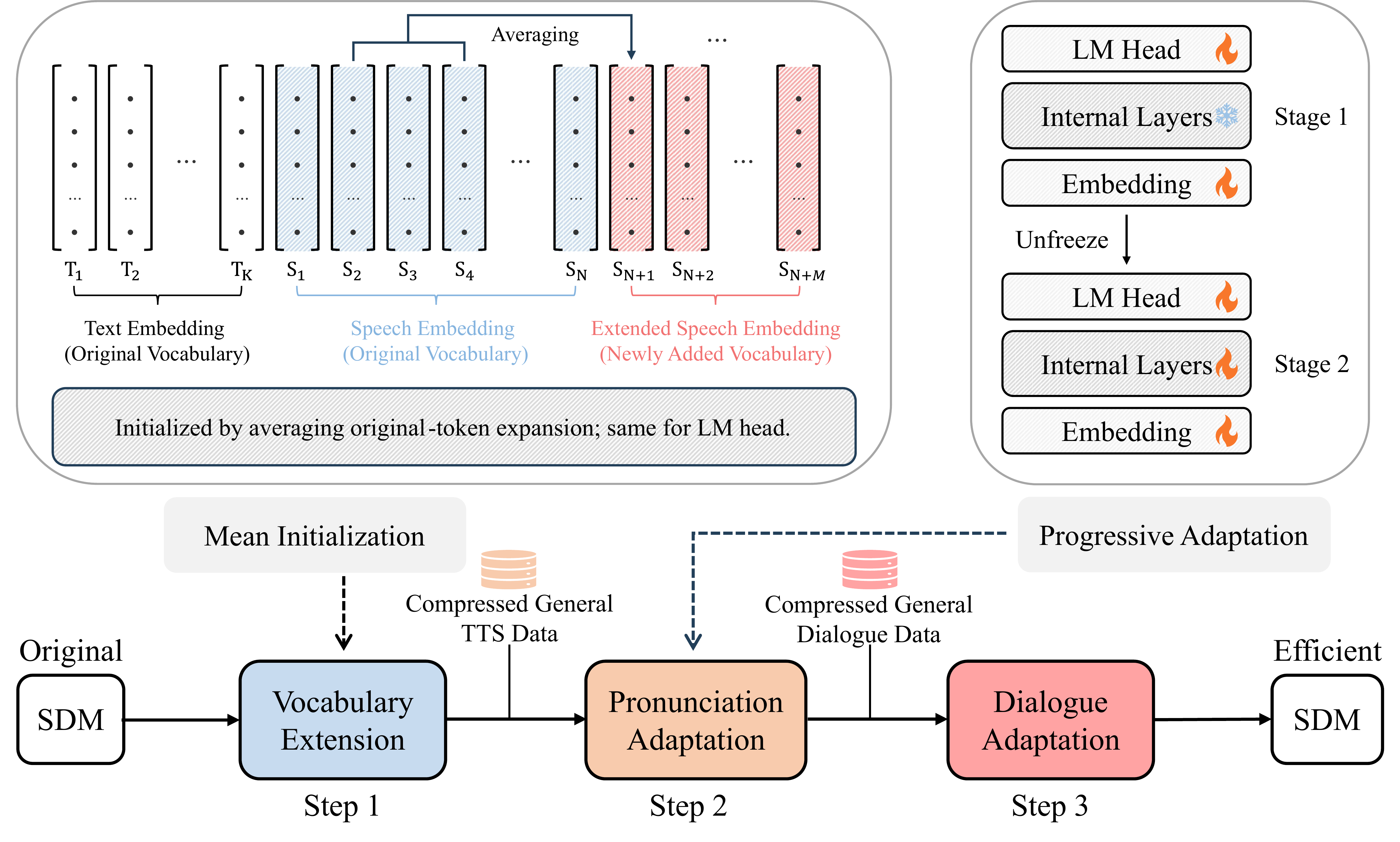}
    \vspace{-1.2em}
    \caption{\method combines vocabulary extension with mean-based initialization, progressive pronunciation adaptation, and dialogue adaptation.}
    \label{fig:method-overview}
    \vspace{-1.2em}
\end{figure*}

\section{From Speech-Token BPE Scaling to Token-Space Decoupling}
\label{sec:compressibility}

\paragraph{Lossless token compression.}
Discrete speech-token sequences contain repeated local patterns, making them naturally amenable for lossless compression.
Here, lossless refers to reversible token segmentation: the compressed sequence can be deterministically expanded back to the original sequence.
We therefore apply BPE to the original speech-token sequences.
Let $\mathcal{S}$ denote the original speech-token vocabulary and $\mathbf{s}$ a speech-token sequence.
BPE constructs an extended vocabulary $\mathcal{Z}\supseteq\mathcal{S}$ by merging frequent adjacent token pairs.
Its encoder $\mathcal{B}$ produces the compressed sequence $\mathbf{z}=\mathcal{B}(\mathbf{s})$, and the deterministic expansion function $\pi$ recovers the original sequence, i.e., $\pi(\mathbf{z})=\mathbf{s}$.
We define the compression rate as
\begin{equation}
    \rho(\mathbf{z}) =
    \frac{|\mathbf{z}|}{|\pi(\mathbf{z})|}.
    \label{eq:compression-rate}
\end{equation}
Lower $\rho$ indicates fewer autoregressive steps for generating the same sequence consumed by the decoder.
The number of added BPE units is denoted as $M=|\mathcal{Z}\setminus\mathcal{S}|$.

\paragraph{Empirical scaling regularity.}
We scale the added BPE vocabulary $M$ from 10K to 320K in powers of two across eight representative speech tokenizer configurations on Chinese dialogue speech.
For each setting, the BPE vocabulary is learned on the training split and fixed for evaluation on held-out speech.
Figure~\ref{fig:tokenizer-scaling} reveals a consistent scaling regularity: $\rho$ decreases approximately linearly with $\log_2 M$.
We describe this empirical trend as
\begin{equation}
    \rho_c(M)\approx a_c-b_c\log_2(M/10\mathrm{K}),
    \qquad b_c>0,
    \label{eq:compression-scaling}
\end{equation}
where $c$ denotes an evaluation setting.
The log-linear model achieves $R^2\geq0.98$ across eight tokenizer configurations and remains consistent across data sources and languages ($R^2\geq0.994$; Appendix~\ref{app:compression-scaling}).
Detailed settings and analyses are provided in Appendix~\ref{app:compression-diagnostics}.

\paragraph{From scaling regularity to generation--synthesis token-space decoupling.}
The original speech-token space produces long sequences for autoregressive generation.
The empirical scaling regularity reveals persistent compressibility: larger BPE vocabularies yield shorter speech-token sequences, motivating their use for autoregressive generation.
However, the speech decoder still requires the original token space.
This motivates \emph{generation--synthesis token-space decoupling}: the SDM generates compact BPE sequences in an extended token space, which are deterministically expanded back to the original space for synthesis.
We instantiate this paradigm through the post-hoc adaptation framework \method in Section~\ref{sec:method}.

\section{Method}
\label{sec:method}

Section~\ref{sec:compressibility} motivates generation--synthesis token-space decoupling.
Realizing this decoupling in an existing SDM, however, requires adapting the model while preserving compatibility with its original tokenizer--decoder pipeline.
We therefore propose \method, a compatibility-preserving post-hoc token-space adaptation framework.
\method adapts the SDM to generate compressed sequences in the extended token space $\mathcal{Z}$, while deterministic expansion $\pi(\mathbf{z})=\mathbf{s}$ restores the original decoder-facing representation in $\mathcal{S}$ before waveform synthesis.

As shown in Figure~\ref{fig:method-overview}, \method consists of three steps:
(1) vocabulary extension with mean-based initialization,
(2) pronunciation adaptation on compressed TTS data, and
(3) dialogue adaptation on compressed speech-to-speech dialogue data.
At inference time, only a lightweight deterministic BPE expansion is inserted before the unchanged speech decoder.
We detail each step below.

\subsection{Vocabulary Extension with Mean-based Initialization}
\label{subsec:mean-init}

To instantiate the extended generation token space $\mathcal{Z}$ in an existing SDM, let $\mathcal{V}$ denote its original model vocabulary, with $\mathcal{S}\subseteq\mathcal{V}$ denoting the original speech-token vocabulary.
We extend $\mathcal{V}$ with the added BPE units:
\begin{equation}
    \mathcal{V}' = \mathcal{V} \cup (\mathcal{Z} \setminus \mathcal{S}).
\end{equation}
For each added BPE unit $z\in\mathcal{Z}\setminus\mathcal{S}$, its deterministic expansion $\pi(z)=(s_1,\ldots,s_k)$ is known.
We therefore initialize its embedding $\E'_z$ and LM-head row $\W'_z$ by averaging the corresponding component rows:
\begin{equation}
\begin{array}{r@{\;}c@{\;}l@{\quad}r@{\;}c@{\;}l}
\E'_z & = & \frac{1}{k}\sum_{i=1}^{k} \E_{s_i},
&
\W'_z & = & \frac{1}{k}\sum_{i=1}^{k} \W_{s_i}.
\end{array}
\label{eq:mean-init}
\end{equation}
Original vocabulary parameters remain unchanged.
This mean-based initialization provides a more stable starting point than random initialization.

\subsection{Pronunciation Adaptation}
\label{subsec:tts-adapt}

After vocabulary extension, the SDM has an extended generation token space but has not yet learned to use the added BPE units reliably.
We therefore perform pronunciation adaptation on compressed TTS data.
Following interleaved-generation SDMs such as Step-Audio 2 mini \cite{Step-Audio2}, each target sequence contains both text and speech tokens.
We apply BPE compression only to speech tokens and leave text tokens unchanged.

Each example consists of a text prompt $p$ and a target response sequence $\mathbf{y}$, where $\mathbf{y}$ contains interleaved text tokens and original speech tokens. Let $\widetilde{\mathbf{y}} = C(\mathbf{y})$ denote the sequence obtained by applying BPE encoder $\mathcal{B}$ only to the speech tokens of $\mathbf{y}$. We train the model to predict the compressed target sequence autoregressively:

\vspace{-1.5em}
\begin{equation}
    \mathcal{L}_{\mathrm{tts}}(\theta)
    = - \sum_{(p,\widetilde{\mathbf{y}}) \in \Dtts}
      \sum_{t=1}^{|\widetilde{\mathbf{y}}|}
      \log P\left(
      \widetilde{y}_t \mid p, \widetilde{\mathbf{y}}_{<t}
      \right).
    \label{eq:tts-loss}
\end{equation}
\vspace{-1.5em}

To stabilize adaptation to the extended generation token space, we adopt a progressive strategy inspired by prior work \cite{TokAlign, TokAlign++}.
In the first stage, we use 50\% of the data to fine-tune only the embedding and LM-head layers while freezing the remaining model parameters.
This calibrates the newly added vocabulary parameters before broader model adaptation.
In the second stage, we use the remaining 50\% of the data and additionally unfreeze the intermediate model layers to complete pronunciation adaptation.

\begin{figure}[t]
    \centering
    \includegraphics[width=\linewidth]{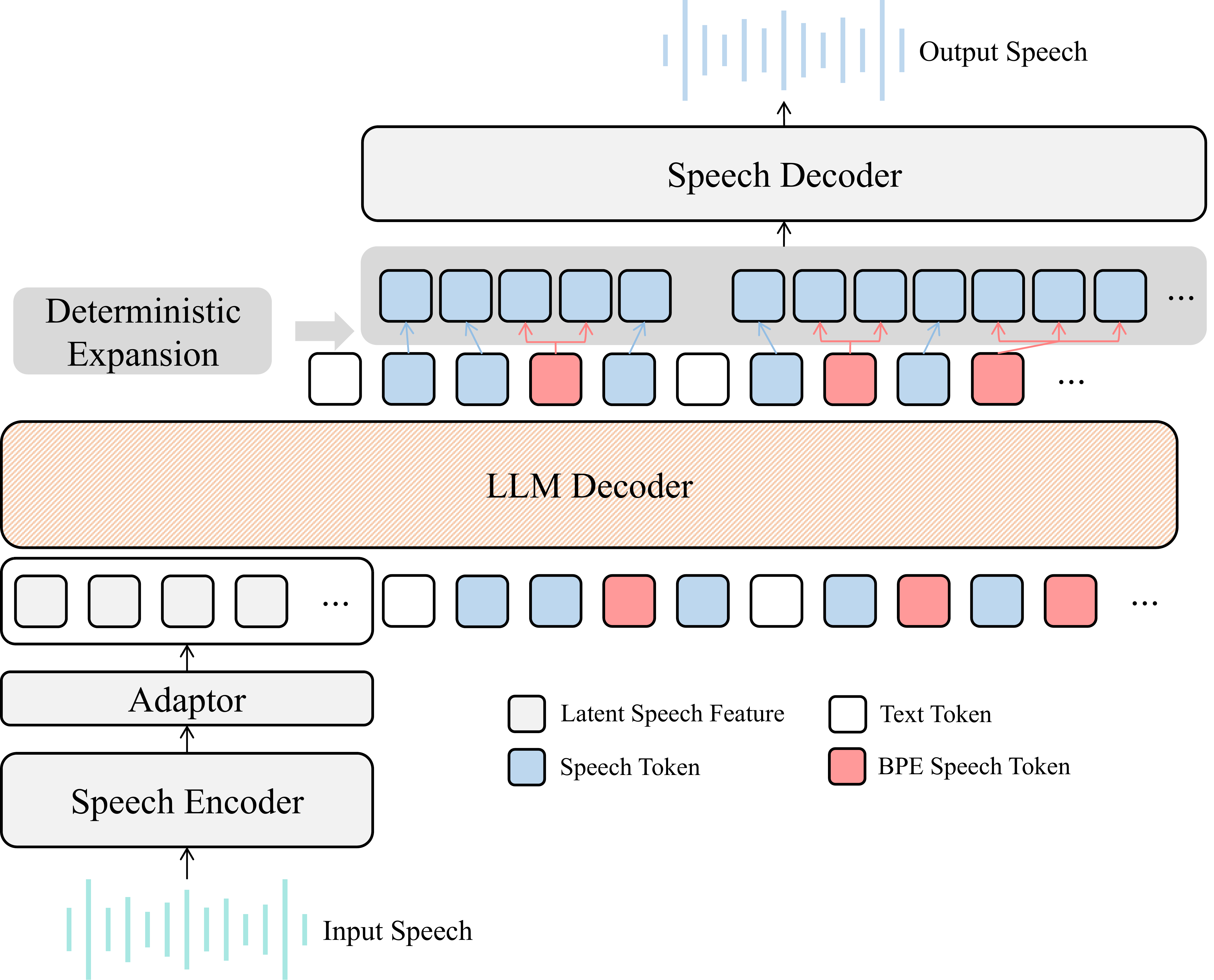}
    \vspace{-1.5em}
    \caption{Inference with deterministic expansion from the extended generation token space to the original space before unchanged decoding.}
    \label{fig:inf}
    \vspace{-1.5em}
\end{figure}

\subsection{Dialogue Adaptation}
\label{subsec:dlg-adapt}

Pronunciation adaptation teaches the model to use the extended generation token space under text prompts, but does not yet adapt this ability to speech-to-speech dialogue.
We therefore perform dialogue adaptation on compressed speech-to-speech dialogue data, updating only the LLM parameters while keeping the speech encoder, adaptor, and speech decoder frozen.
Each dialogue example contains a user's speech query $q$ and a target assistant response $\mathbf{r}$.
As in pronunciation adaptation, only the speech tokens of the interleaved target response are compressed.
Let $\widetilde{\mathbf{r}} = C(\mathbf{r})$.
The objective of dialogue adaptation is:
\begin{equation}
    \mathcal{L}_{\mathrm{dlg}}(\theta)
    = - \sum_{(q,\widetilde{\mathbf{r}}) \in \Ddlg}
      \sum_{t=1}^{|\widetilde{\mathbf{r}}|}
      \log P\left(
      \widetilde{r}_t \mid q, \widetilde{\mathbf{r}}_{<t}
      \right).
    \label{eq:dialogue-loss}
\end{equation}
After this stage, the SDM can use the extended generation token space for speech-to-speech dialogue, with the generated speech-token stream mapped back to $\mathcal{S}$ by $\pi$ before waveform synthesis.

\input{tabs/efficiency}
\input{tabs/system}

\subsection{Inference with Deterministic Expansion}
\label{subsec:inference}

As shown in Figure~\ref{fig:inf}, the adapted SDM generates interleaved text and speech tokens from the extended model vocabulary $\mathcal{V}'$.
The speech-token stream lies in $\mathcal{Z}$ and may contain both original tokens in $\mathcal{S}$ and added BPE units in $\mathcal{Z}\setminus\mathcal{S}$.
We apply the deterministic expansion $\pi$ only to this speech-token stream: original tokens are left unchanged, while each added BPE unit is expanded into its corresponding sequence in $\mathcal{S}$.
The recovered original speech-token sequence is then fed directly into the unchanged speech decoder for waveform synthesis.
This compatibility-preserving mapping allows the SDM to take fewer autoregressive steps while keeping the synthesis interface unchanged.

\section{Experiments}
\label{sec:experiments}

\subsection{Experimental Setup}

\paragraph{Implementation.}
We use the 8B Step-Audio 2 mini~\cite{Step-Audio2} as the SDM for its strong open-source support, simple architecture, and competitive dialogue performance; it uses S3Tokenizer V2 at 25 Hz.
For pronunciation adaptation, we sample 300 hours each of Chinese and English speech from Emilia~\cite{Emilia} and convert them to a unified timbre with Seed-VC~\cite{seedvc}.
For dialogue adaptation, we follow OpenS2S~\cite{OpenS2S} to construct approximately 50K speech-dialogue examples per language, with assistant speech synthesized by CosyVoice 2~\cite{CosyVoice2} using a consistent voice.
We learn the BPE vocabulary from dialogue speech-token sequences and use 80K added units, supported by the vocabulary-size trade-off analysis in Section~\ref{sec:analyses}.
Vocabulary extension adds only embedding and LM-head rows, about 6\% of the original model parameters.
We train CoSTA-ZH, CoSTA-EN, and CoSTA-MIX for Chinese, English, and bilingual adaptation; further training details and data statistics are provided in Appendices~\ref{app:implementation-details} and~\ref{app:train-data}.

\input{tabs/quality}

\paragraph{Baselines and benchmarks.}
We compare against the original Step-Audio 2 mini and Vanilla-
FT-ZH/EN/MIX, which use the same adaptation data and training configuration while retaining the original speech tokens.
This controls for training-data scale and isolates the effect of BPE-compressed token adaptation.
Qwen2.5-Omni~\cite{Qwen2.5-Omni}, Kimi-Audio~\cite{Kimi-Audio}, and GLM-4-Voice~\cite{GLM-4-Voice} serve as external quality references.
We evaluate on 12 datasets from URO-Bench~\cite{uro}, covering aligned Chinese and English settings and spanning summarization, mathematical and common-sense reasoning, open-ended dialogue, and real-world conversation, providing broad coverage of realistic spoken-dialogue scenarios.

\paragraph{Efficiency metrics.}
We report three model-side efficiency metrics.
First, the \textbf{generated-token compression rate} $\rho_g=\rho(\widehat{\mathbf{z}})$ is computed using Eq.~\ref{eq:compression-rate}, where $\widehat{\mathbf{z}}$ denotes the model-generated speech-token sequence; a lower $\rho_g$ indicates fewer autoregressive speech-token steps.
Second, the \textbf{expanded speech-token generation speed} is
\begin{equation}
    \mathrm{Speed}
    = \frac{|\pi(\widehat{\mathbf{z}})|}{T_{\mathrm{decode}}},
    \label{eq:decode-speed}
\end{equation}
where $T_{\mathrm{decode}}$ is LLM decoding time, measuring the effective number of original speech tokens generated per second after expansion.
Third, \textbf{relative speed}, denoted as \textbf{Relative} in the tables, normalizes the average Speed to that of Step-Audio 2 mini under the same language setting.

At the system level, we additionally report mean full-response latency, request throughput, and peak GPU memory on AlpacaEval-zh and AlpacaEval at concurrency 1, 4, and 8 using a single A100 80GB with vLLM.
First-packet latency is reported separately in Appendix~\ref{app:first-packet}.

\paragraph{Quality metrics.}
Following the URO-Bench evaluation protocol, we report \textbf{G-Eval}, \textbf{UTMOS}~\cite{UTMOS}, and \textbf{WER/CER} for response-content quality, speech quality, and text-speech alignment, respectively.
Evaluation details are provided in Appendix~\ref{app:benchmarks}.
We further conduct a MOS-style listening study with six evaluators, yielding 1,080 response-level evaluations of speech fluency and semantic quality; details are provided in Appendix~\ref{app:human-evaluation}.

\subsection{Main Results}

\paragraph{Model-side efficiency.}
Table~\ref{tab:efficiency-results} shows substantial efficiency gains in both languages under the vLLM streaming inference setting.
CoSTA-ZH reaches $\rho_g=66.81\%$ and $1.736\times$ relative speed, while CoSTA-EN reaches $70.17\%$ and $1.659\times$.
CoSTA-MIX shows similar improvements, whereas vanilla fine-tuning retains $\rho_g=100\%$ with nearly unchanged generation speed.
These results show that compressed-token modeling reduces autoregressive speech-token steps and translates this reduction into higher generation efficiency.

\begin{figure}[t]
    \centering
    \includegraphics[width=\linewidth]{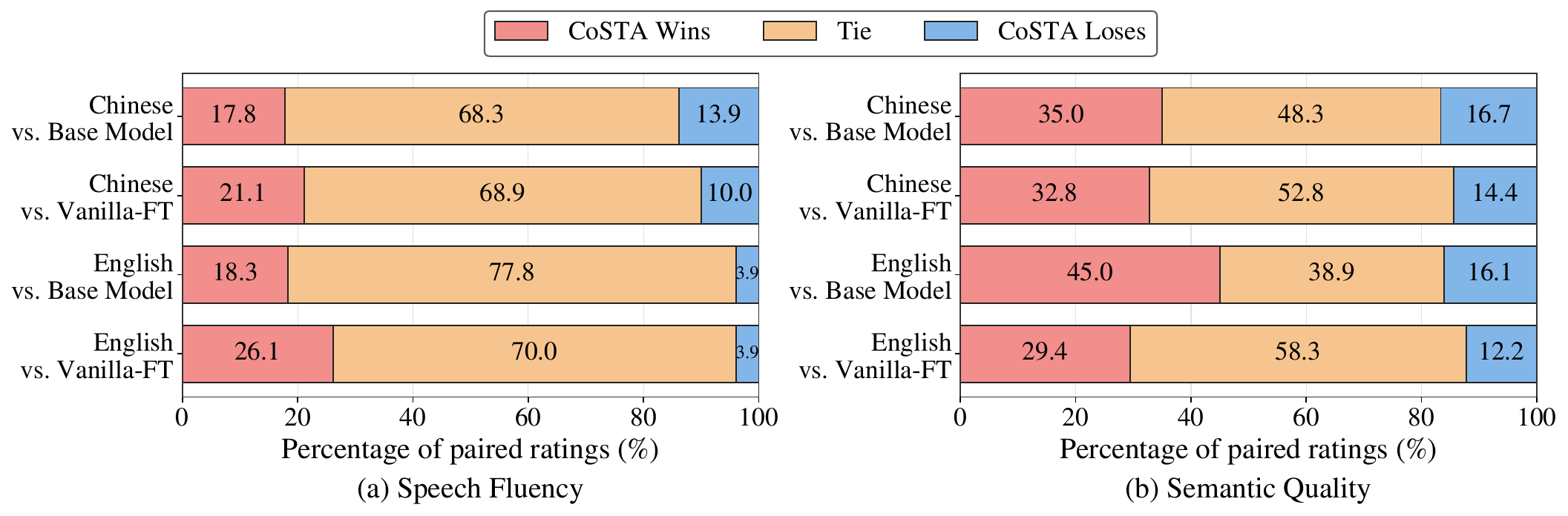}
    \vspace{-1.9em}
    \caption{Paired human evaluation of CoSTA against the base model and Vanilla-FT. Bars show wins, ties, and losses from 180 paired MOS ratings per language and metric.}
    \label{fig:human-evaluation}
    \vspace{-1.9em}
\end{figure}

\paragraph{System efficiency.}
Table~\ref{tab:full-response-efficiency} shows that decoding gains translate into lower full-response latency and higher throughput.
Relative to the base model, CoSTA reduces latency by 32.8--39.0\% in Chinese and 22.4--26.0\% in English across the tested concurrency levels; throughput increases by 51.6--59.2\% and 25.8--38.6\%, respectively.
It also outperforms matched Vanilla-FT at every tested setting, with at most 0.68 GiB additional peak memory.
These workload-level results show that CoSTA's model-side efficiency gains translate into system-level improvements in latency and throughput with minimal memory overhead.

\paragraph{Automatic quality evaluation.}
Table~\ref{tab:main-results-uro} shows that CoSTA preserves or improves average output quality while substantially improving efficiency.
Compared with matched Vanilla-FT, CoSTA improves G-Eval and substantially reduces ASR-CER/WER in both Chinese and English, while UTMOS remains comparable; CoSTA-MIX shows the same overall trend.
These results indicate that compressed-token generation does not trade response quality for efficiency and can improve text-speech alignment, while preserving speech quality after deterministic expansion.
Overall, CoSTA achieves a favorable efficiency--quality trade-off within the adapted SDM.

\paragraph{Human evaluation.}
Figure~\ref{fig:human-evaluation} shows that CoSTA wins more often than it loses against both the base model and Vanilla-FT in Chinese and English.
Win-or-tie rates reach 86.1--96.1\% for speech fluency and 83.3--87.7\% for semantic quality, indicating that efficiency gains preserve perceived quality.
MOS scores show the same trend; detailed results are provided in Appendix~\ref{app:human-evaluation}.

\subsection{Ablation Study}

\input{tabs/ablation_zh}

\begin{figure}[t]
    \centering
    \includegraphics[width=\linewidth]{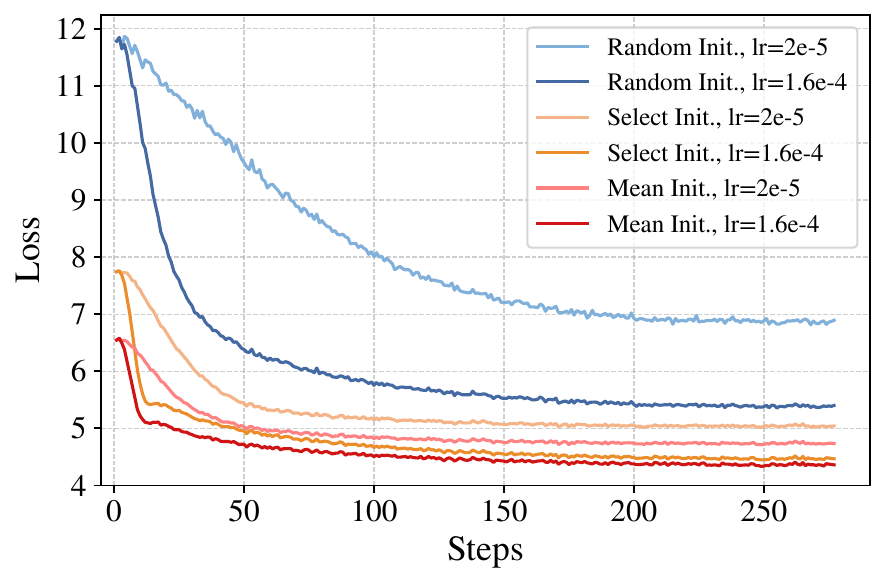}
    \vspace{-2.0em}
    \caption{Training loss during the first stage of pronunciation adaptation. Mean initialization achieves faster early convergence than the alternatives.}
    \label{fig:loss}
    \vspace{-0.6em}
\end{figure}

Table~\ref{tab:ablation-zh} shows that the adaptation components play complementary roles.
Timbre conversion mainly benefits response quality, while progressive adaptation modestly improves compressed-token generation.
Pronunciation adaptation further strengthens compression and text-speech alignment, supporting reliable generation in the extended token space.
Dialogue adaptation has the largest effect: removing it causes G-Eval to drop sharply from 72.27 to 24.35, indicating that learning compressed pronunciation alone is insufficient to recover spoken-dialogue capability.
Appendix~\ref{app:dialogue-adaptation-analysis} further analyzes the associated compressed-token behavior and response failures.
Figures~\ref{fig:loss} and~\ref{fig:grad_norm} further show that mean initialization provides faster convergence and more stable gradients than the alternatives; detailed initialization strategies are described in Appendix~\ref{app:init-analysis}.

\begin{figure*}[t]
    \centering
    \includegraphics[width=0.49\textwidth]{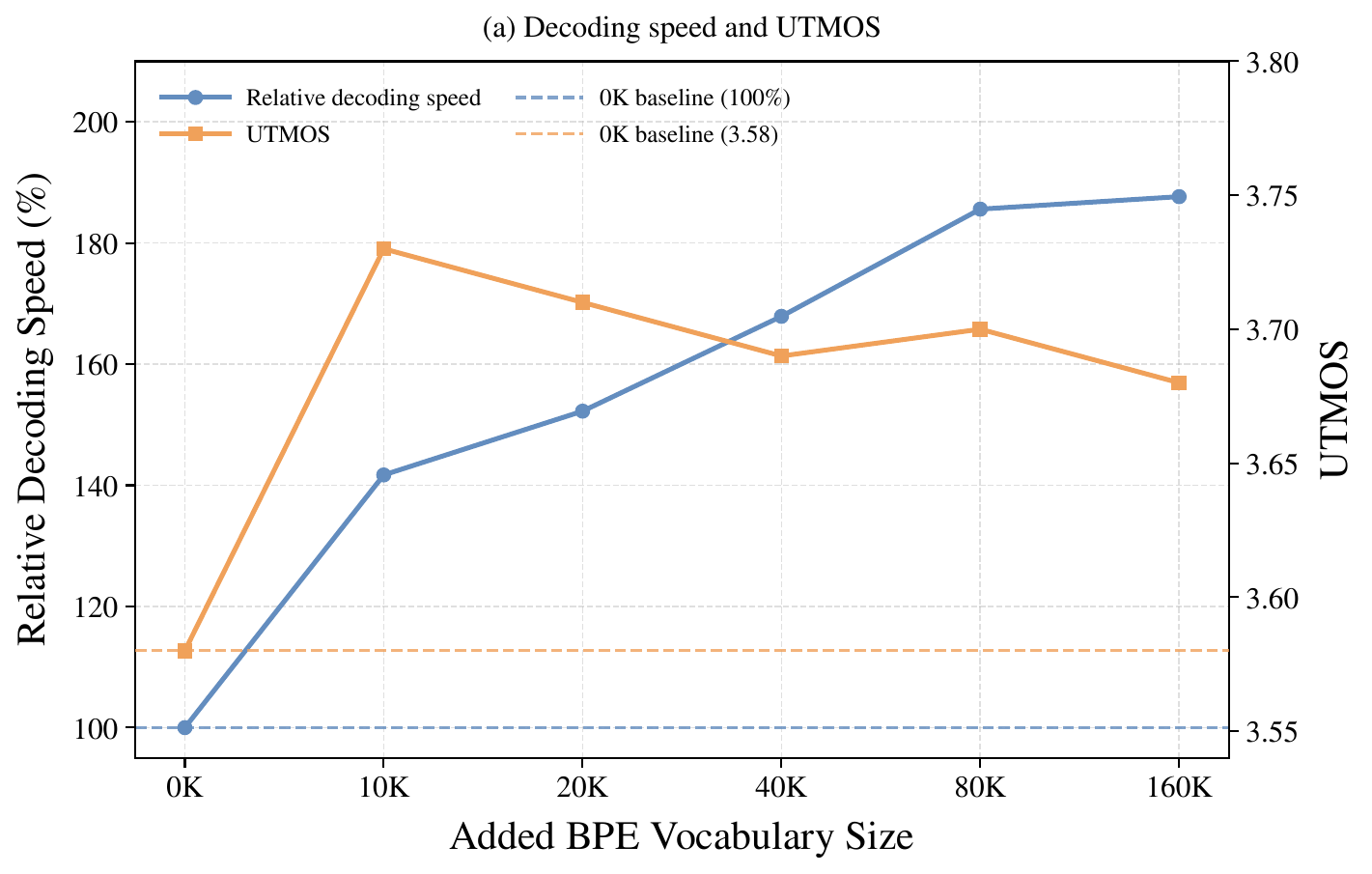}\hfill
    \includegraphics[width=0.49\textwidth]{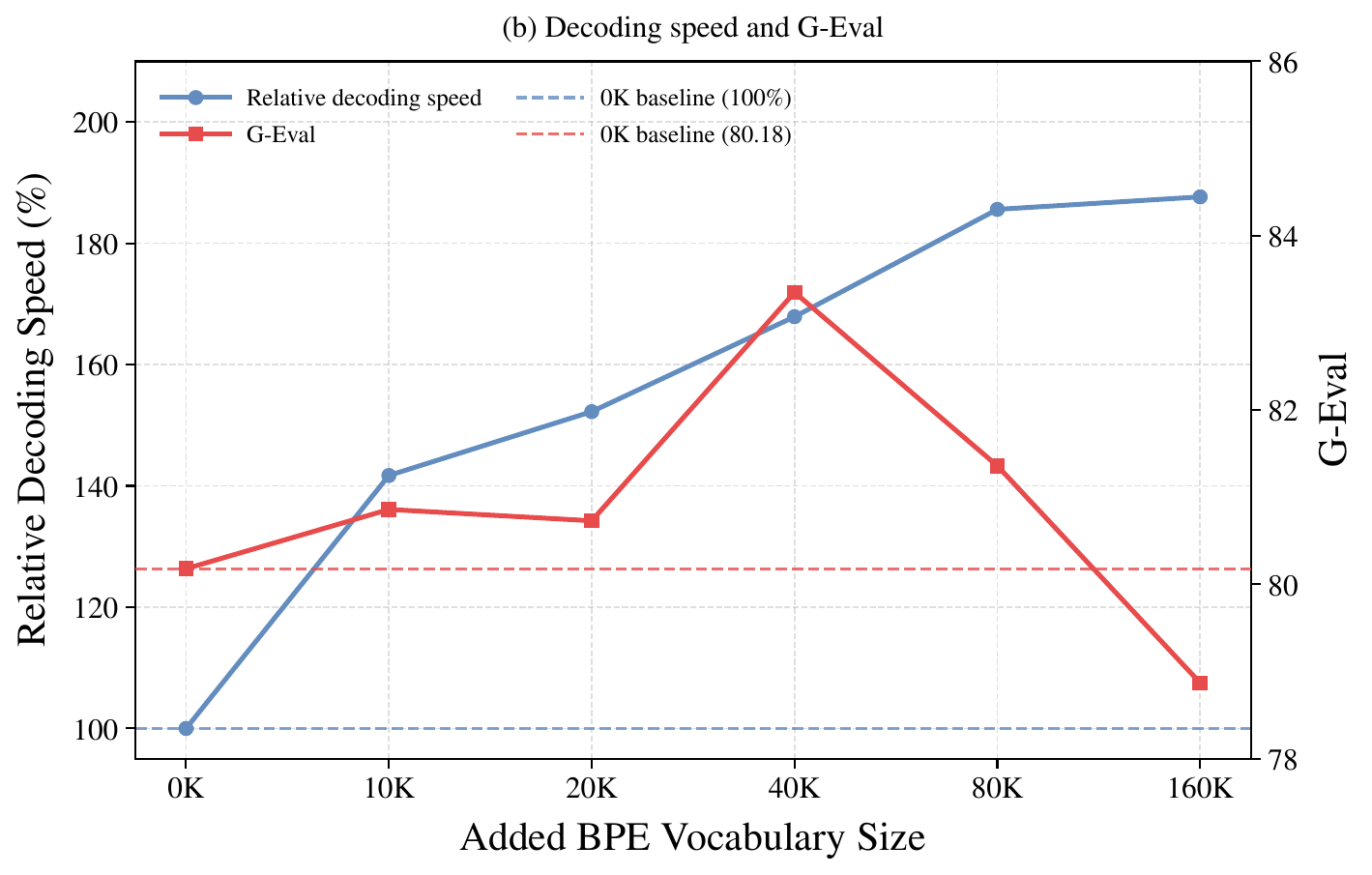}
    \vspace{-1.0em}
    \caption{Vocabulary-size trade-off on AlpacaEval-zh: relative expanded speech-token generation speed with (a) UTMOS and (b) G-Eval. The 0K baseline denotes the original Step-Audio 2 mini; other points denote CoSTA-ZH.}
    \label{fig:vocab-tradeoff}
    \vspace{-1.2em}
\end{figure*}

\subsection{Analyses}
\label{sec:analyses}

\paragraph{Interaction under limited budgets.}
Table~\ref{tab:multi-turn-duration} groups every 10 queries from AlpacaEval-zh and AlpacaEval into a stress-test session, yielding 14 Chinese and 19 English sessions, respectively.
CoSTA sustains 7.21/5.79 effective turns in Chinese/English, compared with 3.71/1.84 for Vanilla-FT, while also producing longer speech within these interactions.
This suggests that compressed-token generation uses the available autoregressive generation budget more efficiently, sustaining more turns and longer speech.
This setting evaluates generation-format sustainability rather than conversational coherence.

\input{tabs/multi_turn_duration}

\paragraph{Vocabulary-size trade-off.}
Figure~\ref{fig:vocab-tradeoff} shows that increasing the added vocabulary from 10K to 80K substantially improves compression and decoding speed while maintaining response quality.
Extending further to 160K reduces $\rho_g$ from 67.50\% to 63.62\%, but yields only a 1.1\% speed gain, while G-Eval decreases from 81.36 to 78.87.
These diminishing full-model gains motivate 80K as a practical balance between compression, decoding efficiency, response-content quality, and speech quality in our setting.
Complete vocabulary-size results are provided in Appendix~\ref{app:full-model-vocab}, with additional analysis of output-layer overhead in Appendix~\ref{app:lm-head-cost}.

\paragraph{Transfer across SDM architectures.}
Table~\ref{tab:opens2s-generality} evaluates CoSTA on OpenS2S~\cite{OpenS2S}, whose Think--Talk architecture, 12.5-Hz GLM-4-Voice tokenizer, and speech decoder differ from Step-Audio 2 mini.
CoSTA reduces speech-token steps by 21.17\% and 18.64\% in Chinese and English, respectively, while improving response-content quality and text-speech alignment over matched Vanilla-FT and maintaining comparable UTMOS.
These results show that post-hoc token-space adaptation transfers to a different SDM architecture in both language settings.
Moreover, the additional compression on top of a native 12.5-Hz tokenizer suggests that CoSTA is complementary to low-rate speech tokenization and can be combined with it for further sequence reduction.
Detailed results are provided in Table~\ref{tab:opens2s-generality-detail}.

Additional data-scaling analyses and examples are provided in Appendices~\ref{app:data-scaling} and~\ref{app:case-study}.

\input{tabs/sdm_gen}

\vspace{-0.8em}
\section{Related Work}
\label{sec:related}
\vspace{-0.8em}

\paragraph{Spoken dialogue models.}

Existing SDMs differ in how text and speech generation are coupled. Think--Talk architectures use separate Transformer backbones~\cite{LLaMA-Omni,Freeze-Omni,LLaMA-Omni2,OpenS2S,Qwen2.5-Omni,Qwen3-Omni,Qwen3.5-Omni,MiniCPM-o4.5,Moshi,MiMo-Audio}; parallel-generation architectures use multiple output heads~\cite{Kimi-Audio,SLAM-Omni,Mini-Omni,Mini-Omni2,LongCat-Flash-Omni}; and interleaved-generation architectures model interleaved text--speech sequences~\cite{SpeechGPT,Baichuan-Audio,GLM-4-Voice,Step-Audio2}.
Despite these differences, the generation token space is typically coupled to the space expected by the speech decoder.
\method instead decouples generation and synthesis through compatibility-preserving post-hoc adaptation, enabling compact, losslessly expandable speech-token generation while retaining the original speech tokenizer and decoder.

\paragraph{Discrete speech tokens.}

Discrete speech tokens include acoustic tokens from codec models~\cite{SoundStream, EnCodec, DAC, WavTokenizer, SpeechTokenizer} and semantic tokens from discriminative~\cite{CosyVoice, CosyVoice2, CosyVoice3} or self-supervised speech models~\cite{vq-wav2vec, wav2vec2.0, HuBERT, WavLM, w2v-BERT}.
Lower-rate tokenizers shorten speech-token sequences~\cite{Moshi, Zeng, SoCodec, TaDiCodec, SpeechTokensReview}, but aggressive rate reduction may introduce compression--quality trade-offs, while tokenizer replacement may require retraining or adapting the tokenizer, speech decoder, and SDM together.
Moreover, this approach still couples generation and synthesis to the same token space.
Acoustic-piece and BPE methods have been explored in ASR~\cite{renshuo, asrbpe, asrbpe2}, speech continuation~\cite{aBPE}, and TTS~\cite{Vec-Tok, aBPE2}, primarily through native training of small, task-specific models.
In contrast, \method enables post-hoc adaptation of existing large-scale SDMs for efficient compressed-token generation.
We further uncover a consistent scaling regularity of speech-token compression and characterize the vocabulary-size trade-off.

\vspace{-0.8em}
\section{Conclusion}
\vspace{-0.8em}

We uncover an approximate log-linear scaling regularity in speech-token BPE and introduce \method, which decouples generation and synthesis token spaces while retaining the original tokenizer and decoder.
\method achieves approximately $1.7\times$ expanded speech-token generation speed and improves system efficiency while preserving response quality.
It also supports more sustained interaction, reveals vocabulary-size trade-offs, and transfers across SDM architectures, complementing low-rate tokenization.

\section*{Limitations}

This work has several limitations and possible extensions.
First, our dialogue-adaptation data is constructed from synthesized speech-to-speech examples. While such data provides controllable and consistent speech responses for studying speech-token compression, it is also a practical and widely adopted choice for training spoken dialogue models, as large-scale, high-quality real speech-to-speech dialogue data is difficult to collect. Future work may further explore how synthetic and real conversational speech data complement each other.
Second, we adopt BPE as a deterministic, lossless, and easily integrated compression method. Future work may investigate other lossless speech-token segmentation strategies and compare their behavior in spoken dialogue modeling.
Finally, our evaluation focuses on general speech-to-speech dialogue and efficiency. Future work may extend the analysis to finer-grained scenarios, such as emotional dialogue, personalized speaking styles, and diverse acoustic environments.

\bibliography{custom}

\clearpage
\appendix

\section{Implementation Details and Hyperparameters}
\label{app:implementation-details}

This section reports the implementation details and hyperparameters used in \method. Unless otherwise specified, all models are initialized from Step-Audio 2 mini \cite{Step-Audio2}. The original speech tokenizer and speech decoder are kept unchanged throughout all experiments. The original speech-token vocabulary contains 6,656 tokens. We train the BPE tokenizer using the HuggingFace \texttt{tokenizers} library\footnote{\url{https://github.com/huggingface/tokenizers}} on speech-token sequences extracted from the general speech-to-speech dialogue corpus, and extend the original vocabulary with 80K additional BPE speech tokens. Since the BPE trainer does not directly operate on integer-valued speech-token sequences, we map each original speech-token ID to a unique Unicode character before BPE training. During BPE expansion, generated BPE-token IDs are first converted into their corresponding Unicode strings, and BPE expansion is performed in the Unicode space; the resulting BPE-expanded Unicode sequence is then mapped back to the corresponding original speech-token IDs. This conversion is deterministic and does not change the underlying speech-token sequence. All adaptation and inference experiments are conducted on NVIDIA A100 80GB GPUs. For all adaptation stages, we use the AdamW optimizer~\cite{AdamW}, train with bfloat16 mixed precision~\cite{MixedPrecisionTraining,Kalamkar2019BF16}, and use DeepSpeed~\cite{DeepSpeed} with ZeRO-1 optimizer-state partitioning~\cite{ZeRO} for distributed training.

\subsection{Pronunciation Adaptation}

Pronunciation adaptation is conducted on compressed general TTS data. Following the progressive adaptation strategy, we split the pronunciation-adaptation data into two halves. In the first stage, only the embedding and LM head layers of the vocabulary-extended model are updated. We use a relatively large learning rate of $1.6\times10^{-4}$ in this stage, which leads to faster loss reduction and more stable optimization as shown in Figures~\ref{fig:loss} and~\ref{fig:grad_norm}. In the second stage, intermediate model layers are further unfrozen and trained with a smaller learning rate of $2.0\times10^{-5}$ to complete pronunciation adaptation. The hyperparameters are shown in Table~\ref{tab:app-pronunciation-hparams}.

\begin{table}[t]
\centering
\small
\renewcommand\arraystretch{1.05}
\setlength{\tabcolsep}{3mm}
\begin{tabular}{l !{\vrule} c}
\toprule
\textbf{Hyperparameter} & \textbf{Value} \\
\midrule
Data split & 50\% / 50\% \\
Stage-1 trainable params & Embedding, LM head \\
Stage-2 trainable params & + Intermediate layers \\
Optimizer & AdamW \\
Stage-1 learning rate & $1.6\times10^{-4}$ \\
Stage-2 learning rate & $2.0\times10^{-5}$ \\
Scheduler & Cosine \\
Warmup ratio & 0.03 \\
Global batch size & 256 \\
Precision & bfloat16 \\
Gradient clipping norm & 1.0 \\
Epochs & 3 \\
GPUs & 8 \\
\bottomrule
\end{tabular}
\caption{Hyperparameters used during pronunciation adaptation.}
\label{tab:app-pronunciation-hparams}
\end{table}

\subsection{Dialogue Adaptation}

After pronunciation adaptation, we perform dialogue adaptation on compressed general speech-to-speech dialogue data. The target responses contain interleaved text and speech tokens, where only speech tokens are BPE-compressed and text tokens remain unchanged. The compressed speech tokens are arranged in the response sequence according to the original interleaving ratio between text tokens and speech tokens used by the base SDM. The hyperparameters are shown in Table~\ref{tab:app-dialogue-hparams}.

\begin{table}[t]
\centering
\small
\renewcommand\arraystretch{1.05}
\setlength{\tabcolsep}{3mm}
\begin{tabular}{l !{\vrule} c}
\toprule
\textbf{Hyperparameter} & \textbf{Value} \\
\midrule
Trainable params & All LLM parameters \\
Optimizer & AdamW \\
Learning rate & $2.0\times10^{-5}$ \\
Scheduler & Cosine \\
Warmup ratio & 0.03 \\
Global batch size & 256 \\
Precision & bfloat16 \\
Gradient clipping norm & 1.0 \\
Epochs & 3 \\
GPUs & 8 \\
\bottomrule
\end{tabular}
\caption{Hyperparameters used during dialogue adaptation.}
\label{tab:app-dialogue-hparams}
\end{table}

\subsection{Inference}

During inference, the adapted SDM generates an interleaved sequence of text tokens and speech tokens from the extended vocabulary. The generated speech tokens may include both original speech tokens and newly added BPE speech tokens. We apply BPE expansion only to the generated speech-token stream: original speech tokens are kept unchanged, while each merged BPE token is deterministically expanded into its original speech-token sequence. The expanded speech tokens are then fed into the unchanged speech decoder for waveform synthesis. The inference hyperparameters are shown in Table~\ref{tab:app-inference-hparams}.

\begin{table}[t]
\centering
\small
\renewcommand\arraystretch{1.05}
\setlength{\tabcolsep}{3mm}
\begin{tabular}{l !{\vrule} c}
\toprule
\textbf{Hyperparameter} & \textbf{Value} \\
\midrule
Decoding mode & Greedy \\
Repetition penalty & 1.2 \\
Max new tokens & 2048 \\
BPE expansion & Enabled \\
Expansion target & Speech-token stream only \\
Text-token stream & Unchanged \\
Speech decoder & Original decoder \\
Hardware & 1 NVIDIA A100 GPU \\
\bottomrule
\end{tabular}
\caption{Hyperparameters used during inference.}
\label{tab:app-inference-hparams}
\end{table}

\subsection{Adaptation Cost}
\label{app:adaptation-cost}

For CoSTA-ZH on eight A100 80GB GPUs, the two pronunciation-adaptation stages take 1:29:41 and 4:34:00, and dialogue adaptation takes 3:52:54. The total is 9:56:35 of training wall-clock time, approximately 79.5 A100 GPU-hours. Data preparation is additional: timbre conversion and dialogue synthesis each take approximately 1--2 days using eight A100 GPUs. These one-time costs should be considered alongside the inference savings; they are not included in the training-only total.

\section{Training Data and Benchmarks}
\label{app:data-and-benchmarks}

This section provides additional details about the training data and evaluation benchmarks used in our experiments.

\subsection{Training Data}
\label{app:train-data}

\paragraph{Pronunciation-adaptation data.}
For pronunciation adaptation, we use Emilia \cite{Emilia}, a large-scale multilingual speech generation dataset constructed from in-the-wild speech. Emilia contains diverse speech data across multiple languages and speaking styles, making it suitable for training speech generation models. In this work, we sample 300 hours of Chinese speech and 300 hours of English speech from Emilia. To reduce speaker variation and improve speech-token pattern consistency, we convert the sampled utterances into a unified timbre using Seed-VC \cite{seedvc}. The converted speech is then tokenized by the original speech tokenizer of Step-Audio 2 mini \cite{Step-Audio2}, and the resulting speech tokens are compressed with the learned BPE tokenizer. Text tokens remain unchanged. The compressed speech tokens are arranged according to the original text-speech interleaving pattern used by the base SDM.

\paragraph{Dialogue-adaptation data.}
For dialogue adaptation, we follow the general-purpose bilingual speech-to-speech instruction-pair construction pipeline of OpenS2S~\cite{OpenS2S}. Specifically, we first collect general text instructions from InstructS2S-200K~\cite{LLaMA-Omni} and translate the English instructions into Chinese using Qwen3-32B~\cite{qwen3}, yielding bilingual text instructions. We then select seed audio with neutral emotion labels and convert these text instructions into user speech queries. Following the speech response generation process in OpenS2S, Qwen3-32B is used to generate dialogue-appropriate text responses, and CosyVoice 2~\cite{CosyVoice2} synthesizes the corresponding assistant speech with a consistent reference voice. Following this pipeline, we initially construct 50K Chinese and 50K English general speech-to-speech instruction examples, each containing a user's speech query and an assistant response. We further filter out examples whose query or response speech exceeds one minute, resulting in 49,664 Chinese and 49,408 English dialogue examples. The statistics of the training data are summarized in Table~\ref{tab:app-training-data}. During training, only speech tokens in the target response are BPE-compressed, while text tokens remain unchanged. The compressed speech tokens are arranged according to the original text-speech interleaving pattern used by the base SDM.

\begin{table}[t]
\centering
\small
\renewcommand\arraystretch{1.1}
\setlength{\tabcolsep}{3mm}
\begin{tabular}{l c c}
\toprule
\textbf{Data Type} & \textbf{Language} & \textbf{Scale} \\
\midrule
\multirow{2}{*}{Text-to-speech}
& Chinese & 300h \\
& English & 300h \\
\midrule
\multirow{2}{*}{Speech dialogue}
& Chinese & 49,664 examples\\
& English & 49,408 examples\\
\bottomrule
\end{tabular}
\caption{Statistics of training data.}
\label{tab:app-training-data}
\end{table}

\begin{table}[t]
\centering
\scriptsize
\renewcommand\arraystretch{1.05}
\setlength{\tabcolsep}{0.8mm}
\resizebox{\linewidth}{!}{
\begin{tabular}{l c c c}
\toprule
\textbf{Dataset} & \textbf{\#Samples} & \textbf{Task / Aspect} & \textbf{Eval. Mode} \\
\midrule

AlpacaEval-zh & 147 & Authentic, open-ended dialogue & Open \\
AlpacaEval & 199 & Authentic, open-ended dialogue & Open \\
APE-zh & 190 & Practical mathematical problems & QA \\
Gsm8kEval & 582 & Practical mathematical problems & QA \\
Claude-zh & 222 & Authentic, open-ended dialogue & Open \\
CommonEval & 200 & Authentic, open-ended dialogue & Open \\
LCSTS-zh & 119 & Summarization & Semi-open \\
Summary & 118 & Summarization & Semi-open \\
MLC-zh & 145 & Math, logic, and common sense & QA \\
MLC & 177 & Math, logic, and common sense & QA \\
Wildchat-zh & 299 & Real-world conversation & Open \\
WildchatEval & 349 & Real-world conversation & Open \\

\bottomrule
\end{tabular}
}
\caption{URO-Bench datasets used in our evaluation. The selected datasets cover understanding, reasoning, and oral conversation in Chinese and English speech-to-speech settings.}
\label{tab:app-urobench-datasets}
\end{table}

\subsection{Evaluation Benchmarks}
\label{app:benchmarks}

We evaluate models on URO-Bench \cite{uro}, a comprehensive benchmark for end-to-end spoken dialogue models in speech-to-speech scenarios. URO-Bench covers three major evaluation dimensions: Understanding, Reasoning, and Oral Conversation, and assesses not only response-content quality but also speech quality and speech-text alignment. In our experiments, we use 12 aligned Chinese and English datasets from URO-Bench, as summarized in Table~\ref{tab:app-urobench-datasets}. These benchmarks better reflect realistic spoken dialogue scenarios, making them well aligned with our goal of improving the efficiency and usability of speech-to-speech SDMs.

Following the URO-Bench evaluation protocol, we evaluate output quality using its prompts and scoring procedures. Specifically, we report the following metrics: (1) \textbf{G-Eval} measures response-content quality by transcribing the generated speech with an ASR model and prompting GPT-4o mini \cite{GPT-4o-mini} to score the transcription. We use Whisper-large-v3 \cite{asr_en} for English transcription and Paraformer-zh \cite{asr_zh} for Chinese transcription. (2) \textbf{UTMOS} \cite{UTMOS} estimates speech quality with a predicted mean opinion score that reflects clarity, naturalness, and fluency. (3) \textbf{WER/CER} evaluates the alignment between the speech and text outputs of SDMs by computing the Word Error Rate (WER) or Character Error Rate (CER) between the generated text and the transcription of the generated speech.

URO-Bench uses different evaluation prompts according to the task format. For open-ended tasks, including AlpacaEval-zh, AlpacaEval, Claude-zh, CommonEval, Wildchat-zh, and WildchatEval, we use \texttt{prompt\_open}. For semi-open tasks with reference answers, including LCSTS-zh and Summary, we use \texttt{prompt\_semi\_open}. For question-answering tasks with deterministic references, including APE-zh, Gsm8kEval, MLC-zh, and MLC, we use \texttt{prompt\_qa}. These prompts are fed into GPT-4o mini together with the user instruction, the model response, and reference answer when available, to obtain the final G-Eval score. The complete prompts are provided below.

\begin{tcolorbox}[
    title=\texttt{prompt\_open},
    label={box:prompt-open},
    colback=gray!5,
    colframe=black!60,
    fonttitle=\bfseries,
    boxrule=0.5pt,
    arc=2pt,
    left=4pt,
    right=4pt,
    top=4pt,
    bottom=4pt,
    breakable
]
\normalsize
I need your help to evaluate the performance of several models in the speech interaction scenario. The models will receive a speech input from the user, which they need to understand and respond to with a speech output.

Your task is to rate the model's responses based on the provided user input transcription [Instruction] and the model's output transcription [Response].

Please evaluate the response on a scale of 1 to 5:

1 point: The response is largely irrelevant, incorrect, or fails to address the user's query. It may be off-topic or provide incorrect information.

2 points: The response is somewhat relevant but lacks accuracy or completeness. It may only partially answer the user's question or include extraneous information.

3 points: The response is relevant and mostly accurate, but it may lack conciseness or include unnecessary details that do not contribute to the main point.

4 points: The response is relevant, accurate, and concise, providing a clear answer to the user's question without unnecessary elaboration.

5 points: The response is exceptionally relevant, accurate, and to the point. It directly addresses the user's query in a highly effective and efficient manner, providing exactly the information needed.

Below are the transcription of user's instruction and model's response:

\#\#\# [Instruction]

\{question\}

\#\#\# [Response]

\{answer\}

After evaluating, please output the score only without anything else. You do not need to provide any explanations.
\end{tcolorbox}

\begin{tcolorbox}[
    title=\texttt{prompt\_semi\_open},
    label={box:prompt-semi-open},
    colback=gray!5,
    colframe=black!60,
    fonttitle=\bfseries,
    boxrule=0.5pt,
    arc=2pt,
    left=4pt,
    right=4pt,
    top=4pt,
    bottom=4pt,
    breakable
]
\normalsize
I need your help to evaluate the performance of several models in the speech interaction scenario. The models will receive a speech input from the user, which they need to understand and respond to with a speech output.

Your task is to rate the model's responses based on the provided user input transcription [Instruction], the model's output transcription [Response] and some suggested answers [Reference]. The model's response does not necessarily have to be identical to the suggested answers, as long as it aligns with the question and is reasonable.

Please evaluate the response on a scale of 1 to 5:

1 point: The response is largely irrelevant, incorrect, or fails to address the user's query. It may be off-topic or provide incorrect information. The response does not align with the question in any meaningful way.

2 points: The response is somewhat relevant but lacks accuracy, completeness, or coherence. It may partially address the query but introduces unnecessary information or deviates from the core issue. The response may not align well with the suggested answer but still provides some value.

3 points: The response is relevant and mostly accurate, but may lack conciseness or clarity. It addresses the question reasonably, but there might be slight deviations in approach or content. While it may not strictly align with the suggested answer, it still effectively addresses the core of the query.

4 points: The response is relevant, accurate, and concise. It provides a clear answer to the user's question and avoids unnecessary details. While it may not exactly mirror the suggested answer, it effectively addresses the user's query in a logical and well-reasoned manner.

5 points: The response is exceptionally relevant, accurate, and concise. It directly addresses the user's query in the most efficient manner, providing exactly the information needed. The response may differ from the suggested answer in phrasing or approach but still aligns perfectly with the intent of the query, demonstrating a high level of reasoning and clarity.

Below are the transcription of user's instruction, model's response and the reference answer:

\#\#\# [Instruction]

\{question\}

\#\#\# [Response]

\{answer\}

\#\#\# [Reference]

\{reference\}

After evaluating, please output the score only without anything else. You do not need to provide any explanations.
\end{tcolorbox}

\begin{tcolorbox}[
    title=\texttt{prompt\_qa},
    label={box:prompt-qa},
    colback=gray!5,
    colframe=black!60,
    fonttitle=\bfseries,
    boxrule=0.5pt,
    arc=2pt,
    left=4pt,
    right=4pt,
    top=4pt,
    bottom=4pt,
    breakable
]
\normalsize
I need your help to evaluate the performance of several models in the speech interaction scenario. The models will receive a speech input from the user, which they need to understand and respond to with a speech output.

Your task is to rate the model's responses based on the provided user input transcription [Question], the model's output transcription [Response] and the correct answer [Reference].

Below are the transcription of user's instruction, model's response and the reference answer:

\#\#\# [Question]

\{question\}

\#\#\# [Response]

\{answer\}

\#\#\# [Reference]

\{reference\}

Is the model's response correct based on the question and reference answer?

Please only output a single ``Yes'' or ``No''. Do not output anything else.
\end{tcolorbox}

\section{Additional Results}
\label{app:additional-results}

This section provides detailed quality results, initialization and data-scaling analyses, compression across corpus settings, first-packet latency, human evaluation, OpenS2S transfer, and vocabulary-size and adaptation analyses.

\subsection{Detailed Main Results}
\label{app:detailed-main-results}

In the main paper, we report per-dataset G-Eval scores and average ASR-CER/WER and UTMOS results for Chinese and English URO-Bench datasets. To provide a more complete comparison, Tables~\ref{tab:main-results-zh-full} and~\ref{tab:main-results-en-full} further report the detailed values of all metrics on each dataset, including G-Eval, ASR-CER/WER, and UTMOS. These results detail the average quality trends discussed in the main paper and the variation across tasks; improvements are not uniform for every metric and dataset.

\subsection{Initialization Strategy Analysis}
\label{app:init-analysis}

\paragraph{Initialization strategies.}
We compare three strategies for initializing the embedding and LM-head rows of newly added BPE units.
\textbf{Random initialization} initializes the new rows randomly, without using their constituent original speech tokens.
\textbf{Select initialization} copies the embedding and LM-head row of one randomly selected token from the deterministic original-token expansion of each BPE unit.
\textbf{Mean initialization}, used by CoSTA, averages the corresponding rows of all constituent tokens as defined in Eq.~\ref{eq:mean-init}.

\begin{figure}[t]
    \centering
    \includegraphics[width=\linewidth]{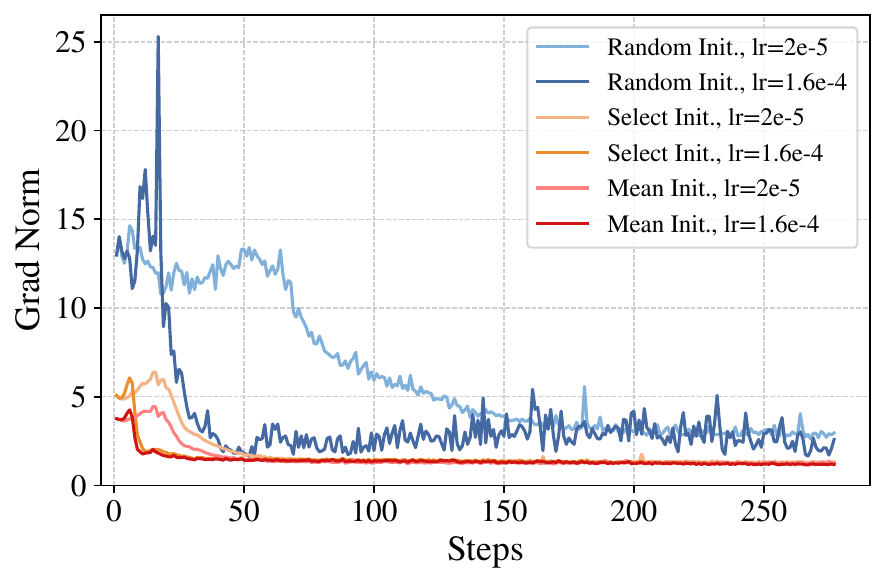}
    \caption{Gradient norm during the first stage of pronunciation adaptation. Mean initialization yields more stable optimization than the alternatives.}
    \label{fig:grad_norm}
\end{figure}

We evaluate these strategies with learning rates of $2\times10^{-5}$ and $1.6\times10^{-4}$ during the first phase of pronunciation adaptation.
Figures~\ref{fig:loss} and~\ref{fig:grad_norm} show that mean initialization converges faster and maintains more stable gradients than the alternatives, particularly at the larger learning rate.
This suggests that aggregating the constituent original-token representations provides a better starting point for adapting the newly added BPE units.

\subsection{Data Scaling Analysis}
\label{app:data-scaling}

We further analyze how the amount of adaptation data affects \method. These scaling experiments were conducted in an early setting without progressive adaptation, and were mainly evaluated on four Chinese URO-Bench datasets: AlpacaEval-zh, APE-zh, Claude-zh, and MLC-zh. We study two scaling settings: varying the amount of pronunciation-adaptation data while keeping the dialogue-adaptation data fixed, and varying the amount of dialogue-adaptation data while keeping the pronunciation-adaptation data fixed.

\paragraph{Scaling pronunciation-adaptation data.}
Table~\ref{tab:tts-data-scale} reports the results when varying the amount of TTS data used for pronunciation adaptation, while fixing the dialogue-adaptation data to 50K examples. Increasing the amount of TTS data generally improves compressed-token learning, especially in terms of compression rate and speech-text alignment. However, the improvement becomes less pronounced as the TTS data size continues to increase. This suggests that pronunciation adaptation benefits from sufficient TTS data, while the marginal gain becomes smaller beyond a certain scale. Based on this trade-off, we use 300h TTS data as the default setting, which provides reliable compressed speech-token generation while keeping the adaptation cost moderate.

\paragraph{Scaling dialogue-adaptation data.}
Table~\ref{tab:dialogue-data-scale} reports the results when varying the amount of speech-to-speech dialogue data, while fixing the pronunciation-adaptation data to 300h. The results show that using 50K dialogue examples provides a strong balance between compression behavior and response quality. Therefore, we adopt 300h TTS data and 50K dialogue examples as the experimental configuration.

\subsection{Compression Across Data Sources and Languages}
\label{app:compression-scaling}

\begin{figure}[t]
    \centering
    \includegraphics[width=0.95\linewidth]{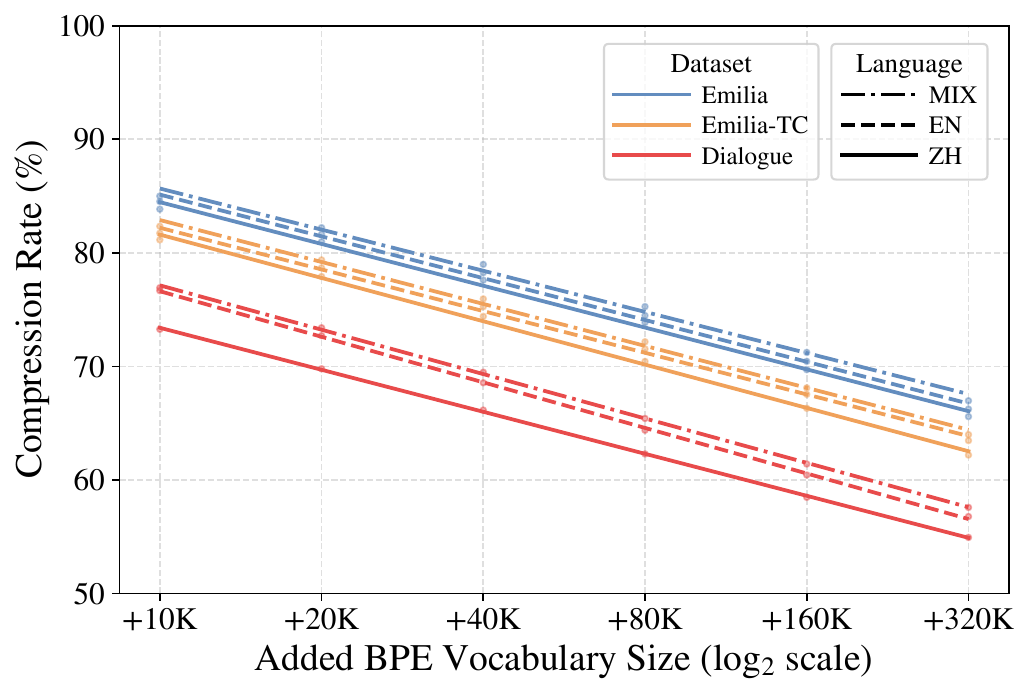}
    \caption{Held-out compression across data sources and languages with S3Tokenizer V2. Points are test measurements; lines are fits against log added vocabulary size. Emilia-TC denotes timbre-converted Emilia.}
    \label{fig:compression_rate}
\end{figure}

Figure~\ref{fig:compression_rate} evaluates S3Tokenizer V2 on test splits of Emilia, timbre-converted Emilia (Emilia-TC), and synthesized dialogue speech, in Chinese, English, and mixed-language settings. Each BPE vocabulary is learned on the corresponding training split and fixed at test time. All nine curves retain the approximate log-linear trend over 10K--320K, with $R^2=0.9942$--$0.9997$ and RMSE $0.10$--$0.47$ pp. Emilia-TC and dialogue speech remain more compressible than original Emilia in these measurements. This demonstrates the trend across settings, rather than transfer of one vocabulary between them. Appendix~\ref{app:compression-diagnostics} gives protocols, complete fits, and training comparisons.

\subsection{First-Packet Latency}
\label{app:first-packet}

Table~\ref{tab:first-packet-latency-breakdown} separates LLM prefill, decoding to the first audio packet, and speech-decoder latency.
CoSTA reduces the LLM decode component from 164.42 to 122.98 ms in Chinese and from 165.08 to 115.96 ms in English, corresponding to reductions of 25.2\% and 29.8\%.
Prefill and speech-decoder latency remain largely unchanged.
The resulting first-packet latency reductions are 9.4\% and 9.6\%, reflecting that CoSTA primarily accelerates autoregressive speech-token decoding while retaining the original speech decoder; over longer responses, these generation-stage savings accumulate, leading to larger full-response latency benefits.

\subsection{Human Evaluation Protocol and Detailed Results}
\label{app:human-evaluation}

We randomly sample five queries from each of the 12 URO-Bench datasets, for 60 queries in total. Six evaluators rate the responses of Step-Audio 2 mini, matched Vanilla-FT, and CoSTA on a 1--5 scale for speech fluency and semantic quality. This yields $60\times3\times6=1{,}080$ response-level evaluations, each with two scores. Tables~\ref{tab:human-evaluation-zh} and~\ref{tab:human-evaluation-en} report the means by dataset and language.

For Figure~\ref{fig:human-evaluation}, we compare the MOS scores given by the same evaluator to the same query for CoSTA and each baseline.
A higher, equal, or lower score counts as a win, tie, or loss based on the MOS ratings.
Each language/metric/baseline comparison therefore contains $30\times6=180$ paired scores.
CoSTA has more wins than losses in every aggregate comparison, with ties forming the majority for fluency.
The paired comparisons provide qualitative evidence of perceived quality preservation, while detailed task-level variations are reported in Appendix~\ref{app:human-evaluation}.

\subsection{Detailed OpenS2S Transfer Results}
\label{app:opens2s-transfer}

We adapt OpenS2S using vocabulary extension followed by both pronunciation and dialogue adaptation in each language. It retains its native GLM-4-Voice tokenizer and speech decoder. Chinese evaluation uses AlpacaEval-zh, Claude-zh, and LCSTS-zh; English evaluation uses AlpacaEval, CommonEval, and Summary. The matched Vanilla-FT baselines use the corresponding adaptation data without BPE compression. Table~\ref{tab:opens2s-generality-detail} reports generated-token compression, ASR-CER for Chinese or ASR-WER for English, UTMOS, and G-Eval. CoSTA improves average content scores and alignment over matched fine-tuning while slightly lowering average UTMOS. These results demonstrate that CoSTA's compressed-token adaptation transfers across SDM architectures while maintaining favorable compression--quality trade-offs.

\subsection{Why Dialogue Adaptation Matters}
\label{app:dialogue-adaptation-analysis}

Removing dialogue adaptation increases $\rho_g$ from 66.73\% to 81.05\% and reduces G-Eval from 72.27 to 24.35 (Table~\ref{tab:ablation-zh}). To examine this degradation, we analyze token usage on all six Chinese benchmarks, with 1,122 responses per model. We pool generated speech tokens across responses, so the usage statistics are token-weighted rather than response-weighted.

\paragraph{Reduced use of merged tokens.}
Table~\ref{tab:dialogue-token-usage} distinguishes original speech tokens from added BPE units. Original and BPE shares are measured before expansion, and Original/BPE is the ratio of their counts. BPE coverage is the fraction of the expanded speech-token sequence contributed by BPE units; average BPE expansion is the mean number of original tokens represented by a generated BPE unit. For example, if $N_{\mathrm{orig}}$ is the number of generated original tokens and $\mathcal{B}$ is the multiset of generated BPE units, coverage is
\begin{equation}
    \frac{\sum_{z\in\mathcal{B}} |\pi(z)|}
    {N_{\mathrm{orig}}+\sum_{z\in\mathcal{B}} |\pi(z)|}.
\end{equation}

Without dialogue adaptation, the original-token share rises from 70.95\% to 78.95\%, an 8.00-percentage-point increase, while the Original/BPE ratio rises from 2.44 to 3.75. BPE coverage falls from 51.64\% to 36.71\%, a 14.93-percentage-point decrease, and average BPE expansion drops from 2.608 to 2.175. Table~\ref{tab:dialogue-token-usage}(b) further groups generated tokens by expansion length $L=|\pi(z)|$, where $L=1$ denotes an original token. Units with $L\geq5$ account for 2.29\% of generated speech tokens in CoSTA-ZH but only 0.45\% without dialogue adaptation. The ablated model therefore both uses merged units less often and selects shorter units when it does use them. These behaviors reduce the number of original tokens represented by each decoding step and help explain the weaker compression.

\paragraph{Failure to answer the user's query.}
The content degradation also appears in representative responses. For an MLC-zh query, translated from Chinese as ``If all flowers are plants, is a rose a plant?'', CoSTA-ZH answers ``Yes, a rose is a plant.'' The model without dialogue adaptation instead repeats the question. This example illustrates a failure to perform the dialogue task, consistent with the sharp decrease in G-Eval, rather than merely a change in token segmentation.

Taken together, these observations suggest that dialogue adaptation connects compressed-token generation learned from TTS data to appropriate spoken responses: it restores question-answering behavior while promoting more effective use of the extended BPE vocabulary. These complementary effects help explain why omitting the stage degrades both compression and response quality.

\subsection{Full-Model Results Across Vocabulary Sizes}
\label{app:full-model-vocab}

Table~\ref{tab:bpe-vocab-full-decoding} provides the complete AlpacaEval-zh results underlying Figure~\ref{fig:vocab-tradeoff}, including generated-token compression and absolute decoding speed. The 0K row is the original Step-Audio 2 mini model; the 10K--160K rows are adapted CoSTA-ZH models. Thus, the vocabulary-size comparison among adapted models uses the nonzero settings.

Increasing the added vocabulary from 10K to 80K reduces $\rho_g$ from 76.66\% to 67.50\% and raises expanded speech-token speed from 78.67 to 103.03 tokens/s. At 160K, $\rho_g$ further decreases to 63.62\%, but speed reaches only 104.18 tokens/s, a 1.1\% increase over 80K. Meanwhile, G-Eval decreases from 81.36 to 78.87, and UTMOS decreases by 0.02. These results motivate 80K as a balance of speed, compression, and quality under the tested conditions, rather than the setting that maximizes every metric.

\subsection{Computational Cost of Vocabulary Extension}
\label{app:lm-head-cost}

\paragraph{Measurement setting.}
Vocabulary extension affects inference asymmetrically.
The enlarged embedding table mainly increases parameter storage, while embedding lookup does not incur vocabulary-wide computation at each decoding step.
In contrast, the LM head projects each hidden state over the full vocabulary, and the subsequent token selection also grows with vocabulary size.
We therefore measure LM-head projection plus greedy argmax as the primary per-step computational overhead introduced by vocabulary extension.
Transformer computation is excluded because its cost depends on the evolving prefix length, which itself changes with token compression and is therefore difficult to attribute to vocabulary extension alone.

Table~\ref{tab:lm-head-scaling} reports the complete 0K--320K measurements for S3Tokenizer V2 using identical BF16 hidden states on a single A100 GPU at batch size 1, with median latency reported per step.
Compression uses exact pooled token counts from the held-out Dialogue-ZH corpus in Figure~\ref{fig:compression_rate}; it is distinct from generated-token compression in Table~\ref{tab:bpe-vocab-full-decoding}.
At 80K, the added output-projection rows contain 286.72M parameters and occupy 546.88 MiB in BF16.
These storage figures cover the LM head alone.

\paragraph{Compression versus per-step overhead.}
Let $t_M$ be the median projection-and-selection latency with $M$ added units and $\rho_M$ the compression rate.
We estimate the normalized output-layer cost for generating the same number of expanded speech tokens as
\begin{equation}
    C_{\mathrm{head}}(M)=\rho_M\frac{t_M}{t_0},
    \label{eq:normalized-head-cost}
\end{equation}
where $\rho_M$ is represented as a fraction.
This estimate captures the isolated output-layer cost and excludes Transformer computation.
At 80K, test compression is 62.29\%, while per-step latency rises from 0.6902 to 1.0251 ms (+48.52\%), yielding a normalized cost of 0.925.
At 160K, compression improves to 58.48\%, but latency reaches 1.3598 ms (+97.01\%), raising the normalized cost to 1.152.
At 320K, compression reaches 54.94\%, while latency increases to 2.0297 ms (+194.08\%) and the normalized cost to 1.616.
Thus, in this isolated measurement, continued compression gains eventually cease to offset the increasing per-step output-layer cost.

\paragraph{Relation to full-model decoding.}
The isolated measurements characterize the additional output-layer cost of vocabulary extension, but do not determine full-model efficiency.
Token compression also reduces the number of Transformer decoding steps, an effect captured by the full-model experiments rather than by $C_{\mathrm{head}}$.
Indeed, AlpacaEval-zh decoding speed still increases slightly from 80K to 160K despite the higher normalized output-layer cost.
Together, these results show why stronger compression does not necessarily translate into proportional full-model speedup and motivate considering compression, runtime, and response quality jointly when selecting vocabulary size.

\section{Held-Out Compression Analysis}
\label{app:compression-diagnostics}

\begin{figure}[t]
    \centering
    \includegraphics[width=\linewidth]{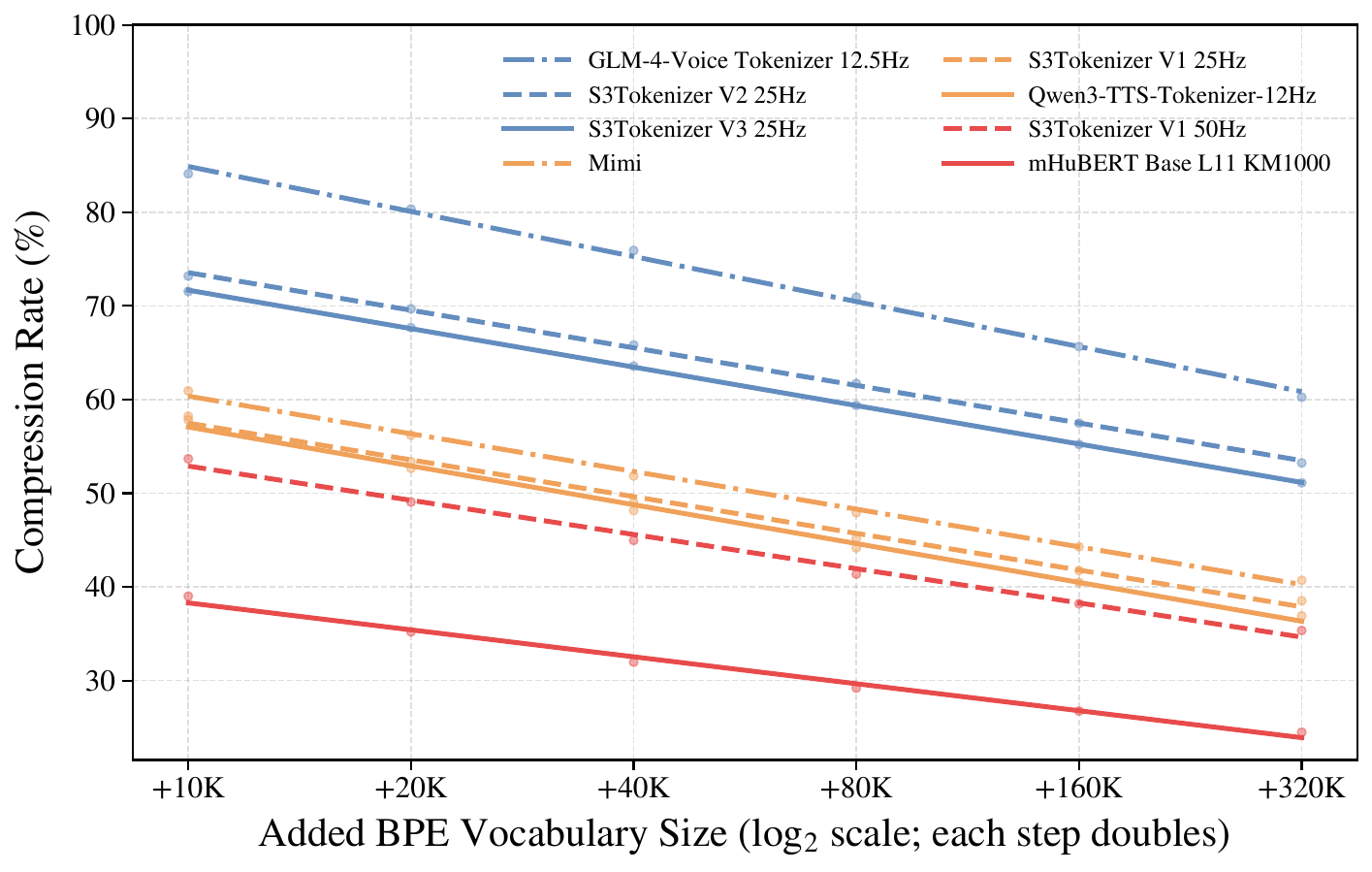}
    \caption{Training-corpus compression across eight tokenizer configurations on Chinese dialogue speech, corresponding to the test results in Figure~\ref{fig:tokenizer-scaling}. Points are measurements and lines are log-linear fits. Mimi and Qwen3-TTS use their first semantic codebook.}
    \label{fig:compression-training-tokenizer}
\end{figure}

\subsection{Evaluation and Fitting Protocol}

\paragraph{Tokenizer configurations and sources.}
Table~\ref{tab:tokenizer-configurations} documents the eight configurations and links their public reference releases. GLM-4-Voice, Mimi, and Qwen3-TTS follow their respective model families~\cite{GLM-4-Voice,Moshi,Qwen3TTS}. The \href{https://github.com/xingchensong/S3Tokenizer}{S3Tokenizer implementation} provides the CosyVoice V1--V3 tokenizers~\cite{CosyVoice,CosyVoice2,CosyVoice3}. mHuBERT Base L11 KM1000 uses layer-11 features with 1,000 $K$-means clusters~\cite{mHuBERTRealData}. Codebook sizes exclude added BPE units.

\input{tabs/tokenizer_configurations}

\paragraph{Training and test evaluation.}
For each tokenizer/corpus setting and added vocabulary size $M\in\{10,20,40,80,160,320\}$K, BPE is learned on the training corpus. The resulting vocabulary and merge rules are fixed when encoding the corresponding test corpus. Figure~\ref{fig:tokenizer-scaling} reports Chinese dialogue test speech for eight tokenizer configurations; Figure~\ref{fig:compression_rate} uses S3Tokenizer V2 at 25 Hz across three data sources and three language settings. Each setting has its own trained BPE vocabulary. The evaluation concerns held-out speech within each setting, not transfer of one vocabulary across settings. For Mimi and Qwen3-TTS, only the first semantic codebook is included.

\begin{figure}[t]
    \centering
    \includegraphics[width=0.95\linewidth]{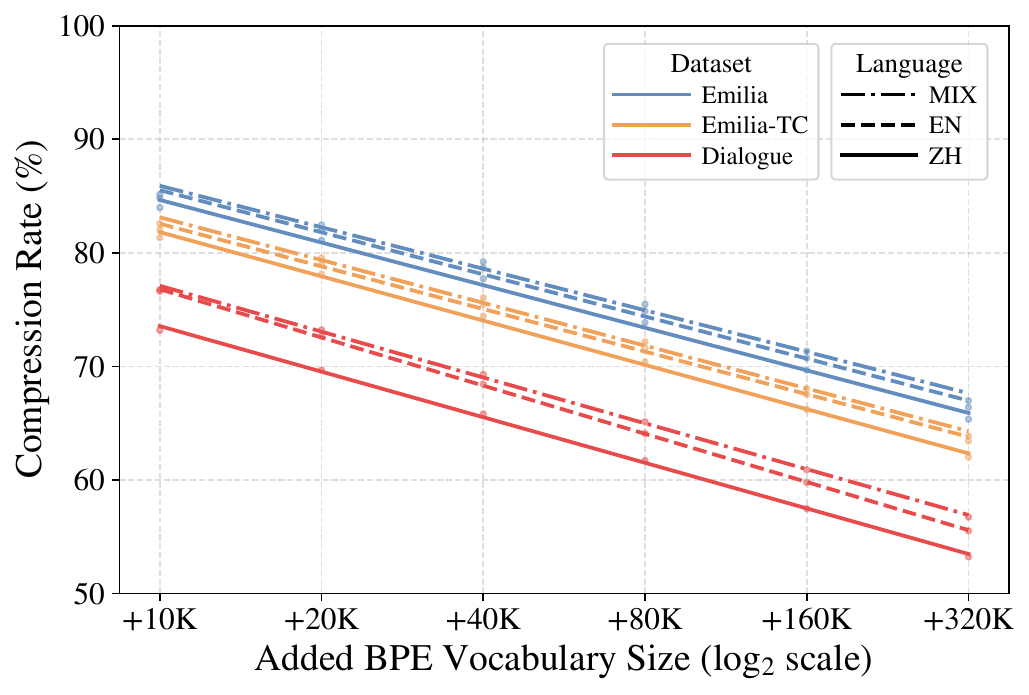}
    \caption{Training-corpus compression across data sources and languages with S3Tokenizer V2, corresponding to the test results in Figure~\ref{fig:compression_rate}. Points are measurements and lines are log-linear fits. Each vocabulary is learned on the training corpus represented by its curve.}
    \label{fig:compression-training-dataset}
\end{figure}

\paragraph{Corpus-level compression.}
We pool token counts across utterances:
\begin{equation}
\rho_{\mathcal{D}}(M)=\frac{\sum_{\mathbf{s}\in\mathcal{D}}|B_M(\mathbf{s})|}{\sum_{\mathbf{s}\in\mathcal{D}}|\mathbf{s}|}.
\end{equation}
Thus, rates are token-weighted, not averages of per-utterance ratios. Figures~\ref{fig:compression-training-tokenizer} and~\ref{fig:compression-training-dataset} preserve the training measurements for comparison. These rates characterize encoding of existing speech and are distinct from compression of model-generated responses.

\paragraph{Fitting and numerical precision.}
We fit $y=\alpha+\beta x$ by unweighted ordinary least squares, where $x=\log_2(M/10\mathrm{K})$ and $y=100\rho$. Here $\alpha$ is the fitted percentage at 10K and $\beta<0$ is the change in percentage points per vocabulary doubling. This is the percentage-scale form of Equation~\ref{eq:compression-scaling}. All six vocabulary settings have equal weight; 0K is excluded because its logarithm is undefined. With residuals $e_j=y_j-\hat y_j$, we report
\begin{equation}
\begin{aligned}
\mathrm{RMSE}&=\sqrt{\frac{1}{6}\sum_{j=1}^{6}e_j^2},\\
R^2&=1-\frac{\sum_j e_j^2}{\sum_j(y_j-\bar y)^2}.
\end{aligned}
\end{equation}
Training fits use the original plotted percentages rounded to 0.01 pp. Tokenizer test fits use the reported four-decimal ratios, also giving 0.01-pp precision; dataset/language test fits use exact pooled token counts. Consequently, the S3 V2 Chinese-dialogue curves can differ slightly between panels despite representing the same setting. Tables~\ref{tab:compression-tokenizer-fits} and~\ref{tab:compression-dataset-fits} report all 34 fits.

\subsection{Fit Quality and Training--Test Consistency}

\paragraph{Persistence on test speech.}
Across the eight tokenizer test curves, $R^2$ ranges from 0.9807 to 0.9997 and RMSE from 0.103 to 0.883 pp. Fitted compression falls by 2.70--4.19 pp per vocabulary doubling. Across the nine dataset/language test curves, $R^2$ ranges from 0.9942 to 0.9997 and RMSE from 0.104 to 0.474 pp, with decreases of 3.62--4.01 pp per doubling. Thus, the approximate trend persists across the evaluated test settings, with setting-dependent levels and slopes.

\paragraph{Training--test consistency.}
The log-linear scaling trend remains consistent between training and held-out speech.
Across all eight tokenizer configurations, test slopes are close to their training counterparts, with differences of 0.25--1.46 pp at 80K and 0.84--3.52 pp at 320K.
For example, at 320K, GLM-4-Voice changes from 60.25\% on training speech to 63.77\% on held-out speech, while Mimi changes from 40.69\% to 43.60\%.
Across data sources, the differences remain small: 320K gaps are at most 0.24 pp for Emilia and Emilia-TC, and 0.85--1.71 pp for dialogue speech.
These results show that the observed scaling regularity consistently transfers from training data to held-out speech across tokenizer configurations, data sources, and languages.

\paragraph{Approximation quality and scope.}
Tables~\ref{tab:compression-tokenizer-fits} and~\ref{tab:compression-dataset-fits} quantify the approximation quality.
The largest absolute test residual is 1.21 pp for Qwen3-TTS and 0.67 pp across dataset/language curves, while the fitted log-linear models achieve high $R^2$ across settings.
These results show that the log-linear scaling trend provides an accurate empirical description over the tested 10K--320K vocabulary range across tokenizer configurations, data sources, and languages.

\section{Case Study}
\label{app:case-study}

We present representative case studies selected from the English AlpacaEval and WildchatEval benchmarks.
Case Studies~1--2 cover factual and commonsense questions, Case Study~3 illustrates sentence rewriting, Case Study~4 presents an open-ended dialogue, and Case Study~5 shows an empathetic customer-service reply.
Across these cases, CoSTA generates compact speech-token sequences while preserving response quality; the generated BPE units are deterministically expanded back to the original speech-token space before waveform synthesis.
These examples qualitatively demonstrate efficient compressed-token generation through token-space decoupling with the unchanged speech decoder.

\section{Licenses of Scientific Artifacts}
\label{app:licenses}

We follow and report the licenses or usage terms of the scientific artifacts used in this work in Table~\ref{tab:licenses}.

\begin{table}[!htb]
\centering
\small
\renewcommand\arraystretch{1.05}
\setlength{\tabcolsep}{2.5mm}
\begin{tabular}{l !{\vrule} l}
\toprule
\textbf{Name} & \textbf{License / Terms} \\
\midrule
Step-Audio 2 mini & Apache 2.0 license \\
Qwen2.5-Omni & Apache 2.0 license \\
Kimi-Audio & Apache 2.0 / MIT licenses \\
GLM-4-Voice / Tokenizer & glm-4-voice Model License \\
OpenS2S & Apache 2.0 license \\
Emilia & CC BY-NC 4.0 license \\
InstructS2S-200K & CC BY-NC 4.0 license \\
Seed-VC & GPL-3.0 license \\
Qwen3-32B & Apache 2.0 license \\
\shortstack[l]{CosyVoice 1/2/3\\(S3Tokenizer V1--V3)}
    & Apache 2.0 license \\
Mimi & CC BY 4.0 license \\
Qwen3-TTS Tokenizer & Apache 2.0 license \\
mHuBERT / fairseq & MIT license \\
URO-Bench & MIT license \\
UTMOS & MIT license \\
HuggingFace tokenizers & Apache 2.0 license \\
Transformers & Apache 2.0 license \\
vLLM & Apache 2.0 license \\
DeepSpeed & Apache 2.0 license \\
Whisper & MIT license \\
GPT-4o-mini API & OpenAI Service Terms \\
\bottomrule
\end{tabular}
\caption{Licenses and usage terms of scientific artifacts used in this work.}
\label{tab:licenses}
\end{table}

\clearpage

\input{tabs/quality_all_zh}

\input{tabs/quality_all_en}

\input{tabs/scaling_tts_in_50k_s2s}

\input{tabs/scaling_s2s_in_300h_tts}

\input{tabs/latency}

\input{tabs/human_zh}

\input{tabs/human_en}

\input{tabs/opens2s}

\clearpage

\input{tabs/dialogue_token_usage}

\input{tabs/vocab_size_ablation}

\input{tabs/cost}

\clearpage

\input{tabs/compression_tokenizer_fits}
\input{tabs/compression_dataset_fits}

\clearpage
\input{tabs/case_study}

\end{document}

%% file: tabs/efficiency.tex
\begin{table*}[t]
\centering
\large
\renewcommand\arraystretch{1.15}
\setlength{\tabcolsep}{0.9mm}

\resizebox{\textwidth}{!}{
\begin{tabular}{l cc !{\vrule} cc !{\vrule} cc !{\vrule} cc !{\vrule} cc !{\vrule} cc !{\vrule} ccc}
\toprule[1.2pt]

\multirow{2}{*}{\textbf{Model}}
& \multicolumn{2}{c}{\textbf{AlpacaEval-zh}}
& \multicolumn{2}{c}{\textbf{APE-zh}}
& \multicolumn{2}{c}{\textbf{Claude-zh}}
& \multicolumn{2}{c}{\textbf{LCSTS-zh}}
& \multicolumn{2}{c}{\textbf{MLC-zh}}
& \multicolumn{2}{c}{\textbf{Wildchat-zh}}
& \multicolumn{3}{c}{\textbf{Avg}} \\

\cmidrule(lr){2-3}
\cmidrule(lr){4-5}
\cmidrule(lr){6-7}
\cmidrule(lr){8-9}
\cmidrule(lr){10-11}
\cmidrule(lr){12-13}
\cmidrule(lr){14-16}

& $\boldsymbol{\rho_g}\downarrow$ & \textbf{Speed}$\uparrow$
& $\boldsymbol{\rho_g}\downarrow$ & \textbf{Speed}$\uparrow$
& $\boldsymbol{\rho_g}\downarrow$ & \textbf{Speed}$\uparrow$
& $\boldsymbol{\rho_g}\downarrow$ & \textbf{Speed}$\uparrow$
& $\boldsymbol{\rho_g}\downarrow$ & \textbf{Speed}$\uparrow$
& $\boldsymbol{\rho_g}\downarrow$ & \textbf{Speed}$\uparrow$
& $\boldsymbol{\rho_g}\downarrow$ & \textbf{Speed}$\uparrow$ & \textbf{Relative}$\uparrow$ \\

\midrule[0.8pt]

Step-Audio 2 mini
& 100.00\% & 55.51
& 100.00\% & 60.88
& 100.00\% & 56.39
& 100.00\% & 57.05
& 100.00\% & 65.22
& 100.00\% & 57.40
& 100.00\% & 58.74 & 100.00\% \\

Vanilla-FT-ZH
& 100.00\% & 57.84
& 100.00\% & 60.40
& 100.00\% & 54.99
& 100.00\% & 55.54
& 100.00\% & 63.47
& 100.00\% & 59.10
& 100.00\% & 58.56 & 99.69\% \\

Vanilla-FT-MIX
& 100.00\% & 59.01
& 100.00\% & 59.65
& 100.00\% & 56.93
& 100.00\% & 56.03
& 100.00\% & 61.57
& 100.00\% & 59.08
& 100.00\% & 58.71 & 99.95\% \\

\noalign{\vskip\aboverulesep}
\cdashline{1-16}
\noalign{\vskip\belowrulesep}

\textbf{CoSTA-ZH}
& \textbf{67.50\%} & \textbf{103.03}
& \textbf{64.63\%} & \underline{101.68}
& \textbf{68.52\%} & \textbf{101.20}
& \textbf{69.40\%} & \textbf{94.46}
& \textbf{62.70\%} & \textbf{110.31}
& \textbf{68.13\%} & \textbf{101.29}
& \textbf{66.81\%} & \textbf{101.99} & \textbf{173.63\%} \\

\textbf{CoSTA-MIX}
& \underline{69.12\%} & \underline{98.44}
& \underline{64.68\%} & \textbf{102.34}
& \underline{70.58\%} & \underline{92.54}
& \underline{71.06\%} & \underline{91.37}
& \underline{63.03\%} & \underline{106.32}
& \underline{70.21\%} & \underline{93.22}
& \underline{68.11\%} & \underline{97.37} & \underline{165.76\%} \\

\midrule[1.0pt]

\multirow{2}{*}{\textbf{Model}}
& \multicolumn{2}{c}{\textbf{AlpacaEval}}
& \multicolumn{2}{c}{\textbf{Gsm8kEval}}
& \multicolumn{2}{c}{\textbf{CommonEval}}
& \multicolumn{2}{c}{\textbf{Summary}}
& \multicolumn{2}{c}{\textbf{MLC}}
& \multicolumn{2}{c}{\textbf{WildchatEval}}
& \multicolumn{3}{c}{\textbf{Avg}} \\

\cmidrule(lr){2-3}
\cmidrule(lr){4-5}
\cmidrule(lr){6-7}
\cmidrule(lr){8-9}
\cmidrule(lr){10-11}
\cmidrule(lr){12-13}
\cmidrule(lr){14-16}

& $\boldsymbol{\rho_g}\downarrow$ & \textbf{Speed}$\uparrow$
& $\boldsymbol{\rho_g}\downarrow$ & \textbf{Speed}$\uparrow$
& $\boldsymbol{\rho_g}\downarrow$ & \textbf{Speed}$\uparrow$
& $\boldsymbol{\rho_g}\downarrow$ & \textbf{Speed}$\uparrow$
& $\boldsymbol{\rho_g}\downarrow$ & \textbf{Speed}$\uparrow$
& $\boldsymbol{\rho_g}\downarrow$ & \textbf{Speed}$\uparrow$
& $\boldsymbol{\rho_g}\downarrow$ & \textbf{Speed}$\uparrow$ & \textbf{Relative}$\uparrow$ \\

\midrule[0.8pt]

Step-Audio 2 mini
& 100.00\% & 57.93
& 100.00\% & 60.03
& 100.00\% & 60.62
& 100.00\% & 58.74
& 100.00\% & 61.80
& 100.00\% & 59.00
& 100.00\% & 59.69 & 100.00\% \\

Vanilla-FT-EN
& 100.00\% & 56.73
& 100.00\% & 58.14
& 100.00\% & 56.96
& 100.00\% & 56.16
& 100.00\% & 61.36
& 100.00\% & 56.67
& 100.00\% & 57.67 & 96.61\% \\

Vanilla-FT-MIX
& 100.00\% & 57.43
& 100.00\% & 59.27
& 100.00\% & 57.29
& 100.00\% & 57.66
& 100.00\% & 61.90
& 100.00\% & 56.43
& 100.00\% & 58.33 & 97.72\% \\

\noalign{\vskip\aboverulesep}
\cdashline{1-16}
\noalign{\vskip\belowrulesep}

\textbf{CoSTA-EN}
& \textbf{73.97\%} & \textbf{87.60}
& \underline{65.60\%} & \textbf{102.04}
& \textbf{71.60\%} & \textbf{98.41}
& \textbf{72.09\%} & \textbf{95.21}
& \underline{64.71\%} & \underline{115.58}
& \textbf{73.06\%} & \textbf{95.46}
& \textbf{70.17\%} & \textbf{99.05} & \textbf{165.94\%} \\

\textbf{CoSTA-MIX}
& \underline{75.45\%} & \underline{81.17}
& \textbf{65.55\%} & \underline{97.92}
& \underline{72.34\%} & \underline{88.91}
& \underline{73.94\%} & \underline{83.29}
& \textbf{64.30\%} & \textbf{123.95}
& \underline{74.56\%} & \underline{87.80}
& \underline{71.02\%} & \underline{93.84} & \underline{157.21\%} \\

\bottomrule[1.2pt]
\end{tabular}
}

\caption{Model-side efficiency on Chinese and English speech-to-speech benchmarks. $\rho_g$ denotes generated-token compression, Speed denotes expanded speech-token generation speed, and Relative normalizes Speed to Step-Audio 2 mini within each language setting.}
\label{tab:efficiency-results}
\vspace{-0.3em}
\end{table*}

%% file: tabs/system.tex
\begin{table*}[t]
\centering
\small
\renewcommand{\arraystretch}{1.08}
\setlength{\tabcolsep}{4.5pt}

\resizebox{\textwidth}{!}{
\begin{tabular}{c l c c c}
\toprule
\textbf{Concurrency}
& \textbf{Model}
& \textbf{Full-Response Latency (s)}$\downarrow$
& \textbf{Throughput (req/s)}$\uparrow$
& \textbf{Peak Memory (GiB)} \\
\midrule

1
& Step-Audio 2 mini / Vanilla-FT-ZH / CoSTA-ZH
& 10.73 / 9.72 / \textbf{7.14}
& 0.103 / 0.114 / \textbf{0.160}
& 78.12 / 77.64 / 78.56 \\

4
& Step-Audio 2 mini / Vanilla-FT-ZH / CoSTA-ZH
& 13.63 / 11.38 / \textbf{9.16}
& 0.316 / 0.369 / \textbf{0.479}
& 78.34 / 78.13 / 79.02 \\

8
& Step-Audio 2 mini / Vanilla-FT-ZH / CoSTA-ZH
& 15.56 / 13.71 / \textbf{9.49}
& 0.515 / 0.601 / \textbf{0.820}
& 78.90 / 79.17 / 79.14 \\

\midrule

1
& Step-Audio 2 mini / Vanilla-FT-EN / CoSTA-EN
& 12.50 / 16.16 / \textbf{9.25}
& 0.088 / 0.068 / \textbf{0.122}
& 78.98 / 77.16 / 78.69 \\

4
& Step-Audio 2 mini / Vanilla-FT-EN / CoSTA-EN
& 15.52 / 20.17 / \textbf{11.59}
& 0.276 / 0.212 / \textbf{0.372}
& 79.24 / 78.33 / 79.12 \\

8
& Step-Audio 2 mini / Vanilla-FT-EN / CoSTA-EN
& 18.62 / 25.04 / \textbf{14.44}
& 0.442 / 0.330 / \textbf{0.556}
& 78.52 / 77.73 / 79.04 \\

\bottomrule
\end{tabular}
}

\caption{
System-level efficiency on AlpacaEval-zh (upper) and AlpacaEval (lower)
at concurrency 1, 4, and 8 on a single A100 80GB GPU.
Values are ordered as Step-Audio 2 mini, Vanilla-FT, and CoSTA.
}
\label{tab:full-response-efficiency}
\vspace{-1.5em}
\end{table*}

%% file: tabs/quality.tex
\begin{table*}[t]
\centering
\large
\renewcommand\arraystretch{0.85}
\setlength{\tabcolsep}{0.8mm}

\resizebox{\textwidth}{!}{
\begin{tabular}{l c c c c c c !{\vrule} c c c}
\toprule[1.2pt]

\multirow{2}{*}{\textbf{Model}}
& \multirow{2}{*}{\textbf{AlpacaEval-zh}}
& \multirow{2}{*}{\textbf{APE-zh}}
& \multirow{2}{*}{\textbf{Claude-zh}}
& \multirow{2}{*}{\textbf{LCSTS-zh}}
& \multirow{2}{*}{\textbf{MLC-zh}}
& \multirow{2}{*}{\textbf{Wildchat-zh}}
& \multicolumn{3}{c}{\textbf{Avg}} \\

\cmidrule(lr){8-10}

&
&
&
&
&
&
& \textbf{G-Eval}$\uparrow$
& \textbf{ASR-CER}$\downarrow$
& \textbf{UTMOS}$\uparrow$ \\

\midrule[0.8pt]

Qwen2.5-Omni
& 74.10 & 62.11 & 78.86 & 69.75 & 82.53 & 78.35
& 74.28 & 2.93\% & 2.95 \\

Kimi-Audio
& 77.05 & 39.12 & 80.24 & 76.13 & 80.92 & 78.08
& 71.93 & 4.88\% & 2.52 \\

GLM-4-Voice
& 84.76 & 23.16 & 83.54 & 76.36 & 78.85 & 84.41
& 71.85 & 3.77\% & 2.94 \\

\midrule[0.8pt]

Step-Audio 2 mini
& 80.18 & \underline{21.93} & 72.40 & \underline{79.16} & 78.16 & \textbf{83.59}
& 69.24 & 7.49\% & 3.62 \\

Vanilla-FT-ZH
& 79.77 & 21.40 & 80.35 & 78.82 & 77.47 & 81.47
& 69.88 & 4.96\% & \textbf{3.77} \\

Vanilla-FT-MIX
& \underline{81.27} & 20.53 & 79.28 & 79.10 & \underline{79.31} & 82.10
& 70.26 & 7.02\% & \textbf{3.77} \\

\noalign{\vskip\aboverulesep}
\cdashline{1-10}
\noalign{\vskip\belowrulesep}

\textbf{CoSTA-ZH}
& \textbf{81.36} & \textbf{24.56} & \textbf{86.49} & \textbf{80.56} & \textbf{79.54} & 81.09
& \textbf{72.27} & \textbf{1.51\%} & \underline{3.75} \\

\textbf{CoSTA-MIX}
& 80.95 & \textbf{24.56} & \underline{85.05} & 78.99 & 77.24 & \underline{82.39}
& \underline{71.53} & \underline{1.65\%} & \underline{3.75} \\

\midrule[1.0pt]

\multirow{2}{*}{\textbf{Model}}
& \multirow{2}{*}{\textbf{AlpacaEval}}
& \multirow{2}{*}{\textbf{Gsm8kEval}}
& \multirow{2}{*}{\textbf{CommonEval}}
& \multirow{2}{*}{\textbf{Summary}}
& \multirow{2}{*}{\textbf{MLC}}
& \multirow{2}{*}{\textbf{WildchatEval}}
& \multicolumn{3}{c}{\textbf{Avg}} \\

\cmidrule(lr){8-10}

&
&
&
&
&
&
& \textbf{G-Eval}$\uparrow$
& \textbf{ASR-WER}$\downarrow$
& \textbf{UTMOS}$\uparrow$ \\

\midrule[0.8pt]

Qwen2.5-Omni
& 76.78 & 54.30 & 72.00 & 80.06 & 77.40 & 76.71
& 72.88 & 1.73\% & 4.28 \\

Kimi-Audio
& 62.01 & 0.57 & 50.77 & 82.15 & 40.87 & 66.40
& 50.46 & 34.92\% & 2.86 \\

GLM-4-Voice
& 80.87 & 29.32 & 66.77 & 92.15 & 69.30 & 81.26
& 69.95 & 11.55\% & 4.16 \\

\midrule[0.8pt]

Step-Audio 2 mini
& 76.31 & 15.92 & 52.53 & 88.14 & \underline{72.69} & 76.75
& 63.73 & 6.49\% & \underline{4.53} \\

Vanilla-FT-EN
& 73.74 & \underline{16.67} & 72.03 & 86.10 & 72.50 & 78.47
& 66.59 & 10.16\% & 4.46 \\

Vanilla-FT-MIX
& 76.18 & 11.05 & 72.63 & 85.37 & 68.55 & 78.93
& 65.45 & 12.00\% & 4.49 \\

\noalign{\vskip\aboverulesep}
\cdashline{1-10}
\noalign{\vskip\belowrulesep}

\textbf{CoSTA-EN}
& \underline{81.34} & 15.52 & \textbf{75.87} & \underline{89.21} & \textbf{74.01} & \textbf{83.84}
& \textbf{69.97} & \textbf{2.50\%} & \textbf{4.54} \\

\textbf{CoSTA-MIX}
& \textbf{83.32} & \textbf{17.53} & \underline{73.47} & \textbf{90.00} & 71.19 & \underline{82.52}
& \underline{69.67} & \underline{2.64\%} & \textbf{4.54} \\

\bottomrule[1.2pt]
\end{tabular}
}

\caption{Automatic quality evaluation on Chinese and English speech-to-speech benchmarks.
Benchmark columns report G-Eval; Avg reports average G-Eval, ASR-CER/WER, and UTMOS.
\textbf{Bold} and \underline{underline} mark the best and second-best results among Step-Audio 2 mini variants; external models serve as quality references.}
\label{tab:main-results-uro}
\vspace{-1.4em}
\end{table*}

%% file: tabs/ablation_zh.tex
\begin{table}[t]
\centering
\large
\renewcommand\arraystretch{1.0}
\setlength{\tabcolsep}{0.8mm}

\resizebox{\linewidth}{!}{
\begin{tabular}{l c c c c}
\toprule[1.2pt]

\textbf{Variant}
& $\boldsymbol{\rho_g}\downarrow$
& \textbf{ASR-CER}$\downarrow$
& \textbf{UTMOS}$\uparrow$
& \textbf{G-Eval}$\uparrow$ \\

\midrule[0.8pt]

Step-Audio 2 mini
& 100.00\% & 7.49\% & 3.62 & 69.24 \\

Vanilla-FT-ZH
& 100.00\% & 4.96\% & \textbf{3.77} & 69.88 \\

\noalign{\vskip\aboverulesep}
\cdashline{1-5}
\noalign{\vskip\belowrulesep}

\textbf{CoSTA-ZH}
& \textbf{66.73\%} & \underline{1.51\%} & \underline{3.75} & \textbf{72.27} \\

\quad w/o Timbre Conversion
& \underline{67.01\%} & 1.52\% & \underline{3.75} & 70.93 \\

\quad w/o Progressive Adaptation
& 67.75\% & \textbf{1.46\%} & \underline{3.75} & \underline{72.19} \\

\quad w/o Pronunciation Adaptation
& 69.56\% & 1.74\% & 3.74 & 72.12 \\

\quad w/o Dialogue Adaptation
& 81.05\% & 2.68\% & 3.74 & 24.35 \\

\bottomrule[1.2pt]
\end{tabular}
}

\caption{Component ablation on Chinese benchmarks. $\rho_g$, ASR-CER, UTMOS, and G-Eval measure generated-token compression, text-speech alignment, speech quality, and response-content quality, respectively.}
\label{tab:ablation-zh}
\vspace{-1.5em}
\end{table}

%% file: tabs/multi_turn_duration.tex
\begin{table}[t]
\centering
\large
\renewcommand\arraystretch{1.0}
\setlength{\tabcolsep}{0.8mm}

\resizebox{\linewidth}{!}{
\begin{tabular}{l c c c c c}
\toprule[1.2pt]

\textbf{Model}
& \textbf{\# Turns}$\uparrow$
& \textbf{Gen Tokens}
& \textbf{Final Tokens}
& \textbf{Duration (s)}
& $\boldsymbol{\rho_g}\downarrow$ \\

\midrule[0.8pt]

Step-Audio 2 mini
& 2.14 & 2208.21 & 2208.21 & 88.33 & 100.00\% \\

Vanilla-FT-ZH
& 3.71 & 2440.50 & 2440.50 & 97.62 & 100.00\% \\

\textbf{CoSTA-ZH}
& \textbf{7.21} & \textbf{2500.79} & \textbf{3766.86} & \textbf{150.67} & \textbf{66.39\%} \\

\midrule[0.8pt]

Step-Audio 2 mini
& 2.58 & 2218.74 & 2218.74 & 88.75 & 100.00\% \\

Vanilla-FT-EN
& 1.84 & 1894.79 & 1894.79 & 75.79 & 100.00\% \\

\textbf{CoSTA-EN}
& \textbf{5.79} & \textbf{2608.53} & \textbf{3650.79} & \textbf{146.03} & \textbf{71.45\%} \\

\bottomrule[1.2pt]
\end{tabular}
}

\caption{Interaction sustainability under a limited generation budget. Gen/Final Tokens denote speech tokens before/after deterministic expansion; Duration denotes generated speech duration.}
\label{tab:multi-turn-duration}
\vspace{-1.5em}
\end{table}

%% file: tabs/sdm_gen.tex
\begin{table}[t]
\centering
\large
\renewcommand\arraystretch{1.0}
\setlength{\tabcolsep}{1.5mm}

\resizebox{\linewidth}{!}{
\begin{tabular}{l c c c c}
\toprule[1.2pt]

\textbf{Model}
& $\boldsymbol{\rho_g}\downarrow$
& \textbf{ASR-CER/WER}$\downarrow$
& \textbf{UTMOS}$\uparrow$
& \textbf{G-Eval}$\uparrow$ \\

\midrule[0.8pt]

OpenS2S
& 100.00\%
& 11.08\%
& \textbf{2.41}
& 65.26 \\

Vanilla-FT-ZH
& 100.00\%
& 11.36\%
& 2.38
& 67.39 \\

\textbf{CoSTA-ZH}
& \textbf{78.83\%}
& \textbf{8.50\%}
& 2.36
& \textbf{69.18} \\

\midrule

OpenS2S
& 100.00\%
& \textbf{4.65\%}
& \textbf{3.81}
& 73.97 \\

Vanilla-FT-EN
& 100.00\%
& 5.16\%
& 3.79
& 75.48 \\

\textbf{CoSTA-EN}
& \textbf{81.36\%}
& 4.66\%
& 3.78
& \textbf{76.23} \\

\bottomrule[1.2pt]
\end{tabular}
}

\caption{Transfer of CoSTA to OpenS2S. Upper/lower blocks report Chinese/English results using ASR-CER/WER, respectively. Matched Vanilla-FT uses the same architecture and language-specific adaptation data.}
\label{tab:opens2s-generality}
\vspace{-1.7em}
\end{table}

%% file: tabs/tokenizer_configurations.tex
\begin{table*}[t]
\centering\small
\renewcommand{\arraystretch}{1.05}
\setlength{\tabcolsep}{3pt}
\begin{tabular}{@{}>{\raggedright\arraybackslash}p{0.265\textwidth}>{\raggedright\arraybackslash}p{0.265\textwidth}rr>{\raggedright\arraybackslash}p{0.25\textwidth}@{}}
\toprule
Tokenizer & Public reference source & Codes & Hz & Token stream used for BPE \\
\midrule
GLM-4-Voice Tokenizer 12.5Hz & \href{https://huggingface.co/zai-org/glm-4-voice-tokenizer}{GLM-4-Voice tokenizer} & 16,384 & 12.5 & Single quantized speech stream \\
\addlinespace[1.5pt]
S3Tokenizer V2 25Hz & \href{https://www.modelscope.cn/models/iic/CosyVoice2-0.5B}{CosyVoice2-0.5B} & 6,561 & 25 & Single semantic-token stream \\
\addlinespace[1.5pt]
S3Tokenizer V3 25Hz & \href{https://huggingface.co/FunAudioLLM/Fun-CosyVoice3-0.5B-2512}{Fun-CosyVoice3-0.5B-2512} & 6,561 & 25 & Single semantic-token stream \\
\addlinespace[1.5pt]
Mimi & \href{https://huggingface.co/kyutai/mimi}{kyutai/mimi} & 2,048 & 12.5 & First semantic codebook only \\
\addlinespace[1.5pt]
S3Tokenizer V1 25Hz & \href{https://www.modelscope.cn/models/iic/CosyVoice-300M-25Hz}{CosyVoice-300M-25Hz} & 4,096 & 25 & Single semantic-token stream \\
\addlinespace[1.5pt]
Qwen3-TTS-Tokenizer-12Hz & \href{https://huggingface.co/Qwen/Qwen3-TTS-Tokenizer-12Hz}{Qwen3-TTS tokenizer} & 2,048 & 12.5 & First semantic codebook only \\
\addlinespace[1.5pt]
S3Tokenizer V1 50Hz & \href{https://www.modelscope.cn/models/iic/CosyVoice-300M}{CosyVoice-300M} & 4,096 & 50 & Single semantic-token stream \\
\addlinespace[1.5pt]
mHuBERT Base L11 KM1000 & \href{https://github.com/facebookresearch/fairseq/blob/main/examples/speech_to_speech/docs/textless_s2st_real_data.md}{fairseq mHuBERT Base (L11, KM1000)} & 1,000 & -- & Layer-11 features quantized by $K$-means ($K=1{,}000$) \\
\bottomrule
\end{tabular}
\caption{Tokenizer configurations for Chinese dialogue compression, using the figure names and linked public sources. Codes excludes added BPE units. Hz is the native rate; mHuBERT is omitted. Qwen3-TTS operates at 12.5 Hz despite its model name (\href{https://huggingface.co/Qwen/Qwen3-TTS-Tokenizer-12Hz/blob/main/config.json}{official configuration}).}
\label{tab:tokenizer-configurations}
\end{table*}

%% file: tabs/quality_all_zh.tex
\begin{table*}[t]
\centering
\normalsize
\renewcommand\arraystretch{1.12}
\setlength{\tabcolsep}{1.2mm}

\resizebox{\textwidth}{!}{
\begin{tabular}{l ccc !{\vrule} ccc !{\vrule} ccc !{\vrule} ccc}
\toprule[1.2pt]

\multirow{2}{*}{\textbf{Model}}
& \multicolumn{3}{c}{\textbf{AlpacaEval-zh}}
& \multicolumn{3}{c}{\textbf{APE-zh}}
& \multicolumn{3}{c}{\textbf{Claude-zh}}
& \multicolumn{3}{c}{\textbf{Avg}} \\

\cmidrule(lr){2-4}
\cmidrule(lr){5-7}
\cmidrule(lr){8-10}
\cmidrule(lr){11-13}

& \textbf{CER}$\downarrow$ & \textbf{UTMOS}$\uparrow$ & \textbf{G-Eval}$\uparrow$
& \textbf{CER}$\downarrow$ & \textbf{UTMOS}$\uparrow$ & \textbf{G-Eval}$\uparrow$
& \textbf{CER}$\downarrow$ & \textbf{UTMOS}$\uparrow$ & \textbf{G-Eval}$\uparrow$
& \textbf{CER}$\downarrow$ & \textbf{UTMOS}$\uparrow$ & \textbf{G-Eval}$\uparrow$ \\

\midrule[0.8pt]

Qwen2.5-Omni
& 2.09\% & 2.93 & 74.10
& 8.50\% & 3.02 & 62.11
& 2.47\% & 2.73 & 78.86
& 2.93\% & 2.95 & 74.28 \\

Kimi-Audio
& 4.71\% & 2.52 & 77.05
& 8.29\% & 2.59 & 39.12
& 11.12\% & 2.08 & 80.24
& 4.88\% & 2.52 & 71.93 \\

GLM-4-Voice
& 3.65\% & 2.86 & 84.76
& 5.51\% & 3.14 & 23.16
& 7.13\% & 2.59 & 83.54
& 3.77\% & 2.94 & 71.85 \\

\midrule[0.8pt]

Step-Audio 2 mini
& 8.48\% & 3.58 & 80.18
& 4.35\% & 3.87 & 21.93
& 20.28\% & 3.33 & 72.40
& 7.49\% & 3.62 & 69.24 \\

Vanilla-FT-ZH
& 5.11\% & 3.76 & 79.77
& 0.95\% & 4.06 & 21.40
& 19.51\% & 3.41 & 80.35
& 4.96\% & 3.77 & 69.88 \\

Vanilla-FT-MIX
& 6.32\% & 3.74 & 81.27
& 1.00\% & 4.12 & 20.53
& 27.73\% & 3.39 & 79.28
& 7.02\% & 3.77 & 70.26 \\

\noalign{\vskip\aboverulesep}
\cdashline{1-13}
\noalign{\vskip\belowrulesep}


CoSTA-ZH
& 1.92\% & 3.70 & 81.36
& 1.37\% & 4.04 & 24.56
& 2.62\% & 3.43 & 86.49
& 1.51\% & 3.75 & 72.27 \\

CoSTA-MIX
& 1.96\% & 3.70 & 80.95
& 2.33\% & 4.07 & 24.56
& 2.55\% & 3.39 & 85.05
& 1.65\% & 3.75 & 71.53 \\

\midrule[1.0pt]

\multirow{2}{*}{\textbf{Model}}
& \multicolumn{3}{c}{\textbf{LCSTS-zh}}
& \multicolumn{3}{c}{\textbf{MLC-zh}}
& \multicolumn{3}{c}{\textbf{Wildchat-zh}}
& \multicolumn{3}{c}{\textbf{Avg}} \\

\cmidrule(lr){2-4}
\cmidrule(lr){5-7}
\cmidrule(lr){8-10}
\cmidrule(lr){11-13}

& \textbf{CER}$\downarrow$ & \textbf{UTMOS}$\uparrow$ & \textbf{G-Eval}$\uparrow$
& \textbf{CER}$\downarrow$ & \textbf{UTMOS}$\uparrow$ & \textbf{G-Eval}$\uparrow$
& \textbf{CER}$\downarrow$ & \textbf{UTMOS}$\uparrow$ & \textbf{G-Eval}$\uparrow$
& \textbf{CER}$\downarrow$ & \textbf{UTMOS}$\uparrow$ & \textbf{G-Eval}$\uparrow$ \\

\midrule[0.8pt]

Qwen2.5-Omni
& 1.86\% & 3.04 & 69.75
& 1.31\% & 3.07 & 82.53
& 1.37\% & 2.88 & 78.35
& 2.93\% & 2.95 & 74.28 \\

Kimi-Audio
& 2.42\% & 2.55 & 76.13
& 0.75\% & 2.83 & 80.92
& 1.96\% & 2.55 & 78.08
& 4.88\% & 2.52 & 71.93 \\

GLM-4-Voice
& 1.80\% & 2.92 & 76.36
& 1.74\% & 3.24 & 78.85
& 2.78\% & 2.86 & 84.41
& 3.77\% & 2.94 & 71.85 \\

\midrule[0.8pt]

Step-Audio 2 mini
& 1.28\% & 3.64 & 79.16
& 1.18\% & 3.85 & 78.16
& 9.39\% & 3.48 & 83.59
& 7.49\% & 3.62 & 69.24 \\

Vanilla-FT-ZH
& 0.91\% & 3.71 & 78.82
& 1.23\% & 3.99 & 77.47
& 2.03\% & 3.70 & 81.47
& 4.96\% & 3.77 & 69.88 \\

Vanilla-FT-MIX
& 3.83\% & 3.67 & 79.10
& 0.80\% & 3.98 & 79.31
& 2.44\% & 3.73 & 82.10
& 7.02\% & 3.77 & 70.26 \\

\noalign{\vskip\aboverulesep}
\cdashline{1-13}
\noalign{\vskip\belowrulesep}


CoSTA-ZH
& 1.18\% & 3.70 & 80.56
& 1.06\% & 3.96 & 79.54
& 0.90\% & 3.71 & 81.09
& 1.51\% & 3.75 & 72.27 \\

CoSTA-MIX
& 1.16\% & 3.70 & 78.99
& 0.77\% & 3.96 & 77.24
& 1.15\% & 3.65 & 82.39
& 1.65\% & 3.75 & 71.53 \\

\bottomrule[1.2pt]
\end{tabular}
}

\caption{Detailed quality results on Chinese speech-to-speech benchmarks. Each benchmark reports CER, UTMOS, and G-Eval for text-speech alignment, speech quality, and response-content quality, respectively.}
\label{tab:main-results-zh-full}
\end{table*}

%% file: tabs/quality_all_en.tex
\begin{table*}[t]
\centering
\normalsize
\renewcommand\arraystretch{1.12}
\setlength{\tabcolsep}{1.2mm}

\resizebox{\textwidth}{!}{
\begin{tabular}{l ccc !{\vrule} ccc !{\vrule} ccc !{\vrule} ccc}
\toprule[1.2pt]

\multirow{2}{*}{\textbf{Model}}
& \multicolumn{3}{c}{\textbf{AlpacaEval}}
& \multicolumn{3}{c}{\textbf{Gsm8kEval}}
& \multicolumn{3}{c}{\textbf{CommonEval}}
& \multicolumn{3}{c}{\textbf{Avg}} \\

\cmidrule(lr){2-4}
\cmidrule(lr){5-7}
\cmidrule(lr){8-10}
\cmidrule(lr){11-13}

& \textbf{WER}$\downarrow$ & \textbf{UTMOS}$\uparrow$ & \textbf{G-Eval}$\uparrow$
& \textbf{WER}$\downarrow$ & \textbf{UTMOS}$\uparrow$ & \textbf{G-Eval}$\uparrow$
& \textbf{WER}$\downarrow$ & \textbf{UTMOS}$\uparrow$ & \textbf{G-Eval}$\uparrow$
& \textbf{WER}$\downarrow$ & \textbf{UTMOS}$\uparrow$ & \textbf{G-Eval}$\uparrow$ \\

\midrule[0.8pt]

Qwen2.5-Omni
& 2.05\% & 4.22 & 76.78
& 4.25\% & 4.32 & 54.30
& 1.48\% & 4.27 & 72.00
& 1.73\% & 4.28 & 72.88 \\

Kimi-Audio
& 20.77\% & 2.53 & 62.01
& 99.08\% & 2.77 & 0.57
& 37.62\% & 2.95 & 50.77
& 34.92\% & 2.86 & 50.46 \\

GLM-4-Voice
& 13.92\% & 4.00 & 80.87
& 8.29\% & 4.25 & 29.32
& 26.28\% & 4.10 & 66.77
& 11.55\% & 4.16 & 69.95 \\

\midrule[0.8pt]

Step-Audio 2 mini
& 9.82\% & 4.51 & 76.31
& 17.07\% & 4.53 & 15.92
& 1.64\% & 4.53 & 52.53
& 6.49\% & 4.53 & 63.73 \\

Vanilla-FT-EN
& 23.82\% & 4.35 & 73.74
& 4.15\% & 4.53 & 16.67
& 9.73\% & 4.47 & 72.03
& 10.16\% & 4.46 & 66.59 \\

Vanilla-FT-MIX
& 27.61\% & 4.43 & 76.18
& 3.63\% & 4.54 & 11.05
& 9.89\% & 4.51 & 72.63
& 12.00\% & 4.49 & 65.45 \\

\noalign{\vskip\aboverulesep}
\cdashline{1-13}
\noalign{\vskip\belowrulesep}

CoSTA-EN
& 3.08\% & 4.54 & 81.34
& 3.01\% & 4.54 & 15.52
& 2.80\% & 4.55 & 75.87
& 2.50\% & 4.54 & 69.97 \\

CoSTA-MIX
& 3.19\% & 4.54 & 83.32
& 3.26\% & 4.55 & 17.53
& 2.61\% & 4.55 & 73.47
& 2.64\% & 4.54 & 69.67 \\

\midrule[1.0pt]

\multirow{2}{*}{\textbf{Model}}
& \multicolumn{3}{c}{\textbf{Summary}}
& \multicolumn{3}{c}{\textbf{MLC}}
& \multicolumn{3}{c}{\textbf{WildchatEval}}
& \multicolumn{3}{c}{\textbf{Avg}} \\

\cmidrule(lr){2-4}
\cmidrule(lr){5-7}
\cmidrule(lr){8-10}
\cmidrule(lr){11-13}

& \textbf{WER}$\downarrow$ & \textbf{UTMOS}$\uparrow$ & \textbf{G-Eval}$\uparrow$
& \textbf{WER}$\downarrow$ & \textbf{UTMOS}$\uparrow$ & \textbf{G-Eval}$\uparrow$
& \textbf{WER}$\downarrow$ & \textbf{UTMOS}$\uparrow$ & \textbf{G-Eval}$\uparrow$
& \textbf{WER}$\downarrow$ & \textbf{UTMOS}$\uparrow$ & \textbf{G-Eval}$\uparrow$ \\

\midrule[0.8pt]

Qwen2.5-Omni
& 0.38\% & 4.34 & 80.06
& 1.02\% & 4.30 & 77.40
& 1.20\% & 4.24 & 76.71
& 1.73\% & 4.28 & 72.88 \\

Kimi-Audio
& 3.43\% & 3.14 & 82.15
& 27.72\% & 3.17 & 40.87
& 20.91\% & 2.58 & 66.40
& 34.92\% & 2.86 & 50.46 \\

GLM-4-Voice
& 1.39\% & 4.26 & 92.15
& 10.79\% & 4.23 & 69.30
& 8.61\% & 4.11 & 81.26
& 11.55\% & 4.16 & 69.95 \\

\midrule[0.8pt]

Step-Audio 2 mini
& 0.28\% & 4.55 & 88.14
& 3.23\% & 4.53 & 72.69
& 6.86\% & 4.52 & 76.75
& 6.49\% & 4.53 & 63.73 \\

Vanilla-FT-EN
& 4.18\% & 4.51 & 86.10
& 3.63\% & 4.53 & 72.50
& 15.45\% & 4.41 & 78.47
& 10.16\% & 4.46 & 66.59 \\

Vanilla-FT-MIX
& 10.09\% & 4.49 & 85.37
& 3.96\% & 4.52 & 68.55
& 16.85\% & 4.46 & 78.93
& 12.00\% & 4.49 & 65.45 \\

\noalign{\vskip\aboverulesep}
\cdashline{1-13}
\noalign{\vskip\belowrulesep}

CoSTA-EN
& 1.34\% & 4.55 & 89.21
& 2.00\% & 4.53 & 74.01
& 2.76\% & 4.54 & 83.84
& 2.50\% & 4.54 & 69.97 \\

CoSTA-MIX
& 1.25\% & 4.55 & 90.00
& 2.07\% & 4.53 & 71.19
& 3.45\% & 4.54 & 82.52
& 2.64\% & 4.54 & 69.67 \\

\bottomrule[1.2pt]
\end{tabular}
}

\caption{Detailed quality results on English speech-to-speech benchmarks. Each benchmark reports WER, UTMOS, and G-Eval for text-speech alignment, speech quality, and response-content quality, respectively.}
\label{tab:main-results-en-full}
\end{table*}

%% file: tabs/scaling_tts_in_50k_s2s.tex
\begin{table*}[t]
\centering
\normalsize
\renewcommand\arraystretch{1.12}
\setlength{\tabcolsep}{1.2mm}

\resizebox{\textwidth}{!}{
\begin{tabular}{l c !{\vrule} cccc !{\vrule} cccc !{\vrule} cccc}
\toprule[1.2pt]

\multirow{2}{*}{\textbf{Model}}
& \multirow{2}{*}{\textbf{TTS Data}}
& \multicolumn{4}{c}{\textbf{AlpacaEval-zh}}
& \multicolumn{4}{c}{\textbf{APE-zh}}
& \multicolumn{4}{c}{\textbf{Avg}} \\

\cmidrule(lr){3-6}
\cmidrule(lr){7-10}
\cmidrule(lr){11-14}

&
& $\boldsymbol{\rho_g}\downarrow$ & \textbf{ASR-CER}$\downarrow$ & \textbf{UTMOS}$\uparrow$ & \textbf{G-Eval}$\uparrow$
& $\boldsymbol{\rho_g}\downarrow$ & \textbf{ASR-CER}$\downarrow$ & \textbf{UTMOS}$\uparrow$ & \textbf{G-Eval}$\uparrow$
& $\boldsymbol{\rho_g}\downarrow$ & \textbf{ASR-CER}$\downarrow$ & \textbf{UTMOS}$\uparrow$ & \textbf{G-Eval}$\uparrow$ \\

\midrule[0.8pt]

Step-Audio 2 mini
& --
& 100.00\% & 8.48\% & 3.54 & 79.95
& 100.00\% & 4.28\% & 3.87 & 22.11
& 100.00\% & 8.57\% & 3.66 & 63.34 \\

\noalign{\vskip\aboverulesep}
\cdashline{1-14}
\noalign{\vskip\belowrulesep}

\multirow{5}{*}{CoSTA-ZH}
& 500h
& 68.17\% & 1.62\% & 3.70 & 80.86
& 65.32\% & 1.11\% & 4.06 & 21.05
& \textbf{66.48\%} & 1.52\% & 3.79 & 66.35 \\

& 400h
& 68.73\% & 1.69\% & 3.70 & 82.40
& 65.88\% & 1.48\% & 4.05 & 25.26
& 67.35\% & 1.62\% & 3.79 & 67.71 \\

& 300h
& 68.54\% & 1.91\% & 3.72 & 81.59
& 65.09\% & 1.02\% & 4.06 & 25.09
& \underline{66.83\%} & 1.66\% & 3.78 & \underline{67.75} \\

& 200h
& 69.29\% & 2.15\% & 3.68 & 82.36
& 65.69\% & 0.83\% & 4.07 & 26.84
& 67.24\% & 1.71\% & 3.78 & \textbf{67.99} \\

& 100h
& 68.78\% & 1.75\% & 3.73 & 82.86
& 65.50\% & 1.55\% & 4.03 & 22.11
& 67.39\% & 1.80\% & 3.78 & 66.88 \\

\midrule[1.0pt]

\multirow{2}{*}{\textbf{Model}}
& \multirow{2}{*}{\textbf{TTS Data}}
& \multicolumn{4}{c}{\textbf{Claude-zh}}
& \multicolumn{4}{c}{\textbf{MLC-zh}}
& \multicolumn{4}{c}{\textbf{Avg}} \\

\cmidrule(lr){3-6}
\cmidrule(lr){7-10}
\cmidrule(lr){11-14}

&
& $\boldsymbol{\rho_g}\downarrow$ & \textbf{ASR-CER}$\downarrow$ & \textbf{UTMOS}$\uparrow$ & \textbf{G-Eval}$\uparrow$
& $\boldsymbol{\rho_g}\downarrow$ & \textbf{ASR-CER}$\downarrow$ & \textbf{UTMOS}$\uparrow$ & \textbf{G-Eval}$\uparrow$
& $\boldsymbol{\rho_g}\downarrow$ & \textbf{ASR-CER}$\downarrow$ & \textbf{UTMOS}$\uparrow$ & \textbf{G-Eval}$\uparrow$ \\

\midrule[0.8pt]

Step-Audio 2 mini
& --
& 100.00\% & 20.21\% & 3.36 & 73.15
& 100.00\% & 1.30\% & 3.85 & 78.16
& 100.00\% & 8.57\% & 3.66 & 63.34 \\

\noalign{\vskip\aboverulesep}
\cdashline{1-14}
\noalign{\vskip\belowrulesep}

\multirow{5}{*}{CoSTA-ZH}
& 500h
& 69.35\% & 2.39\% & 3.41 & 84.86
& 63.09\% & 0.95\% & 3.99 & 78.62
& \textbf{66.48\%} & 1.52\% & 3.79 & 66.35 \\

& 400h
& 70.07\% & 2.42\% & 3.42 & 85.47
& 64.70\% & 0.88\% & 3.99 & 77.70
& 67.35\% & 1.62\% & 3.79 & 67.71 \\

& 300h
& 69.77\% & 2.98\% & 3.39 & 84.77
& 63.93\% & 0.71\% & 3.93 & 79.54
& \underline{66.83\%} & 1.66\% & 3.78 & \underline{67.75} \\

& 200h
& 70.12\% & 2.84\% & 3.40 & 85.53
& 63.85\% & 1.02\% & 3.97 & 77.24
& 67.24\% & 1.71\% & 3.78 & \textbf{67.99} \\

& 100h
& 70.38\% & 2.69\% & 3.39 & 85.77
& 64.90\% & 1.22\% & 3.95 & 76.78
& 67.39\% & 1.80\% & 3.78 & 66.88 \\

\bottomrule[1.2pt]
\end{tabular}
}

\caption{Effect of pronunciation-adaptation data scale on Chinese benchmarks. Dialogue-adaptation data is fixed to 50K examples. $\rho_g$ denotes the generated-token compression rate, while ASR-CER, UTMOS, and G-Eval evaluate text-speech alignment, speech quality, and response-content quality, respectively.}
\label{tab:tts-data-scale}
\end{table*}

%% file: tabs/scaling_s2s_in_300h_tts.tex
\begin{table*}[t]
\centering
\normalsize
\renewcommand\arraystretch{1.12}
\setlength{\tabcolsep}{1.2mm}

\resizebox{\textwidth}{!}{
\begin{tabular}{l c !{\vrule} cccc !{\vrule} cccc !{\vrule} cccc}
\toprule[1.2pt]

\multirow{2}{*}{\textbf{Model}}
& \multirow{2}{*}{\textbf{Dialogue Data}}
& \multicolumn{4}{c}{\textbf{AlpacaEval-zh}}
& \multicolumn{4}{c}{\textbf{APE-zh}}
& \multicolumn{4}{c}{\textbf{Avg}} \\

\cmidrule(lr){3-6}
\cmidrule(lr){7-10}
\cmidrule(lr){11-14}

&
& $\boldsymbol{\rho_g}\downarrow$ & \textbf{ASR-CER}$\downarrow$ & \textbf{UTMOS}$\uparrow$ & \textbf{G-Eval}$\uparrow$
& $\boldsymbol{\rho_g}\downarrow$ & \textbf{ASR-CER}$\downarrow$ & \textbf{UTMOS}$\uparrow$ & \textbf{G-Eval}$\uparrow$
& $\boldsymbol{\rho_g}\downarrow$ & \textbf{ASR-CER}$\downarrow$ & \textbf{UTMOS}$\uparrow$ & \textbf{G-Eval}$\uparrow$ \\

\midrule[0.8pt]

Step-Audio 2 mini
& --
& 100.00\% & 8.48\% & 3.54 & 79.95
& 100.00\% & 4.28\% & 3.87 & 22.11
& 100.00\% & 8.57\% & 3.66 & 63.34 \\

\noalign{\vskip\aboverulesep}
\cdashline{1-14}
\noalign{\vskip\belowrulesep}

\multirow{5}{*}{CoSTA-ZH}
& 50K
& 68.54\% & 1.91\% & 3.72 & 81.59
& 65.09\% & 1.02\% & 4.06 & 25.09
& \textbf{66.83\%} & 1.66\% & 3.78 & \textbf{67.75} \\

& 40K
& 69.24\% & 1.79\% & 3.71 & 78.59
& 65.61\% & 1.36\% & 4.04 & 22.81
& \underline{67.28\%} & 1.72\% & 3.78 & 66.02 \\

& 30K
& 68.82\% & 1.90\% & 3.73 & 79.23
& 66.04\% & 1.00\% & 4.04 & 18.07
& 67.50\% & 1.80\% & 3.78 & 64.95 \\

& 20K
& 68.32\% & 2.02\% & 3.70 & 80.45
& 66.20\% & 0.95\% & 4.06 & 25.44
& 67.49\% & 1.97\% & 3.79 & 66.84 \\

& 10K
& 68.82\% & 1.31\% & 3.75 & 82.59
& 68.10\% & 0.67\% & 4.02 & 24.56
& 68.19\% & 2.09\% & 3.80 & \underline{67.25} \\

\midrule[1.0pt]

\multirow{2}{*}{\textbf{Model}}
& \multirow{2}{*}{\textbf{Dialogue Data}}
& \multicolumn{4}{c}{\textbf{Claude-zh}}
& \multicolumn{4}{c}{\textbf{MLC-zh}}
& \multicolumn{4}{c}{\textbf{Avg}} \\

\cmidrule(lr){3-6}
\cmidrule(lr){7-10}
\cmidrule(lr){11-14}

&
& $\boldsymbol{\rho_g}\downarrow$ & \textbf{ASR-CER}$\downarrow$ & \textbf{UTMOS}$\uparrow$ & \textbf{G-Eval}$\uparrow$
& $\boldsymbol{\rho_g}\downarrow$ & \textbf{ASR-CER}$\downarrow$ & \textbf{UTMOS}$\uparrow$ & \textbf{G-Eval}$\uparrow$
& $\boldsymbol{\rho_g}\downarrow$ & \textbf{ASR-CER}$\downarrow$ & \textbf{UTMOS}$\uparrow$ & \textbf{G-Eval}$\uparrow$ \\

\midrule[0.8pt]

Step-Audio 2 mini
& --
& 100.00\% & 20.21\% & 3.36 & 73.15
& 100.00\% & 1.30\% & 3.85 & 78.16
& 100.00\% & 8.57\% & 3.66 & 63.34 \\

\noalign{\vskip\aboverulesep}
\cdashline{1-14}
\noalign{\vskip\belowrulesep}

\multirow{5}{*}{CoSTA-ZH}
& 50K
& 69.77\% & 2.98\% & 3.39 & 84.77
& 63.93\% & 0.71\% & 3.93 & 79.54
& \textbf{66.83\%} & 1.66\% & 3.78 & \textbf{67.75} \\

& 40K
& 69.99\% & 2.68\% & 3.41 & 86.13
& 64.28\% & 1.07\% & 3.97 & 76.55
& \underline{67.28\%} & 1.72\% & 3.78 & 66.02 \\

& 30K
& 70.25\% & 2.90\% & 3.38 & 84.80
& 64.88\% & 1.39\% & 3.98 & 77.70
& 67.50\% & 1.80\% & 3.78 & 64.95 \\

& 20K
& 70.19\% & 2.93\% & 3.45 & 86.05
& 65.25\% & 1.97\% & 3.95 & 75.40
& 67.49\% & 1.97\% & 3.79 & 66.84 \\

& 10K
& 70.30\% & 2.50\% & 3.50 & 84.59
& 65.55\% & 3.86\% & 3.95 & 77.24
& 68.19\% & 2.09\% & 3.80 & \underline{67.25} \\

\bottomrule[1.2pt]
\end{tabular}
}

\caption{Effect of dialogue-adaptation data scale on Chinese benchmarks. Pronunciation-adaptation data is fixed to 300h. $\rho_g$ denotes the generated-token compression rate, while ASR-CER, UTMOS, and G-Eval evaluate text-speech alignment, speech quality, and response-content quality, respectively.}
\label{tab:dialogue-data-scale}
\end{table*}

%% file: tabs/latency.tex
\begin{table*}[t]
\centering
\small
\renewcommand{\arraystretch}{1.08}
\setlength{\tabcolsep}{4.0pt}

\resizebox{\textwidth}{!}{
\begin{tabular}{
    l
    c
    c
    !{\hspace{3pt}\vrule width 0.6pt\hspace{3pt}}
    c
    c
    c
    c
    c
}
\toprule
\textbf{Model}
& $\boldsymbol{\rho_g}\downarrow$
& \textbf{Speed}$\uparrow$
& \textbf{Prefill (ms)}$\downarrow$
& \textbf{Decode (ms)}$\downarrow$
& \textbf{LLM Total (ms)}$\downarrow$
& \textbf{Speech Decoder (ms)}$\downarrow$
& \textbf{Total (ms)}$\downarrow$ \\
\midrule

Step-Audio 2 mini
& 100.00\%
& 58.74
& 56.02
& 164.42
& 220.44
& 337.46
& 557.90 \\

Vanilla-FT-ZH
& 100.00\%
& 58.56
& 55.37
& 164.22
& 219.59
& 340.51
& 560.10 \\

\textbf{CoSTA-ZH}
& \textbf{66.81\%}
& \textbf{101.99}
& \textbf{54.55}
& \textbf{122.98}
& \textbf{177.53}
& \textbf{328.06}
& \textbf{505.59} \\

\midrule

Step-Audio 2 mini
& 100.00\%
& 59.69
& \textbf{53.07}
& 165.08
& 218.15
& 344.44
& 562.60 \\

Vanilla-FT-EN
& 100.00\%
& 57.67
& 55.72
& 165.10
& 220.82
& 339.52
& 560.34 \\

\textbf{CoSTA-EN}
& \textbf{70.17\%}
& \textbf{99.05}
& 53.24
& \textbf{115.96}
& \textbf{169.20}
& \textbf{339.17}
& \textbf{508.36} \\

\bottomrule
\end{tabular}
}

\caption{
First-packet latency breakdown for Chinese and English settings.
The upper and lower blocks report Chinese and English results, respectively.
}
\label{tab:first-packet-latency-breakdown}
\end{table*}

%% file: tabs/human_zh.tex
\begin{table*}[t]
\centering
\small
\renewcommand{\arraystretch}{1.12}
\setlength{\tabcolsep}{1.2mm}

\resizebox{\textwidth}{!}{
\begin{tabular}{ll cccccc !{\vrule} c}
\toprule[1.2pt]
\textbf{Metric}
& \textbf{Model}
& \textbf{AlpacaEval-zh}
& \textbf{APE-zh}
& \textbf{Claude-zh}
& \textbf{LCSTS-zh}
& \textbf{MLC-zh}
& \textbf{Wildchat-zh}
& \textbf{Avg.} \\
\midrule[0.8pt]

\multirow{3}{*}{\shortstack{Speech\\Fluency}}
& Step-Audio 2 mini
& 4.70 & \textbf{4.77} & 3.77 & 4.47 & \textbf{4.83} & 4.37 & 4.48 \\
& Vanilla-FT-ZH
& 3.80 & 4.50 & 3.60 & \textbf{4.83} & 4.60 & \textbf{4.77} & 4.35 \\
& \textbf{CoSTA-ZH}
& \textbf{4.73} & 4.33 & \textbf{4.60}
& 4.67 & 4.70 & 4.73 & \textbf{4.63} \\

\midrule[0.8pt]

\multirow{3}{*}{\shortstack{Semantic\\Quality}}
& Step-Audio 2 mini
& 4.17 & \textbf{2.23} & 3.17 & 3.60 & 3.83 & 4.10 & 3.52 \\
& Vanilla-FT-ZH
& 4.00 & 1.83 & 3.43 & 3.80 & 3.70 & \textbf{4.23} & 3.50 \\
& \textbf{CoSTA-ZH}
& \textbf{4.40} & 2.17 & \textbf{3.60}
& \textbf{4.10} & \textbf{4.00} & \textbf{4.23} & \textbf{3.75} \\

\bottomrule[1.2pt]
\end{tabular}
}

\caption{MOS-style human evaluation on Chinese URO-Bench datasets.
Scores are averaged over six evaluators and five randomly sampled queries
per dataset. Higher scores indicate better performance.}
\label{tab:human-evaluation-zh}
\end{table*}

%% file: tabs/human_en.tex
\begin{table*}[t]
\centering
\small
\renewcommand{\arraystretch}{1.12}
\setlength{\tabcolsep}{1.2mm}

\resizebox{\textwidth}{!}{
\begin{tabular}{ll cccccc !{\vrule} c}
\toprule[1.2pt]
\textbf{Metric}
& \textbf{Model}
& \textbf{AlpacaEval}
& \textbf{Gsm8kEval}
& \textbf{CommonEval}
& \textbf{Summary}
& \textbf{MLC}
& \textbf{WildchatEval}
& \textbf{Avg.} \\
\midrule[0.8pt]

\multirow{3}{*}{\shortstack{Speech\\Fluency}}
& Step-Audio 2 mini
& 4.23 & 4.03 & 4.73 & 4.70 & 4.50 & \textbf{4.80} & 4.50 \\
& Vanilla-FT-EN
& 3.53 & 4.63 & 4.23 & 4.77 & 4.30 & 4.37 & 4.31 \\
& \textbf{CoSTA-EN}
& \textbf{4.80} & \textbf{4.70} & \textbf{4.77}
& \textbf{4.87} & \textbf{4.83} & 4.73 & \textbf{4.78} \\

\midrule[0.8pt]

\multirow{3}{*}{\shortstack{Semantic\\Quality}}
& Step-Audio 2 mini
& 3.50 & \textbf{2.33} & 1.23 & 4.57 & 4.20 & 4.27 & 3.35 \\
& Vanilla-FT-EN
& 3.13 & 1.77 & 4.00 & 4.40 & 4.80 & 4.27 & 3.73 \\
& \textbf{CoSTA-EN}
& \textbf{4.37} & 1.20 & \textbf{4.17}
& \textbf{4.90} & \textbf{4.83} & \textbf{4.73} & \textbf{4.03} \\

\bottomrule[1.2pt]
\end{tabular}
}

\caption{MOS-style human evaluation on English URO-Bench datasets.
Scores are averaged over six evaluators and five randomly sampled queries
per dataset. Higher scores indicate better performance.}
\label{tab:human-evaluation-en}
\end{table*}

%% file: tabs/opens2s.tex
\begin{table*}[t]
\centering
\normalsize
\renewcommand\arraystretch{1.12}
\setlength{\tabcolsep}{1.2mm}

\resizebox{\textwidth}{!}{
\begin{tabular}{l cccc !{\vrule} cccc !{\vrule} cccc !{\vrule} cccc}
\toprule[1.2pt]

\multirow{2}{*}{\textbf{Model}}
& \multicolumn{4}{c}{\textbf{AlpacaEval-zh}}
& \multicolumn{4}{c}{\textbf{Claude-zh}}
& \multicolumn{4}{c}{\textbf{LCSTS-zh}}
& \multicolumn{4}{c}{\textbf{Avg}} \\

\cmidrule(lr){2-5}
\cmidrule(lr){6-9}
\cmidrule(lr){10-13}
\cmidrule(lr){14-17}

& $\boldsymbol{\rho_g}\downarrow$ & \textbf{ASR-CER}$\downarrow$ & \textbf{UTMOS}$\uparrow$ & \textbf{G-Eval}$\uparrow$
& $\boldsymbol{\rho_g}\downarrow$ & \textbf{ASR-CER}$\downarrow$ & \textbf{UTMOS}$\uparrow$ & \textbf{G-Eval}$\uparrow$
& $\boldsymbol{\rho_g}\downarrow$ & \textbf{ASR-CER}$\downarrow$ & \textbf{UTMOS}$\uparrow$ & \textbf{G-Eval}$\uparrow$
& $\boldsymbol{\rho_g}\downarrow$ & \textbf{ASR-CER}$\downarrow$ & \textbf{UTMOS}$\uparrow$ & \textbf{G-Eval}$\uparrow$ \\

\midrule[0.8pt]

OpenS2S
& 100.00\% & 9.02\% & \textbf{2.54} & 69.71
& 100.00\% & 14.40\% & \textbf{2.18} & 66.97
& 100.00\% & 9.82\% & \textbf{2.51} & 59.10
& 100.00\% & 11.08\% & \textbf{2.41} & 65.26 \\

Vanilla-FT-ZH
& 100.00\% & 6.26\% & 2.49 & \textbf{72.29}
& 100.00\% & 20.00\% & 2.14 & 66.64
& 100.00\% & 7.81\% & 2.50 & 63.25
& 100.00\% & 11.36\% & 2.38 & 67.39 \\

\textbf{CoSTA-ZH}
& \textbf{78.34\%} & \textbf{5.98\%} & 2.52 & 70.61
& \textbf{79.92\%} & \textbf{11.77\%} & 2.08 & \textbf{71.65}
& \textbf{78.23\%} & \textbf{7.75\%} & 2.49 & \textbf{65.27}
& \textbf{78.83\%} & \textbf{8.50\%} & 2.36 & \textbf{69.18} \\

\midrule[1.0pt]

\multirow{2}{*}{\textbf{Model}}
& \multicolumn{4}{c}{\textbf{AlpacaEval}}
& \multicolumn{4}{c}{\textbf{CommonEval}}
& \multicolumn{4}{c}{\textbf{Summary}}
& \multicolumn{4}{c}{\textbf{Avg}} \\

\cmidrule(lr){2-5}
\cmidrule(lr){6-9}
\cmidrule(lr){10-13}
\cmidrule(lr){14-17}

& $\boldsymbol{\rho_g}\downarrow$ & \textbf{ASR-WER}$\downarrow$ & \textbf{UTMOS}$\uparrow$ & \textbf{G-Eval}$\uparrow$
& $\boldsymbol{\rho_g}\downarrow$ & \textbf{ASR-WER}$\downarrow$ & \textbf{UTMOS}$\uparrow$ & \textbf{G-Eval}$\uparrow$
& $\boldsymbol{\rho_g}\downarrow$ & \textbf{ASR-WER}$\downarrow$ & \textbf{UTMOS}$\uparrow$ & \textbf{G-Eval}$\uparrow$
& $\boldsymbol{\rho_g}\downarrow$ & \textbf{ASR-WER}$\downarrow$ & \textbf{UTMOS}$\uparrow$ & \textbf{G-Eval}$\uparrow$ \\

\midrule[0.8pt]

OpenS2S
& 100.00\% & 6.96\% & \textbf{3.72} & 70.65
& 100.00\% & 4.92\% & \textbf{3.85} & 63.47
& 100.00\% & \textbf{2.07\%} & \textbf{3.87} & 87.80
& 100.00\% & \textbf{4.65\%} & \textbf{3.81} & 73.97 \\

Vanilla-FT-EN
& 100.00\% & 8.44\% & 3.67 & 72.16
& 100.00\% & \textbf{4.67\%} & 3.84 & 66.27
& 100.00\% & 2.36\% & 3.86 & \textbf{88.02}
& 100.00\% & 5.16\% & 3.79 & 75.48 \\

\textbf{CoSTA-EN}
& \textbf{82.37\%} & \textbf{6.35\%} & 3.65 & \textbf{74.47}
& \textbf{80.87\%} & 5.18\% & 3.84 & \textbf{66.53}
& \textbf{80.83\%} & 2.47\% & 3.85 & 87.68
& \textbf{81.36\%} & 4.66\% & 3.78 & \textbf{76.23} \\

\bottomrule[1.2pt]
\end{tabular}
}

\caption{Detailed SDM-level generalization results on OpenS2S. 
CoSTA is compared with the original OpenS2S model and matched vanilla fine-tuning on Chinese and English benchmarks.
$\rho_g$ denotes the generated-token compression rate, while ASR-CER (Chinese) / ASR-WER (English), UTMOS, and G-Eval evaluate text-speech alignment, speech quality, and response-content quality, respectively.}
\label{tab:opens2s-generality-detail}
\end{table*}

%% file: tabs/dialogue_token_usage.tex
\begin{table*}[t]
\centering
\small
\setlength{\tabcolsep}{5pt}
\renewcommand{\arraystretch}{1.1}
\textbf{(a) Token usage and BPE coverage}\par\smallskip
\begin{tabular}{lccccc}
\toprule
\textbf{Model} & \textbf{Original} & \textbf{BPE} & \textbf{Original/BPE} & \textbf{BPE coverage} & \textbf{Avg. expansion} \\
\midrule
CoSTA-ZH & 70.95\% & 29.05\% & 2.44 & 51.64\% & 2.608 \\
w/o Dialogue Adaptation & 78.95\% & 21.05\% & 3.75 & 36.71\% & 2.175 \\
\bottomrule
\end{tabular}
\par\medskip
\textbf{(b) Distribution of token expansion lengths}\par\smallskip
\begin{tabular}{lcccccc}
\toprule
\textbf{Model} & $L=1$ & $L=2$ & $L=3$ & $L=4$ & $5\leq L\leq 8$ & $L\geq 9$ \\
\midrule
CoSTA-ZH & 70.95\% & 24.28\% & 1.69\% & 0.79\% & 1.36\% & 0.93\% \\
w/o Dialogue Adaptation & 78.95\% & 19.22\% & 1.12\% & 0.27\% & 0.40\% & 0.05\% \\
\bottomrule
\end{tabular}
\caption{Speech-token usage with and without dialogue adaptation on all six Chinese benchmarks (1,122 responses per model). Statistics are pooled over generated speech tokens. (a) Original/BPE shares and their ratio use counts before expansion; BPE coverage uses counts after expansion. Avg. expansion is the number of original tokens per generated BPE unit. (b) $L$ is the expansion length, with $L=1$ denoting an original token. Each percentage uses all generated speech tokens before expansion as its denominator.}
\label{tab:dialogue-token-usage}
\end{table*}

%% file: tabs/vocab_size_ablation.tex
\begin{table*}[t]
\centering
\small
\renewcommand\arraystretch{1.0}
\setlength{\tabcolsep}{5pt}

\begin{tabular}{l c c c c c c}
\toprule[1.2pt]

\textbf{Model}
& \textbf{Size}
& $\boldsymbol{\rho_g}\downarrow$
& \textbf{Speed}$\uparrow$
& \textbf{Relative}$\uparrow$
& \textbf{UTMOS}$\uparrow$
& \textbf{G-Eval}$\uparrow$ \\

\midrule[0.8pt]

Step-Audio 2 mini
& 0K
& 100.00\%
& 55.51
& 100.00\%
& 3.58
& 80.18 \\

\noalign{\vskip\aboverulesep}
\cdashline{1-7}
\noalign{\vskip\belowrulesep}

\multirow{5}{*}{CoSTA-ZH}
& 10K
& 76.66\%
& 78.67
& 141.71\%
& 3.73
& 80.86 \\

& 20K
& 74.33\%
& 84.51
& 152.24\%
& 3.71
& 80.73 \\

& 40K
& 71.25\%
& 93.20
& 167.89\%
& 3.69
& 83.35 \\

& 80K
& 67.50\%
& 103.03
& 185.59\%
& 3.70
& 81.36 \\

& 160K
& 63.62\%
& 104.18
& 187.65\%
& 3.68
& 78.87 \\

\bottomrule[1.2pt]
\end{tabular}

\caption{Complete vocabulary-size results on AlpacaEval-zh underlying Figure~\ref{fig:vocab-tradeoff}. Size is the number of added BPE units; Speed is expanded speech tokens/s, normalized to the original Step-Audio 2 mini in Relative. The 0K row is the unadapted baseline.}
\label{tab:bpe-vocab-full-decoding}
\end{table*}

%% file: tabs/cost.tex
\begin{table*}[t]
\centering
\small
\renewcommand{\arraystretch}{1.08}
\setlength{\tabcolsep}{7pt}

\begin{tabular}{ccccc}
\toprule
\textbf{Added units}
& \textbf{Compression}$\downarrow$
& \textbf{Median ms/step}$\downarrow$
& \textbf{Overhead}$\downarrow$
& \textbf{Norm. Cost}$\downarrow$ \\
\midrule

0K   & 100.00\% & 0.6902 & +0.00\%   & 1.000 \\

\noalign{\vskip\aboverulesep}
\cdashline{1-5}
\noalign{\vskip\belowrulesep}

10K  & 73.27\% & 0.7320 & +6.06\%   & 0.777 \\
20K  & 69.79\% & 0.7739 & +12.13\%  & 0.783 \\
40K  & 66.15\% & 0.8576 & +24.25\%  & 0.822 \\
80K  & 62.29\% & 1.0251 & +48.52\%  & 0.925 \\
160K & 58.48\% & 1.3598 & +97.01\%  & 1.152 \\
320K & 54.94\% & 2.0297 & +194.08\% & 1.616 \\

\bottomrule
\end{tabular}

\caption{
Complete corpus-compression and isolated LM-head projection plus greedy-argmax costs for 0K--320K added BPE units, measured on a single A100 GPU with identical BF16 hidden states at batch size 1. Compression uses held-out Dialogue-ZH token counts. Derived costs use unrounded compression rates. Overhead is per-step latency relative to 0K; Norm. Cost is $\rho_M t_M/t_0$ for a fixed expanded token count (Equation~\ref{eq:normalized-head-cost}).
}
\label{tab:lm-head-scaling}
\end{table*}

%% file: tabs/compression_tokenizer_fits.tex
\begin{table*}[t]
\centering\small
\renewcommand{\arraystretch}{1.02}
\setlength{\tabcolsep}{3pt}
\begin{tabular}{llrrrrr}
\toprule
Tokenizer & Split & $\alpha$ (\%) & $\beta$ (pp/doubling) & $R^2$ & RMSE (pp) & Max $|e|$ (pp) \\
\midrule
GLM-4-Voice Tokenizer 12.5Hz & Train & 84.884 & -4.808 & 0.9959 & 0.528 & 0.784 \\
 & Test & 84.866 & -4.193 & 0.9989 & 0.242 & 0.386 \\
\addlinespace[2pt]
S3Tokenizer V2 25Hz & Train & 73.550 & -4.013 & 0.9988 & 0.241 & 0.370 \\
 & Test & 73.397 & -3.698 & 0.9997 & 0.103 & 0.149 \\
\addlinespace[2pt]
S3Tokenizer V3 25Hz & Train & 71.678 & -4.107 & 0.9999 & 0.085 & 0.138 \\
 & Test & 71.569 & -3.809 & 0.9991 & 0.191 & 0.309 \\
\addlinespace[2pt]
Mimi & Train & 60.365 & -4.023 & 0.9968 & 0.389 & 0.565 \\
 & Test & 60.217 & -3.539 & 0.9852 & 0.740 & 1.077 \\
\addlinespace[2pt]
S3Tokenizer V1 25Hz & Train & 57.479 & -3.922 & 0.9939 & 0.526 & 0.741 \\
 & Test & 57.488 & -3.664 & 0.9864 & 0.734 & 1.001 \\
\addlinespace[2pt]
Qwen3-TTS-Tokenizer-12Hz & Train & 57.064 & -4.147 & 0.9947 & 0.516 & 0.766 \\
 & Test & 57.018 & -3.684 & 0.9807 & 0.883 & 1.212 \\
\addlinespace[2pt]
S3Tokenizer V1 50Hz & Train & 52.896 & -3.652 & 0.9920 & 0.559 & 0.774 \\
 & Test & 52.812 & -3.501 & 0.9874 & 0.675 & 0.928 \\
\addlinespace[2pt]
mHuBERT Base L11 KM1000 & Train & 38.288 & -2.878 & 0.9902 & 0.490 & 0.712 \\
 & Test & 38.425 & -2.699 & 0.9812 & 0.637 & 0.875 \\
\bottomrule
\end{tabular}
\caption{Complete training and test fits for the eight tokenizer configurations on Chinese dialogue speech. Full names, configurations, and public reference sources are given in Table~\ref{tab:tokenizer-configurations}. Fits use $y=\alpha+\beta\log_2(M/10\mathrm{K})$, where $y=100\rho$. Each row fits six vocabulary settings; errors are in percentage points.}
\label{tab:compression-tokenizer-fits}
\end{table*}

%% file: tabs/compression_dataset_fits.tex
\begin{table*}[t]
\centering\small
\renewcommand{\arraystretch}{1.02}
\setlength{\tabcolsep}{7pt}
\begin{tabular}{llrrrrr}
\toprule
Setting & Split & $\alpha$ (\%) & $\beta$ (pp/doubling) & $R^2$ & RMSE (pp) & Max $|e|$ (pp) \\
\midrule
Emilia-MIX & Train & 85.902 & -3.653 & 0.9929 & 0.529 & 0.752 \\
 & Test & 85.664 & -3.621 & 0.9942 & 0.474 & 0.670 \\
\addlinespace[2pt]
Emilia-EN & Train & 85.513 & -3.706 & 0.9942 & 0.485 & 0.693 \\
 & Test & 85.127 & -3.680 & 0.9957 & 0.414 & 0.605 \\
\addlinespace[2pt]
Emilia-ZH & Train & 84.656 & -3.751 & 0.9947 & 0.468 & 0.676 \\
 & Test & 84.464 & -3.681 & 0.9954 & 0.426 & 0.618 \\
\addlinespace[2pt]
Emilia-TC-MIX & Train & 83.130 & -3.768 & 0.9966 & 0.377 & 0.560 \\
 & Test & 82.889 & -3.694 & 0.9965 & 0.375 & 0.551 \\
\addlinespace[2pt]
Emilia-TC-EN & Train & 82.556 & -3.751 & 0.9973 & 0.334 & 0.496 \\
 & Test & 82.207 & -3.669 & 0.9972 & 0.334 & 0.481 \\
\addlinespace[2pt]
Emilia-TC-ZH & Train & 81.824 & -3.897 & 0.9977 & 0.322 & 0.494 \\
 & Test & 81.605 & -3.815 & 0.9977 & 0.316 & 0.460 \\
\addlinespace[2pt]
Dialogue-MIX & Train & 77.096 & -4.036 & 0.9991 & 0.211 & 0.336 \\
 & Test & 77.138 & -3.908 & 0.9996 & 0.136 & 0.212 \\
\addlinespace[2pt]
Dialogue-EN & Train & 76.790 & -4.242 & 0.9998 & 0.103 & 0.170 \\
 & Test & 76.621 & -4.012 & 0.9996 & 0.142 & 0.222 \\
\addlinespace[2pt]
Dialogue-ZH & Train & 73.550 & -4.013 & 0.9988 & 0.241 & 0.370 \\
 & Test & 73.400 & -3.699 & 0.9997 & 0.104 & 0.148 \\
\bottomrule
\end{tabular}
\caption{Complete training and test fits across data sources and languages with S3Tokenizer V2. Fits use $y=\alpha+\beta\log_2(M/10\mathrm{K})$, where $y=100\rho$. Each row fits six vocabulary settings; errors are in percentage points.}
\label{tab:compression-dataset-fits}
\end{table*}

%% file: tabs/case_study.tex

\providecommand{\fieldname}[1]{%
\vspace{0.10em}
\par\noindent\textbf{\footnotesize #1:}\hspace{0.25em}%
}

\providecommand{\tokenfield}[1]{%
\vspace{0.12em}
\par\noindent\textbf{\footnotesize #1:}\par
\vspace{0.02em}
}

\providecommand{\tokenseq}[1]{%
\vspace{0.02em} 
{\ttfamily\fontsize{7.5pt}{8.5pt}\selectfont 
\raggedright \setlength{\emergencystretch}{2em} #1\par} 
\vspace{0.10em}
}

\providecommand{\audtok}[1]{\texttt{[audio\_#1]}}

\clearpage
\onecolumn

\begingroup
\setlength{\parindent}{0pt}
\setlength{\parskip}{0.08em}
\small

\begin{tcolorbox}[
    enhanced,
    breakable,
    colback=gray!2,
    colframe=black!55,
    colbacktitle=black!80,
    coltitle=white,
    title=\textbf{Case Study 1: Short Factual QA},
    fonttitle=\bfseries,
    boxrule=0.80pt,
    arc=1.5mm,
    left=1.2mm,
    right=1.2mm,
    top=0.70mm,
    bottom=0.70mm,
    boxsep=0.70mm,
    before skip=0.35em,
    after skip=0.55em,
    pad at break*=0.8mm
]

\fieldname{Input}
\textit{What breed of dog is the smallest?}

\vspace{0.25em}

\begin{tcolorbox}[
    enhanced,
    breakable,
    colback=cyan!2,
    colframe=gray!35,
    colbacktitle=cyan!10,
    coltitle=black,
    title=\textbf{Step-Audio 2 mini},
    fonttitle=\bfseries\footnotesize,
    boxrule=0.30pt,
    arc=0.8mm,
    borderline west={1.6pt}{0pt}{cyan!55!black},
    left=0.9mm,
    right=0.9mm,
    top=0.45mm,
    bottom=0.45mm,
    boxsep=0.45mm,
    before skip=0.22em,
    after skip=0.22em,
    pad at break*=0.6mm
]

\fieldname{Text Content}
The Chihuahua is considered one of the smallest dog breeds, though there are other tiny dogs like the Yorkshire Terrier and Pomeranian that can be even smaller than some Chihuahuas.

\tokenfield{Speech Tokens}
\tokenseq{[1475, 1978, 4218, 4137, 2031, 2004, 72, 876, 2490, 5322, 3111, 3887,
2183, 1456, 710, 3404, 5591, 5026, 5830, 5092, 5023, 4699, 1490, 1763,
4769, 4450, 5644, 1623, 168, 12, 3908, 3910, 5645, 4673, 2217, 4526,
138, 3151, 3872, 1686, 1854, 2185, 2915, 5024, 3744, 443, 4275, 2085,
315, 65, 4672, 4916, 4915, 4914, 2124, 6501, 2849, 5597, 4958, 5074,
4909, 4492, 4411, 4753, 3459, 1950, 2112, 60, 2333, 2759, 2681, 4949,
5078, 1128, 2076, 2166, 2160, 4367, 1685, 962, 395, 557, 79, 1833,
6379, 6374, 5401, 27, 1458, 1707, 3648, 3645, 1701, 1704, 1947, 3891,
3894, 1950, 1959, 1914, 306, 2243, 2810, 2912, 5828, 1044, 4194, 4442,
2384, 3107, 4106, 1892, 1865, 4079, 1919, 2003, 1976, 6320, 4779, 543,
4528, 6059, 1687, 303, 2214, 6075, 5599, 5189, 2285, 3666, 2726, 2753,
1599, 66, 461, 494, 1223, 1220, 704, 1209, 81, 5482, 6372, 6475,
2702, 3002, 475, 2139, 63, 80, 1294, 3480, 3884, 5090, 5806, 4114,
732, 6048, 5319, 3141, 3875, 1680, 2247, 2274, 2922, 2366, 5294, 6293,
4115, 3158, 3401, 959, 1928, 1676, 1520, 737, 22, 3667, 2175, 1431,
5589, 5598, 5671, 2845, 4311, 2135, 2171, 2180, 1652, 4553, 5047, 2373,
2938, 2699, 539, 125, 5, 1476, 1480, 1723, 1716, 2040, 1272, 2216,
197, 1059, 2355, 25, 6178, 2172, 969, 4939, 2005, 2014, 5019, 4858,
4347, 2315, 3664, 1726, 6377, 5402, 4918, 1170, 4314, 662, 5678, 5025,
5062, 2750, 380, 1688, 884, 1281, 74, 20, 3913, 6376, 5648, 4671,
4619, 658, 4072, 1041, 4593, 2382, 1700, 2186, 1457, 575, 4133, 6077,
2894, 4372, 719, 515, 29, 794, 4428, 5726, 5645, 2484, 1702, 6405,
4218, 4218, 4218]}

\fieldname{Length}
\textbf{267} Speech Tokens 

\end{tcolorbox}

\vspace{0.8em}

\begin{tcolorbox}[
    enhanced,
    breakable,
    colback=orange!2,
    colframe=gray!35,
    colbacktitle=orange!10,
    coltitle=black,
    title=\textbf{Vanilla-FT-EN},
    fonttitle=\bfseries\footnotesize,
    boxrule=0.30pt,
    arc=0.8mm,
    borderline west={1.6pt}{0pt}{orange!50!black},
    left=0.9mm,
    right=0.9mm,
    top=0.45mm,
    bottom=0.45mm,
    boxsep=0.45mm,
    before skip=0.22em,
    after skip=0.22em,
    pad at break*=0.6mm
]

\fieldname{Text Content}
The smallest breed of dog is the Chihuahua.

\tokenfield{Speech Tokens}
\tokenseq{[3704, 4299, 6486, 6405, 4218, 1950, 1560, 801, 2246, 4428, 6377, 1035,
2130, 2125, 4949, 4945, 5029, 5059, 5985, 4409, 4733, 4444, 6454, 4191,
2232, 1188, 2166, 2169, 4367, 3872, 2420, 5102, 231, 70, 1760, 3404,
4868, 2894, 6534, 2112, 60, 2423, 5051, 4946, 494, 491, 515, 1125,
84, 168, 1707, 5838, 5109, 2922, 979, 1951, 386, 4841, 4454, 5409,
2085, 72, 2327, 150, 2733, 5326, 3111, 3806, 2102, 2185, 710, 6077,
6157, 5050, 3643, 707, 3161, 2918, 5832, 1701, 4299, 4299, 4299, 4299,
6486, 6486, 6486, 4299, 4299, 4299, 6486, 6486, 6405, 4218, 4218, 4218,
4299, 6159]}

\fieldname{Length}
\textbf{98} Speech Tokens 

\end{tcolorbox}

\vspace{0.8em}

\begin{tcolorbox}[
    enhanced,
    breakable,
    colback=red!2,
    colframe=gray!35,
    colbacktitle=red!10,
    coltitle=black,
    title=\textbf{CoSTA-EN},
    fonttitle=\bfseries\footnotesize,
    boxrule=0.30pt,
    arc=0.8mm,
    borderline west={1.6pt}{0pt}{red!45!black},
    left=0.9mm,
    right=0.9mm,
    top=0.45mm,
    bottom=0.45mm,
    boxsep=0.45mm,
    before skip=0.22em,
    after skip=0.22em,
    pad at break*=0.6mm
]

\fieldname{Text Content}
The smallest dog breed is the chihuahua.

\tokenfield{Speech Tokens}
\tokenseq{[3704, 7192, 6725, 6676, 1959, 3831, 45, 4442,
6792, 1035, 14382, 3410, 9992, 5068, 6148, 4459,
4498, 4429, 6979, 2209, 795, 7411, 353, 4946,
13225, 4770, 975, 6705, 2169, 9941, 5300, 2627,
11198, 484, 786, 7553, 17346, 4454, 8462, 3225,
140, 2336, 1356, 17640, 1072, 626, 1456, 3643,
3488, 25896, 5062, 5830, 2786, 17267, 974, 2927,
5832, 6669, 13230, 22981, 9416]}

\tokenfield{Expanded Speech Tokens}
\tokenseq{[3704, 4299, 6486, 6405, 4218, 4218, 4218, 6405,
6405, 1959, 3831, 45, 4442, 6373, 6377, 1035,
2130, 2125, 3410, 4865, 4942, 5068, 6148, 4459,
4498, 4429, 2139, 4431, 2209, 795, 2139, 66,
353, 4946, 5675, 4942, 4770, 975, 4299, 2166,
2169, 4358, 6050, 5300, 2627, 638, 566, 484,
786, 1763, 1951, 2573, 4760, 4454, 4429, 4272,
3225, 140, 2336, 1356, 5326, 3840, 1072, 626,
1456, 3643, 3488, 6157, 5508, 5062, 5830, 2786,
3161, 974, 974, 2927, 5832, 4299, 6486, 6486,
6486, 6486, 6486, 6486, 6486, 6486, 6486, 6486,
6405, 6405, 6405, 6405, 6405, 6405, 4218, 4218,
4218, 4218, 4218, 4299, 6078]}

\fieldname{Length}
\textbf{61} BPE Speech Tokens / \textbf{101} Expanded Speech Tokens 

\fieldname{Compression Rate} \textbf{60.40\%}

\end{tcolorbox}

\label{fig:case-study-1}
\end{tcolorbox}

\newpage

\begin{tcolorbox}[
    enhanced,
    breakable,
    colback=gray!2,
    colframe=black!55,
    colbacktitle=black!80,
    coltitle=white,
    title=\textbf{Case Study 2: Simple Commonsense QA},
    fonttitle=\bfseries,
    boxrule=0.80pt,
    arc=1.5mm,
    left=1.2mm,
    right=1.2mm,
    top=0.70mm,
    bottom=0.70mm,
    boxsep=0.70mm,
    before skip=0.35em,
    after skip=0.55em,
    pad at break*=0.8mm
]

\fieldname{Input}
\textit{What color is the sky?}

\vspace{0.25em}

\begin{tcolorbox}[
    enhanced,
    breakable,
    colback=cyan!2,
    colframe=gray!35,
    colbacktitle=cyan!10,
    coltitle=black,
    title=\textbf{Step-Audio 2 mini},
    fonttitle=\bfseries\footnotesize,
    boxrule=0.30pt,
    arc=0.8mm,
    borderline west={1.6pt}{0pt}{cyan!55!black},
    left=0.9mm,
    right=0.9mm,
    top=0.45mm,
    bottom=0.45mm,
    boxsep=0.45mm,
    before skip=0.22em,
    after skip=0.22em,
    pad at break*=0.6mm
]

\fieldname{Text Content}
The sky is blue.

\tokenfield{Speech Tokens}
\tokenseq{[1499, 4218, 6486, 4299, 2112, 1029, 74, 2261, 6455, 6374, 1890, 1788,
183, 2270, 5591, 5348, 5186, 4543, 4821, 5011, 1978, 1034, 4517, 5179,
6382, 1044, 2175, 2166, 721, 6475, 4690, 4613, 3155, 884, 2099, 2102,
2186, 2149, 4299, 6486, 6405, 4218]}

\fieldname{Length}
\textbf{42} Speech Tokens

\end{tcolorbox}

\vspace{0.8em}

\begin{tcolorbox}[
    enhanced,
    breakable,
    colback=orange!2,
    colframe=gray!35,
    colbacktitle=orange!10,
    coltitle=black,
    title=\textbf{Vanilla-FT-EN},
    fonttitle=\bfseries\footnotesize,
    boxrule=0.30pt,
    arc=0.8mm,
    borderline west={1.6pt}{0pt}{orange!50!black},
    left=0.9mm,
    right=0.9mm,
    top=0.45mm,
    bottom=0.45mm,
    boxsep=0.45mm,
    before skip=0.22em,
    after skip=0.22em,
    pad at break*=0.6mm
]

\fieldname{Text Content}
The sky is blue.

\tokenfield{Speech Tokens}
\tokenseq{[1517, 4299, 6486, 6405, 4218, 4218, 4137, 1950, 3667, 1563, 73, 4514,
4441, 6455, 6373, 2106, 2031, 174, 2459, 4865, 4871, 4951, 4536, 4813,
1250, 1951, 557, 5093, 4444, 6454, 3456, 2112, 2160, 1208, 5988, 2263,
4606, 4531, 5420, 1775, 2018, 2099, 4286, 2021, 2140, 4299, 4299, 4299,
4299, 4299, 4299, 6486, 6486, 6486, 6486, 6486, 6405, 4218, 6162]}

\fieldname{Length}
\textbf{59} Speech Tokens 

\end{tcolorbox}

\vspace{0.8em}

\begin{tcolorbox}[
    enhanced,
    breakable,
    colback=red!2,
    colframe=gray!35,
    colbacktitle=red!10,
    coltitle=black,
    title=\textbf{CoSTA-EN},
    fonttitle=\bfseries\footnotesize,
    boxrule=0.30pt,
    arc=0.8mm,
    borderline west={1.6pt}{0pt}{red!45!black},
    left=0.9mm,
    right=0.9mm,
    top=0.45mm,
    bottom=0.45mm,
    boxsep=0.45mm,
    before skip=0.22em,
    after skip=0.22em,
    pad at break*=0.6mm
]

\fieldname{Text Content}
The sky is blue.

\tokenfield{Speech Tokens}
\tokenseq{[3704, 9189, 6657, 1764, 2990, 4447, 7478, 3456, 7246, 4514, 5348, 10357,
5269, 4813, 7115, 4858, 4449, 6390, 1776, 8886, 1450, 6476, 5635, 4771,
4772, 2585, 398, 11408, 2096, 6734, 6672, 43408]}

\tokenfield{Expanded Speech Tokens}
\tokenseq{[3704, 4299, 4299, 4218, 4218, 4218, 4299, 4299, 1764, 2990, 4447, 6455,
6380, 3456, 2112, 84, 4514, 5348, 6320, 6158, 5269, 4813, 1277, 2006,
4858, 4449, 6390, 1776, 2148, 2166, 1450, 6476, 5635, 4771, 4772, 2585,
398, 1370, 2102, 2096, 4299, 4299, 4299, 4299, 4299, 4299, 6486, 6486,
6486, 4299, 4218, 4218, 4218, 4218, 4218, 4218, 4218, 4218, 4299, 6078]}

\fieldname{Length}
\textbf{32} BPE Speech Tokens / \textbf{60} Expanded Speech Tokens 

\fieldname{Compression Rate} \textbf{53.33\%}

\end{tcolorbox}

\label{fig:case-study-2}
\end{tcolorbox}

\newpage

\begin{tcolorbox}[
    enhanced,
    breakable,
    colback=gray!2,
    colframe=black!55,
    colbacktitle=black!80,
    coltitle=white,
    title=\textbf{Case Study 3: Sentence Rewriting},
    fonttitle=\bfseries,
    boxrule=0.80pt,
    arc=1.5mm,
    left=1.2mm,
    right=1.2mm,
    top=0.70mm,
    bottom=0.70mm,
    boxsep=0.70mm,
    before skip=0.35em,
    after skip=0.55em,
    pad at break*=0.8mm
]

\fieldname{Input}
\textit{The order in which you add  vectors does not matter.  Please rewrite in simple and same meaning.}

\vspace{0.25em}

\begin{tcolorbox}[
    enhanced,
    breakable,
    colback=cyan!2,
    colframe=gray!35,
    colbacktitle=cyan!10,
    coltitle=black,
    title=\textbf{Step-Audio 2 mini},
    fonttitle=\bfseries\footnotesize,
    boxrule=0.30pt,
    arc=0.8mm,
    borderline west={1.6pt}{0pt}{cyan!55!black},
    left=0.9mm,
    right=0.9mm,
    top=0.45mm,
    bottom=0.45mm,
    boxsep=0.45mm,
    before skip=0.22em,
    after skip=0.22em,
    pad at break*=0.6mm
]

\fieldname{Text Content}
The order of adding vectors doesn't matter.

\tokenfield{Speech Tokens}
\tokenseq{[1499, 4218, 6486, 4299, 2112, 57, 2510, 566, 1763, 2032, 1358, 2906,
5102, 5815, 6058, 1518, 6067, 6059, 5329, 6050, 1760, 1001, 4970, 4275,
6453, 1517, 1951, 3650, 5834, 3647, 5837, 138, 4445, 5047, 6181, 4235,
4284, 6534, 110, 5918, 4570, 1788, 1869, 2247, 3105, 3098, 1685, 1688,
1697, 5912, 4671, 6374, 6373, 4245, 4299, 2085, 2253, 2, 2702, 5653,
5652, 4438, 5853, 1744, 1803, 4068, 5105, 5194, 5508, 5185, 867, 3146,
1676, 1679, 3872, 3847, 3729, 6486, 6486, 6405, 3975]}

\fieldname{Length}
\textbf{81} Speech Tokens

\end{tcolorbox}

\vspace{0.8em}
\begin{tcolorbox}[
    enhanced,
    breakable,
    colback=orange!2,
    colframe=gray!35,
    colbacktitle=orange!10,
    coltitle=black,
    title=\textbf{Vanilla-FT-EN},
    fonttitle=\bfseries\footnotesize,
    boxrule=0.30pt,
    arc=0.8mm,
    borderline west={1.6pt}{0pt}{orange!50!black},
    left=0.9mm,
    right=0.9mm,
    top=0.45mm,
    bottom=0.45mm,
    boxsep=0.45mm,
    before skip=0.22em,
    after skip=0.22em,
    pad at break*=0.6mm
]

\fieldname{Text Content}
When you add vectors, it doesn't matter which one comes first.

\tokenfield{Speech Tokens}
\tokenseq{[3704, 4299, 6486, 6405, 4218, 4218, 4218, 2121, 4316, 4346, 3644, 2567,
4379, 4471, 4966, 4885, 3482, 4130, 6071, 1517, 1978, 764, 4463, 4382,
5111, 4459, 57, 2417, 2152, 2074, 2097, 2219, 5191, 4536, 2112, 162,
1113, 4431, 3012, 3116, 1676, 1679, 1688, 1616, 5911, 4672, 6134, 762,
4056, 3975, 4299, 4299, 1124, 5102, 4697, 1536, 2049, 1770, 59, 3650,
4456, 5412, 5418, 4494, 5937, 2733, 132, 4074, 1639, 2195, 5108, 4540,
138, 3824, 1658, 1685, 1928, 1454, 2186, 2185, 2912, 2408, 4849, 438,
4758, 3864, 1380, 4344, 5101, 5029, 4862, 11, 50, 1726, 1805, 1833,
165, 5346, 5356, 5591, 666, 6498, 5409, 6374, 2916, 6537, 6534, 5780,
5320, 3863, 1685, 1619, 4509, 6376, 2139, 2247, 2193, 736, 4299, 4299,
6486, 6486, 6486, 6486, 6486, 6486, 6486, 6486, 4299, 4218, 4218, 4218,
4218, 4218, 4218, 4218, 4299, 6159]}

\fieldname{Length}
\textbf{138} Speech Tokens

\end{tcolorbox}

\vspace{0.8em}

\begin{tcolorbox}[
    enhanced,
    breakable,
    colback=red!2,
    colframe=gray!35,
    colbacktitle=red!10,
    coltitle=black,
    title=\textbf{CoSTA-EN},
    fonttitle=\bfseries\footnotesize,
    boxrule=0.30pt,
    arc=0.8mm,
    borderline west={1.6pt}{0pt}{red!45!black},
    left=0.9mm,
    right=0.9mm,
    top=0.45mm,
    bottom=0.45mm,
    boxsep=0.45mm,
    before skip=0.22em,
    after skip=0.22em,
    pad at break*=0.6mm
]

\fieldname{Text Content}
The order doesn't matter when you add vectors.

\tokenfield{Speech Tokens}
\tokenseq{[3704, 9189, 6676, 1725, 14093, 137, 8715, 2789, 2708, 4976, 4842, 297,
4601, 6864, 13239, 5189, 5348, 14410, 4439, 13344, 27059, 11544, 189, 2332,
7868, 2171, 18623, 110, 2276, 4643, 3454, 4129, 4367, 9321, 5435, 5920,
5108, 57, 1855, 2073, 1854, 73052, 11434, 7732, 3078, 3116, 11361, 3725,
4683, 4673, 7404, 1731, 6678, 20923, 6078]}

\tokenfield{Expanded Speech Tokens}
\tokenseq{[3704, 4299, 4299, 4218, 4218, 4218, 6405, 6405, 1725, 1941, 63, 137,
2003, 2006, 2789, 2708, 4976, 4842, 297, 4601, 6059, 6058, 1689, 57,
5189, 5348, 5409, 5652, 4439, 5853, 5857, 4068, 173, 5192, 5917, 189,
2332, 3863, 1685, 2171, 1456, 1456, 110, 2276, 4643, 3454, 4129, 4367,
1763, 1979, 5435, 5920, 5108, 57, 1855, 2073, 1854, 2195, 5998, 2383,
6486, 2112, 4431, 3078, 3116, 1685, 1697, 3725, 4683, 4673, 5644, 5643,
1731, 4299, 4299, 4299, 4299, 4299, 4299, 4299, 4299, 4218, 4218, 4218,
4218, 4218, 4218, 4218, 4218, 4218, 4218, 4218, 4218, 6078]}

\fieldname{Length}
\textbf{55} BPE Speech Tokens / \textbf{94} Expanded Speech Tokens

\fieldname{Compression Rate} \textbf{58.51\%}

\end{tcolorbox}

\label{fig:case-study-3}
\end{tcolorbox}

\newpage

\begin{tcolorbox}[
    enhanced,
    breakable,
    colback=gray!2,
    colframe=black!55,
    colbacktitle=black!80,
    coltitle=white,
    title=\textbf{Case Study 4: Open-ended Dialogue},
    fonttitle=\bfseries,
    boxrule=0.80pt,
    arc=1.5mm,
    left=1.2mm,
    right=1.2mm,
    top=0.70mm,
    bottom=0.70mm,
    boxsep=0.70mm,
    before skip=0.35em,
    after skip=0.55em,
    pad at break*=0.8mm
]

\fieldname{Input}
\textit{What is the effect of saying hello to the person?}

\vspace{0.25em}

\begin{tcolorbox}[
    enhanced,
    breakable,
    colback=cyan!2,
    colframe=gray!35,
    colbacktitle=cyan!10,
    coltitle=black,
    title=\textbf{Step-Audio 2 mini},
    fonttitle=\bfseries\footnotesize,
    boxrule=0.30pt,
    arc=0.8mm,
    borderline west={1.6pt}{0pt}{cyan!55!black},
    left=0.9mm,
    right=0.9mm,
    top=0.45mm,
    bottom=0.45mm,
    boxsep=0.45mm,
    before skip=0.22em,
    after skip=0.22em,
    pad at break*=0.6mm
]

\fieldname{Text Content}
Saying "Hello" is an act of greeting, which can have several effects on both parties involved.

\tokenfield {Speech Tokens}
\tokenseq{[1499, 4218, 6486, 4299, 3894, 5158, 6455, 6458, 3222, 2246, 2195, 2573,
2825, 2834, 2753, 2807, 3995, 6182, 6181, 3984, 2031, 5835, 5838, 3651,
2112, 5109, 5832, 2918, 1046, 3962, 6148, 2486, 4780, 4806, 5077, 2813,
635, 728, 1457, 3644, 5827, 2028, 3645, 2922, 3165, 2032, 1951, 557,
4526, 6382, 3231, 2243, 3666, 1480, 1735, 1708, 5834, 5915, 5914, 4543,
1869, 165, 1896, 4431, 1520, 1978, 515, 5078, 4284, 6534, 2109, 1221,
657, 4367, 6071, 5587, 5100, 231, 75, 2816, 620, 1808, 4076, 3994,
4227, 4299, 3975, 5919, 5916, 3648, 3888, 4131, 4215, 4299, 2113, 3643,
5582, 240, 5244, 3135, 1788, 2355, 4390, 19, 1479, 5106, 5841, 3647,
326, 2169, 4509, 6374, 6377, 5652, 4406, 4537, 4356, 6543, 6310, 5086,
5097, 2003, 1028, 4519, 6453, 6537, 6540, 2656, 4540, 2924, 737, 197,
1203, 813, 4191, 6131, 5402, 5644, 1759, 1979, 5591, 4874, 587, 4314,
2130, 2139, 712, 626, 5087, 5094, 4770, 6375, 1893, 2166, 5670, 5589,
5753, 6266, 6050, 957, 2267, 566, 1295, 2996, 6373, 795, 2006, 4769,
1638, 4073, 2074, 6546, 713, 4946, 5022, 5053, 2789, 629, 719, 2915,
4368, 6507, 1032, 2241, 732, 4299]}

\fieldname{Length}
\textbf{186} Speech Tokens

\end{tcolorbox}

\vspace{0.8em}

\begin{tcolorbox}[
    enhanced,
    breakable,
    colback=orange!2,
    colframe=gray!35,
    colbacktitle=orange!10,
    coltitle=black,
    title=\textbf{Vanilla-FT-EN},
    fonttitle=\bfseries\footnotesize,
    boxrule=0.30pt,
    arc=0.8mm,
    borderline west={1.6pt}{0pt}{orange!50!black},
    left=0.9mm,
    right=0.9mm,
    top=0.45mm,
    bottom=0.45mm,
    boxsep=0.45mm,
    before skip=0.22em,
    after skip=0.22em,
    pad at break*=0.6mm
]

\fieldname{Text Content}
Saying hello greets them warmly, fostering connection and mutual respect. A simple gesture that often softens interactions, especially with those we've not met before.

\tokenfield{Speech Tokens}
\tokenseq{[3704, 6486, 6486, 4299, 4218, 5133, 5401, 6377, 5648, 4451, 2222, 4841,
5102, 4994, 5047, 6180, 4642, 5106, 5832, 5842, 3647, 3719, 5986, 47,
4457, 4807, 4997, 623, 632, 1370, 2184, 489, 1450, 4367, 3872, 4598,
4857, 5100, 1194, 4679, 4435, 1734, 1833, 1035, 59, 2216, 2124, 4316,
4319, 2130, 1430, 1457, 2870, 6482, 6238, 4311, 6174, 3206, 4043, 29,
386, 647, 1376, 2807, 2517, 6486, 6486, 3975, 5838, 5835, 3645, 2112,
6426, 6534, 3537, 491, 4942, 4411, 5644, 2139, 4431, 2290, 6059, 6554,
6302, 5336, 5075, 6181, 6502, 1815, 168, 2203, 1479, 3669, 2228, 5920,
4539, 1869, 3108, 5319, 4593, 159, 143, 32, 1482, 1484, 1483, 1735,
1951, 2924, 4382, 750, 1725, 161, 2127, 2128, 1646, 3563, 5342, 4851,
951, 6055, 6051, 3099, 4772, 5009, 5000, 5086, 4284, 4366, 3863, 4499,
4428, 6376, 1896, 2166, 4511, 5111, 1466, 815, 1923, 921, 3054, 1275,
4432, 2193, 735, 6486, 6486, 6486, 6486, 6486, 6486, 6405, 6162, 3894,
3648, 1950, 1703, 2918, 2254, 6455, 6380, 807, 4445, 6012, 2175, 3591,
5065, 5005, 885, 2490, 3147, 2330, 4459, 4414, 4672, 2814, 5487, 6054,
1673, 1679, 1676, 1685, 3875, 3813, 3651, 5919, 2922, 1545, 2085, 63,
2195, 220, 1516, 2003, 3407, 4949, 5026, 3429, 6507, 1032, 2217, 4495,
5124, 6099, 5645, 6374, 5643, 4727, 4945, 5053, 6345, 6537, 6453, 146,
68, 1480, 1481, 3671, 6385, 5404, 1763, 1285, 5021, 3751, 1887, 4515,
1632, 4115, 1598, 1493, 5921, 5998, 2379, 1569, 6021, 5322, 3138, 890,
65, 1477, 3668, 5858, 5402, 5405, 33, 2031, 5109, 3732, 5838, 5838,
1464, 1978, 1490, 4379, 4441, 6373, 2112, 2079, 2189, 5188, 4498, 4590,
4509, 4654, 4910, 4934, 4442, 5020, 4285, 1455, 1208, 2411, 4444, 6130,
1959, 2100, 73, 2300, 4835, 2810, 641, 2099, 1856, 5909, 5653, 2970,
1788, 5916, 2919, 1701, 3570, 3643, 3314, 4859, 3067, 6264, 1722, 3670,
1476, 5105, 5671, 5022, 111, 2733, 2220, 1641, 4071, 911, 3653, 5108,
138, 4920, 924, 2163, 3157, 5750, 2986, 6507, 6534, 2884, 5084, 2906,
719, 680, 1865, 1649, 3836, 3844, 3726, 6486, 6486, 6486, 6486, 6486,
6486, 6486, 6486, 6486, 6486, 6486, 6486, 6486, 6486, 6486, 4299, 4299,
4299, 6159]}

\fieldname{Length}
\textbf{350} Speech Tokens

\end{tcolorbox}

\vspace{0.8em}

\begin{tcolorbox}[
    enhanced,
    breakable,
    colback=red!2,
    colframe=gray!35,
    colbacktitle=red!10,
    coltitle=black,
    title=\textbf{CoSTA-EN},
    fonttitle=\bfseries\footnotesize,
    boxrule=0.30pt,
    arc=0.8mm,
    borderline west={1.6pt}{0pt}{red!45!black},
    left=0.9mm,
    right=0.9mm,
    top=0.45mm,
    bottom=0.45mm,
    boxsep=0.45mm,
    before skip=0.22em,
    after skip=0.22em,
    pad at break*=0.6mm
]

\fieldname{Text Content}
Saying hello fosters connection and breaks ice, especially meaningful from experience.

\tokenfield{Speech Tokens}
\tokenseq{[2975, 9039, 1950, 5136, 5404, 7084, 2982, 2267, 2573, 4742, 9727, 4885,
7184, 5839, 5, 3722, 5989, 47, 4540, 4726, 2729, 16524, 728, 725,
2028, 6768, 6991, 8720, 685, 10458, 13823, 7472, 2196, 3107, 11971, 4531,
46605, 7046, 93, 747, 5856, 2233, 9501, 1896, 141, 9701, 10551, 6955,
3668, 1501, 6776, 6949, 11175, 3671, 2155, 2175, 10911, 7151, 9950, 2904,
7161, 6912, 801, 1735, 1948, 7536, 4871, 731, 12873, 3563, 4510, 5405,
10063, 762, 6710, 8056, 7153, 2582, 4535, 7282, 6921, 2297, 13982, 4404,
8154, 4682, 4910, 5744, 2252, 5020, 913, 15772, 944, 5021, 4821, 6015,
4480, 4966, 4238, 2076, 6674, 716, 8929, 2915, 4278, 8502, 3872, 3242,
26855, 3586, 4312, 7604, 4849, 1140, 24474, 1839, 11667, 5101, 5335, 6293,
11772, 152, 14798, 8080, 5421, 8485, 5407, 771, 6669, 28091]}

\tokenfield{Expanded Speech Tokens}
\tokenseq{[2975, 4299, 6486, 6486, 6486, 6486, 4299, 4218, 1950, 5136, 5404, 6376,
6377, 2982, 2267, 2573, 4742, 5048, 5047, 4885, 5352, 5109, 5839, 5,
3722, 5989, 47, 4540, 4726, 2729, 551, 635, 728, 725, 2028, 3888,
3888, 1701, 2112, 6426, 6534, 6534, 685, 4949, 5022, 4410, 6454, 2112,
2244, 2196, 3107, 3863, 6062, 4531, 5725, 3213, 2112, 87, 93, 747,
5856, 2233, 5921, 4539, 1896, 141, 5319, 4593, 2328, 890, 59, 1479,
1484, 3668, 1501, 2112, 1736, 4463, 4382, 49, 3669, 3671, 2155, 2175,
2179, 4367, 6302, 6050, 4598, 4849, 2904, 4299, 168, 5158, 6374, 801,
1735, 1948, 6320, 5671, 4871, 731, 359, 647, 3563, 4510, 5405, 6133,
6376, 762, 4299, 4299, 4299, 4218, 3651, 2922, 2112, 1951, 2582, 4535,
5158, 6376, 2139, 2166, 2297, 5191, 4543, 4404, 5319, 4509, 4682, 4910,
5744, 2252, 5020, 913, 2131, 2127, 944, 5021, 4821, 6015, 4480, 4966,
4238, 2076, 6534, 6534, 716, 626, 626, 2915, 4278, 6534, 6534, 4356,
3872, 3242, 5591, 4955, 3586, 4312, 1763, 1277, 4849, 1140, 3024, 6373,
1839, 2166, 954, 5101, 5335, 6293, 4358, 1685, 152, 386, 35, 1480,
3671, 5421, 5402, 5405, 5407, 771, 4299, 6486, 6486, 6486, 6486, 6486,
6486, 6486, 6486, 4299, 4299, 4299, 4299, 4299, 4299, 4299, 4299, 4299,
4299, 4299, 6159]}

\fieldname{Length}
\textbf{130} BPE Speech Tokens / \textbf{207} Expanded Speech Tokens

\fieldname{Compression Rate} \textbf{62.80\%}

\end{tcolorbox}

\label{fig:case-study-4}
\end{tcolorbox}

\newpage

\begin{tcolorbox}[
    enhanced,
    breakable,
    colback=gray!2,
    colframe=black!55,
    colbacktitle=black!80,
    coltitle=white,
    title=\textbf{Case Study 5: Empathetic Customer Dialogue},
    fonttitle=\bfseries,
    boxrule=0.80pt,
    arc=1.5mm,
    left=1.2mm,
    right=1.2mm,
    top=0.70mm,
    bottom=0.70mm,
    boxsep=0.70mm,
    before skip=0.35em,
    after skip=0.55em,
    pad at break*=0.8mm
]

\fieldname{Input}
\textit{What to say a customer that mentioned their father is in the hospital?}

\vspace{0.25em}

\begin{tcolorbox}[
    enhanced,
    breakable,
    colback=cyan!2,
    colframe=gray!35,
    colbacktitle=cyan!10,
    coltitle=black,
    title=\textbf{Step-Audio 2 mini},
    fonttitle=\bfseries\footnotesize,
    boxrule=0.30pt,
    arc=0.8mm,
    borderline west={1.6pt}{0pt}{cyan!55!black},
    left=0.9mm,
    right=0.9mm,
    top=0.45mm,
    bottom=0.45mm,
    boxsep=0.45mm,
    before skip=0.22em,
    after skip=0.22em,
    pad at break*=0.6mm
]

\fieldname{Text Content}
I'm really sorry to hear that your father is in the hospital. Is there anything I can do for you?

\tokenfield{Speech Tokens}
\tokenseq{[1502, 1978, 4218, 4137, 1950, 1951, 3404, 3647, 203, 2124, 2128, 2152,
2179, 3881, 5587, 4603, 6066, 2267, 566, 4431, 4672, 5645, 2484, 2702,
4949, 5779, 6293, 1685, 404, 303, 2217, 323, 4380, 4704, 2760, 5749,
6061, 3872, 63, 315, 2, 1763, 2023, 4129, 1928, 6534, 6537, 3618, 5675,
6319, 4699, 543, 2252, 4115, 1676, 152, 4680, 2493, 20, 3912, 46, 1217,
5589, 6318, 5591, 5671, 4384, 4915, 1191, 2160, 227, 873, 1352, 629,
2825, 2689, 2031, 1944, 3645, 3648, 5835, 3651, 1464, 1707, 3894, 1707,
1950, 1950, 1735, 2258, 5179, 6381, 2013, 2255, 3863, 1919, 2005, 764,
8, 3666, 4966, 4993, 6426, 4185, 4415, 4805, 5452, 1760, 1946, 4544,
447, 168, 25, 3667, 1563, 60, 1700, 4127, 5824, 6507, 6534, 3600, 6059,
5668, 4939, 5345, 5342, 5068, 5043, 3495, 4299, 4218, 4137, 4137, 4137]}

\fieldname{Length}
\textbf{135} Speech Tokens

\end{tcolorbox}

\vspace{0.8em}

\begin{tcolorbox}[
    enhanced,
    breakable,
    colback=orange!2,
    colframe=gray!35,
    colbacktitle=orange!10,
    coltitle=black,
    title=\textbf{Vanilla-FT-EN},
    fonttitle=\bfseries\footnotesize,
    boxrule=0.30pt,
    arc=0.8mm,
    borderline west={1.6pt}{0pt}{orange!50!black},
    left=0.9mm,
    right=0.9mm,
    top=0.45mm,
    bottom=0.45mm,
    boxsep=0.45mm,
    before skip=0.22em,
    after skip=0.22em,
    pad at break*=0.6mm
]

\fieldname{Text Content}
I'm sorry to hear about your father. Is there anything I can do to help?

\tokenfield{Speech Tokens}
\tokenseq{[3704, 4299, 6486, 6405, 4218, 4218, 1951, 4133, 6320, 5105, 4382, 4985,
214, 2127, 1481, 5401, 6377, 6373, 74, 3161, 5753, 6545, 6302, 5297,
4849, 1167, 2490, 2382, 6044, 3725, 5190, 4947, 2778, 5100, 5101, 4859,
962, 1679, 1688, 1601, 1490, 5348, 387, 2166, 380, 6077, 4862, 5006,
1110, 4431, 2760, 4291, 6071, 3620, 4079, 6534, 6537, 6453, 2459, 5185,
4455, 153, 2016, 32, 1652, 3875, 3884, 1610, 1843, 4299, 4299, 4299,
4299, 3894, 5838, 5109, 1788, 2031, 395, 4850, 4444, 6130, 2112, 1029,
2267, 197, 3836, 3866, 1763, 764, 2924, 4462, 3018, 4399, 5073, 4407,
6457, 6129, 4535, 5021, 5047, 6097, 1705, 1948, 3890, 5837, 4814, 681,
168, 2283, 2195, 753, 1727, 1914, 2247, 1862, 5993, 5500, 5580, 951,
4434, 897, 4367, 68, 4380, 5118, 8, 734, 2675, 4972, 5088, 2139, 2160,
729, 1869, 6486, 6486, 6486, 6486, 6486, 6486, 6486, 4299, 4299, 4299,
4299, 4218, 4218, 4218, 4218, 4299, 6078]}

\fieldname{Length}
\textbf{152} Speech Tokens

\end{tcolorbox}

\vspace{0.8em}

\begin{tcolorbox}[
    enhanced,
    breakable,
    colback=red!2,
    colframe=gray!35,
    colbacktitle=red!10,
    coltitle=black,
    title=\textbf{CoSTA-EN},
    fonttitle=\bfseries\footnotesize,
    boxrule=0.30pt,
    arc=0.8mm,
    borderline west={1.6pt}{0pt}{red!45!black},
    left=0.9mm,
    right=0.9mm,
    top=0.45mm,
    bottom=0.45mm,
    boxsep=0.45mm,
    before skip=0.22em,
    after skip=0.22em,
    pad at break*=0.6mm
]

\fieldname{Text Content}
I'm sorry to hear that. Is there anything I can do to help?

\tokenfield{Speech Tokens}
\tokenseq{[3713, 7315, 7077, 7491, 4462, 5065, 4984, 2474, 8, 909, 1805, 3721,
6792, 6373, 1046, 245, 4943, 5779, 6544, 7358, 4490, 10646, 240, 7328,
903, 4286, 2946, 4785, 9778, 4939, 4606, 6061, 6062, 805, 21388, 9488,
737, 14223, 1761, 8607, 6933, 6720, 6750, 11002, 17816, 8103, 2267, 224,
9508, 7131, 2195, 17344, 5064, 4740, 6507, 6348, 2943, 4526, 4985, 8428,
1474, 1948, 7288, 4814, 681, 9299, 9677, 1727, 17688, 1862, 3887, 5258,
63397, 876, 7309, 912, 1370, 2919, 7800, 9462, 2675, 21749, 1437, 6935,
999, 1809, 6678, 21513]}

\tokenfield{Expanded Speech Tokens}
\tokenseq{[3713, 4299, 4299, 4299, 4299, 4299, 4299, 4218, 1951, 1946, 4133, 6077,
4462, 5065, 4984, 2474, 8, 909, 1805, 3721, 6373, 6377, 6373, 1046, 245,
4943, 5779, 6544, 6553, 6050, 4490, 2825, 2834, 240, 2166, 4435, 903,
4286, 2946, 4785, 4947, 2697, 4939, 4606, 6061, 6062, 805, 2094, 65,
2195, 2921, 737, 59, 1923, 1761, 4435, 2274, 735, 4299, 4299, 6486,
4299, 4299, 4299, 4299, 4218, 1951, 1295, 5021, 5182, 6373, 2112, 1029,
2267, 224, 3863, 1679, 1763, 737, 2195, 2370, 2938, 5064, 4740, 6507,
6348, 2943, 4526, 4985, 5047, 6180, 1474, 1948, 4133, 5834, 4814, 681,
168, 2283, 2228, 3669, 1727, 2184, 2490, 1862, 3887, 5258, 5743, 5823,
876, 546, 4596, 912, 1370, 2919, 5109, 2932, 737, 1463, 2675, 5080,
5088, 1437, 2166, 1431, 999, 1809, 4299, 4299, 4299, 4299, 4299, 4299,
4299, 4299, 4299, 4299, 4299, 4299, 4299, 4299, 4299, 4299, 4299, 4299,
4299, 4299, 6078]}

\fieldname{Length}
\textbf{88} BPE Speech Tokens / \textbf{148} Expanded Speech Tokens

\fieldname{Compression Rate} \textbf{59.46\%}

\end{tcolorbox}

\label{fig:case-study-5}
\end{tcolorbox}

\endgroup
\clearpage
\twocolumn